\documentclass[journal]{IEEEtran}

\usepackage[english]{babel}
\usepackage[utf8x]{inputenc}
\usepackage[T1]{fontenc}

\usepackage{amsmath,amssymb,mathrsfs,bbm}
\usepackage{graphicx}
\usepackage[colorinlistoftodos]{todonotes}
\usepackage[colorlinks=true, allcolors=blue]{hyperref}
\usepackage{multirow}
\usepackage[lined,linesnumbered,ruled]{algorithm2e}
\usepackage[font=footnotesize]{subcaption}
\usepackage[font=footnotesize]{caption}

\newcommand{\cp}[1]{\ifmmode {\mathcal{#1}}\else ${\mathcal{#1}}$\fi}
	
\newcommand{\bA}{\boldsymbol{A}}
\newcommand{\bB}{\boldsymbol{B}}

\newcommand{\bE}{\boldsymbol{E}}

\newcommand{\bI}{\boldsymbol{I}}

\newcommand{\bM}{\boldsymbol{M}}

\newcommand{\bW}{\boldsymbol{W}}

\newcommand{\bY}{\boldsymbol{Y}}

\newcommand{\ba}{\boldsymbol{a}}

\newcommand{\bd}{\boldsymbol{d}}
\newcommand{\bm}{\boldsymbol{m}}

\newcommand{\be}{\boldsymbol{e}}

\newcommand{\by}{\boldsymbol{y}}

\newcommand{\bx}{\boldsymbol{x}}

\newcommand{\calC}{\mathcal{C}}
\newcommand{\calD}{\mathcal{D}}
\newcommand{\calW}{\mathcal{W}}

\newcommand{\bbM}{\mathbb{M}}

\newcommand{\bpsi}{\boldsymbol{\psi}}

\newcommand{\btheta}{\boldsymbol{\theta}}

\newcommand{\bPsi}{\boldsymbol{\Psi}}

\newcommand{\bTheta}{\boldsymbol{\Theta}}

\newcommand{\cb}[1]{\boldsymbol{#1}}
\newcommand{\tr}{\mathrm{tr}}
\newcommand{\diag}{\mathrm{diag}}
\newcommand{\vect}{\mathrm{vec}}

\usepackage{color}  

\definecolor{darkgreen}{rgb}{0.0, 0.85, 0.0}

\makeatletter
\newcommand{\doublewidetilde}[1]{{\mathpalette\double@widetilde{#1}}}
\newcommand{\double@widetilde}[2]{\sbox\z@{$\m@th#1\widetilde{#2}$} \ht\z@=.9\ht\z@\widetilde{\box\z@}}
\makeatother

\title{A Data Dependent Multiscale Model for Hyperspectral Unmixing With Spectral Variability}

\author{Ricardo~Augusto~Borsoi,~\IEEEmembership{Student Member,~IEEE},
Tales Imbiriba,~\IEEEmembership{Member,~IEEE}, 
Jos\'e~Carlos~Moreira~Bermudez,~\IEEEmembership{Senior~Member,~IEEE}
\thanks{This work has been supported by the National Council for Scientific and Technological Development (CNPq) under grants 304250/2017-1, 409044/2018-0, 141271/2017-5 and 204991/2018-8, and by the Brazilian Education Ministry (CAPES) under grant PNPD/1811213.}
\thanks{The authors would like to thank Lucas Drumetz and his collaborators for providing part of the data used in the experimental section of the manuscript.}
\thanks{R.A. Borsoi is with the Department of Electrical Engineering, Federal University of Santa Catarina (DEE--UFSC), Florian\'opolis, SC, Brazil, and with the Lagrange Laboratory, Universit\'e  C\^ote  d'Azur, Nice, France. e-mail: \mbox{raborsoi@gmail.com}.
T. Imbiriba was with DEE--UFSC, Florian\'opolis, SC, Brazil, and is with the ECE department of the Northeastern University, Boston, MA, USA. e-mail: \mbox{talesim@ece.neu.edu}.
J.C.M. Bermudez is with the DEE--UFSC, Florian\'opolis, SC, Brazil, and with the Graduate Program on Electronic Engineering and Computing, Catholic University of Pelotas (UCPel) Pelotas, Brazil. e-mail: \mbox{j.bermudez@ieee.org}.}
\thanks{This paper has supplementary downloadable material available at http://ieeexplore.ieee.org., provided by the authors. The material includes more detailed experimental validations. Contact \mbox{raborsoi@gmail.com} for further questions about this work.}
\thanks{Manuscript received Month day, year; revised Month day, year.}
}

\begin{document}
\maketitle

\begin{abstract}
Spectral variability in hyperspectral images can result from factors including environmental, illumination, atmospheric and temporal changes. Its occurrence may lead to the propagation of significant estimation errors in the unmixing process. To address this issue, extended linear mixing models have been proposed which lead to large scale nonsmooth ill-posed inverse problems. Furthermore, the regularization strategies used to obtain meaningful results have introduced interdependencies among abundance solutions that further increase the complexity of the resulting optimization problem. In this paper we present a novel data dependent multiscale model for hyperspectral unmixing accounting for spectral variability. The new method incorporates spatial contextual information to the abundances in extended linear mixing models by using a multiscale transform based on superpixels. The proposed method results in a fast algorithm that solves the abundance estimation problem only once in each scale during each iteration. Simulation results using synthetic and real images compare the performances, both in accuracy and execution time, of the proposed algorithm and other state-of-the-art solutions.
\end{abstract}

\begin{IEEEkeywords}
Hyperspectral data, spectral variability, spatial regularization, multiscale, superpixels.
\end{IEEEkeywords}

\section{Introduction}

Hyperspectral devices acquire hundreds of contiguous reflectance samples from the observed electromagnetic spectra. This observed reflectance is often mixed at the pixel level and requires unmixing strategies to correctly unveil important information about the materials and their proportion in a target scene~\cite{Keshava:2002p5667}.  
Hyperspectral unmixing (HU) aims at decomposing the observed reflectance in pure spectral components, \emph{i.e.}, \emph{endmembers}, and their proportions~\cite{Keshava:2002p5667}, commonly referred as fractional \emph{abundances}. 
Different models and strategies have been proposed to solve this problem~\cite{Bioucas-Dias-2013-ID307,Dobigeon-2014-ID322,Zare-2014-ID324}. The vast majority of methods considers the Linear Mixing Model (LMM)~\cite{Keshava:2002p5667}, which assumes that the observed reflectance vector (\emph{i.e.} a hyperspectral image pixel) can be modeled as a convex combination of the endmembers present in the scene.
Although this assumption naturally leads to fast and reliable unmixing strategies, the intrinsic limitation of the LMM cannot cope with relevant nonideal effects intrinsic to practical applications~\cite{Dobigeon-2014-ID322,Imbiriba2016_tip, Imbiriba2017_bs_tip}. One such important nonideal effect is
endmember variability~\cite{Zare-2014-ID324, drumetz2016variabilityReviewRecent}.

Endmember variability can be caused by a myriad of factors including environmental conditions, illumination, atmospheric and temporal changes~\cite{Zare-2014-ID324}. Its occurrence may result in significant estimation errors being propagated
throughout the unmixing process~\cite{thouvenin2016hyperspectralPLMM}. The most common approaches to deal with spectral variability can be divided into three basic classes. 1) to group endmembers in variational sets, 2) to model endmembers as statistical distributions, and 3) to incorporate the variability in the mixing model, often using physically motivated concepts~\cite{drumetz2016variabilityReviewRecent}. This work follows the third approach. Recently, \cite{thouvenin2016hyperspectralPLMM}, \cite{drumetz2016blindUnmixingELMMvariability} and \cite{imbiriba2018GLMM} introduced variations of the LMM to cope with  spectral variability. 
The Perturbed LMM model (PLMM)~\cite{thouvenin2016hyperspectralPLMM} introduces an additive perturbation to the endmember matrix. Such perturbation matrix then needs to be estimated jointly with the abundances. Though the perturbation matrix can model arbitrary endmember variations, it lacks physical motivation. 
The Extended Linear Mixing Model (ELMM) proposed in~\cite{drumetz2016blindUnmixingELMMvariability} increased the flexibility of the LMM model by associating a pixel-dependent multiplicative term to each endmember. This generalization can efficiently model changes in the observed reflectances due to illumination, an important effect~\cite{drumetz2016blindUnmixingELMMvariability}. 
This model addresses a physically motivated problem, with the advantage of estimating a variability parameter vector of much lower dimension when compared with the additive perturbation matrix in PLMM.
%
Although the ELMM performs well in situations where spectral variability is mainly caused by illumination variations, it lacks flexibility when the endmembers are subject to more complex spectral distortions.  This motivated the use of additive low-rank terms to the ELMM to deal with more complex types of spectral variability~\cite{hong2019augmentedLMMvariability}. However, this approach does not provide an explicit representation for the endmembers. In~\cite{imbiriba2018GLMM} a physically-motivated generalization to the ELMM was proposed resulting in the Generalized Linear Mixing Model (GLMM). The GLMM accounted for arbitrary spectral variations in each endmember by allowing the multiplicative scaling factors to vary according to the spectral bands, leading to an increased amount of freedom when compared \mbox{to the ELMM.}


Though the above described models were shown to be capable to model endmember variability effects with good accuracy, their use in HU leads to severely ill-posed inverse problems, which require sound regularization strategies to yield meaningful solutions.
%
One way to mitigate this ill-posedness is to explore spatial correlations found in typical abundance \cite{eches2011enhancingHSspatialCorrelations} and variability \cite{drumetz2016blindUnmixingELMMvariability} maps.
%
%
%
For instance, spatial information has been employed both for endmember extraction~\cite{zortea2009spatialPreProcessingEMextraction,torres2014SpatialMultiscaleEMextraction} and for regularization in linear~\cite{zymnis2007HSlinearUnmixingTV}, nonlinear~\cite{Jchen2014nonlinearSpatialTVreg}, Bayesian~\cite{eches2011enhancingHSspatialCorrelations,chen2017sparseBayesianMRFunmixingHS,altmann2015robustUnmixingAnomaly} and sparse~\cite{iordache2012sparseUnmixingTV} unmixing strategies.
Total variation (TV) deserves special mention as a spatial regularization approach that promotes spatially piecewise homogeneous solutions without compromising sharp discontinuities between neighboring pixels. This property is important to handle the type of spatial correlation found in many hyperspectral unmixing applications~\cite{shi2014surveySpatialRegUnmixing,afonso2011augmented}.

Although important to mitigate the ill-posedness of the inverse problem, the use of spatial regularization in spectral-variability-aware HU introduces interdependencies among abundance solutions for different image pixels. This in turn leads to intricate, large scale and computationally demanding optimization problems.
Even though some approaches have been investigated to accelerate the minimization of convex TV-regularized functionals~\cite{chambolle2017acceleratedAlternatingDescentDykstra,chambolle2011primalDualTV}, this is still a computationally demanding operation which, in the context of HU, have been primarily addressed using variable splitting (e.g. ADMM) techniques~\cite{thouvenin2016hyperspectralPLMM,drumetz2016blindUnmixingELMMvariability,imbiriba2018GLMM}.
%
Such complexity is usually incompatible with recent demands for timely processing of vast amounts of remotely sensed data required by many modern real world applications~\cite{ma2015BigDataRemoteSensing,chi2016BigDataRemoteSensing}.
Thus, it is desirable to search for faster and lower complexity strategies that yield comparable unmixing performances.



Two recent works have proposed new regularization techniques for ill-posed HU problems aimed at avoiding the interdependency between pixels introduced by standard regularization methods. In~\cite{imbiriba2018_ULTRA} a low-rank tensor regularization strategy named ULTRA was proposed for regularizing ill-posed HU problems. Although ULTRA avoids the pixel interdependency, it requires a canonical polyadic decomposition at every algorithm iteration, which may negatively impact the complexity of the problem for large datasets.

In~\cite{Borsoi2017_multiscale} a multiscale spatial regularization approach was proposed for sparse unmixing. The method uses a signal-adaptive spatial multiscale decomposition to break the unmixing problem down into two simpler problems, one in an approximation domain and another in the original domain. The spatial contextual information is obtained by solving an unregularized unmixing problem in the approximation domain. This information is then mapped back to the original image domain and used to regularize the original unmixing problem. The multiscale approach resulted in a fast algorithm that outperformed competing methods, both in accuracy and in execution time, and promoted piecewise homogeneity in the estimated abundances without compromising sharp discontinuities among neighboring pixels.



Motivated by the excellent results in~\cite{Borsoi2017_multiscale}, we propose in this paper a novel data dependent multiscale mixture model for use in hyperspectral unmixing accounting for spectral variability of the endmembers. The new model uses a multiscale transform to incorporate spatial contextual information into the abundances of a generic mixing model considering spectral variability. The unmixing problem is then formulated as the minimization of a cost function in which a parametric endmember model (e.g. ELMM, PLMM or GLMM) is used to constraint and reduce the ill-posedness of the endmember estimation problem. However, the dimensionality of this problem is still very high since the spatial regularization ties the abundance solutions of all pixels together. Nevertheless, under a few mild assumptions we are able to devise a computationally efficient solution to the abundance estimation problem that can also be computed separately in the two domains.

The contributions of this paper include:
\begin{enumerate}
    \item The proposal of a new regularization strategy based on a multiscale representation of the hyperspectral images and abundance maps. This regularization is significantly different from and improves our previous work~\cite{Borsoi2017_multiscale}. 
    While in~\cite{Borsoi2017_multiscale} the static/fixed endmember matrix for all pixels allowed the easy separation of the abundance estimation process in different domains, the same approach is not applicable to the present case since the variability of the endmember matrix ties the abundances in the approximation and original image domains.
    \item A new approximate multiscale decomposition of the generic mixing model considering spectral variability. The new decomposition leads to a separable abundance estimation problem that allows a simple and efficient solution without significantly sacrificing accuracy. Moreover, the solution can be determined in  parallel for all image pixels.
\end{enumerate}

When compared with approaches that rely on standard spatial regularization strategies and on variable splitting techniques such as ADMM, the proposed strategy leads to a faster iterative algorithm that at each iteration solves the abundance problem only once in each domain. The new algorithm is named \textit{Multiscale Unmixing Algorithm Accounting for Spectral Variability} (MUA-SV). Simulation results clearly show the advantage of the proposed algorithm, both in accuracy and in execution time, over the competing methods.



In this paper we represent scalars by lowercase italic letters, vectors by lowercase bold letters and matrices by uppercase bold letters. We denote the number of pixels in an image by $N$, the number of bands by $L$, and the number of endmembers in the scene by $P$. We denote the hyperspectral image and the abundance maps by $\bY\in\mathbb{R}^{L\times N}$ and $\bA\in\mathbb{R}^{P\times N}$, respectively, whose $n$-th columns are given by $\by_n$ and $\ba_n$. The endmember matrix for each pixel is denoted by $\bM_n\in\mathbb{R}^{L\times P}$, and $\bM_0\in\mathbb{R}^{L\times P}$ represents a reference endmember matrix extracted from the image~$\bY$. Matrix $\bW\in\mathbb{R}^{N\times S}$ denotes the multiscale transformation and $\bW^*\in\mathbb{R}^{S\times N}$ its conjugate, and subscripts $\calC$ and $\calD$ in variables denote their representation in the coarse and detail spatial scales, respectively.

The paper is organized as follows. Section~\ref{sec:LMM} briefly reviews the linear mixing models and its variants accounting for spectral variability. In Section~\ref{sec:multi_scale}, we present the proposed multiscale formulation for the mixture model. In Section~\ref{sec:unmixing_prob_general} we formulate the unmixing problem using the multiscale approach. The optimization of the resulting cost function is presented in Section~\ref{sec:algorithm}. In Section~\ref{sec:probA_reformulation}, we propose an approximate formulation of the abundance estimation problem that leads to a simple and efficient solution. The resulting MUA-SV algorithm is detailed in Section~\ref{sec:mua_sv_alg}.
Simulation results with synthetic and real data are presented in Section~\ref{sec:results}.
Section~\ref{sec:conclusions} presents the conclusions.

\section{Linear mixing models considering spectral variability}\label{sec:LMM}

The classical Linear Mixing Model (LMM)~\cite{Keshava:2002p5667} assumes that an $L$-band pixel $\by_n = [y_{n,1},\,\ldots, \,y_{n,L} ]^\top$ is represented as
\begin{align} \label{eq:LMM}
 &\by_n = \bM \ba_n + \be_n, 
\quad \text{subject to }\,\cb{1}^\top\ba_n = 1 \text{ and } \ba_n \geq \cb{0} 
\end{align}
where $\bM\in\mathbb{R}^{L\times P} $ is the endmember matrix whose columns $\bm_i = [m_{i,1},\,\ldots,\,m_{i,L}]^\top$ are the spectral signatures of pure materials\footnote{Note that the definition of a pure material depends on convention and may change depending on the problem.},
$\ba_n = [a_{n,1},\,\ldots,\,a_{n,P}]^\top$ is the abundance vector, $\be_n\sim\mathcal{N}(0, \sigma_n^2\bI)$ is an additive white Gaussian noise (WGN), and $\bI$ is the identity matrix. 
The LMM assumes that the endmember spectra are fixed for all pixels $\by_n$, $n=1,\ldots,N$, in the hyperspectral image. This assumption can compromise the accuracy of estimated abundances in many circumstances due to the spectral variability existing in a typical scene. Recently, variations of the LMM have been proposed to cope with the variability phenomenon~\cite{thouvenin2016hyperspectralPLMM,drumetz2016blindUnmixingELMMvariability,imbiriba2018GLMM,borsoi2019deepGun,hong2019augmentedLMMvariability}. This work considers models that are linear on the abundances.

%

The most general form of the LMM considering spectral variability generalizes~\eqref{eq:LMM} to allow a different endmember matrix for each pixel. This results in the following observation model for the $n$-th pixel
\begin{align} \label{eq:model_variab_general}
	\by_n = \bM_n \ba_n + \be_n, \qquad n = 1, \dots, N
\end{align}
where $\bM_n\in\mathbb{R}^{L\times P}$ is the $n$-th pixel endmember matrix. This model can also be written for all pixels as
\begin{align} \label{eq:model_variab_general_allpx}
    \bY = \big[\bM_1\ba_1, \ldots, \bM_N\ba_N\big] + \bE\,,
\end{align}
where $\bY=[\by_1,\ldots,\by_N]$ is the matrix with all observed pixels and $\bE=[\be_1,\ldots,\be_N]$ is the noise.

Different models have been recently proposed to represent endmember variability as a parametric function of some reference endmember spectral signatures~\cite{thouvenin2016hyperspectralPLMM,drumetz2016blindUnmixingELMMvariability,imbiriba2018GLMM,borsoi2019deepGun}. These models are generically denoted by
\begin{align} \label{eq:param_mdl_generic}
    \bM_n = f(\bM_0,\btheta_n)
\end{align}
where~$f$ is a parametric function, $\bM_0\in\mathbb{R}^{L\times P}$ is a fixed reference endmember matrix and $\btheta_n$ is a vector of parameters used to describe the endmember signatures for the $n$-th pixel.
Although different forms have been proposed for~$f$, two notable examples are given by the Perturbed Linear Mixing Model (PLMM)~\cite{thouvenin2016hyperspectralPLMM} and the Extended Linear Mixing \mbox{Model (ELMM)~\cite{drumetz2016blindUnmixingELMMvariability}.}

The PLMM proposed in~\cite{thouvenin2016hyperspectralPLMM} models $\bM_n$ as a fixed matrix $\bM_0$ plus a pixel-dependent variation matrix that can accommodate generic spatial variations. Mathematically,
\begin{align} \label{eq:plmm_model_i}
	\by_n {}={} \big(\bM_0 + \bd\bM_n\big) \ba_n + \be_n \,,
\end{align}
where the parameters of this model are related to those of~\eqref{eq:param_mdl_generic} by $\btheta_n\equiv\vect(\bd\bM_n)$, where $\vect(\cdot)$ is the vectorization operator. This model is not physically motivated. Hence, in most cases all elements of $\bd\bM_n$ must be included as independent variables in the solution of the ill-posed unmixing problem, making the inverse problem very hard to solve. This limitation motivated the development of simpler, physically motivated variability models.

The ELMM is a simpler model proposed in~\cite{drumetz2016blindUnmixingELMMvariability}. It incorporates a multiplicative diagonal matrix to LMM, which maintains the directional information of the reference endmembers, but allows them to be independently scaled. The ELMM is expressed as
\begin{align} \label{eq:model_variab_elmm}
	\by_n = \bM_0 \,\diag (\bpsi_n) \ba_n + \be_n
    \,,
\end{align}
where $\bpsi_n\in\mathbb{R}^{P}$ is a vector containing a (positive) scaling factor for each endmember, which is related to the parameters of~\eqref{eq:param_mdl_generic} by $\btheta_n\equiv\bpsi_n$, and $\diag(\bx)$ denotes a diagonal matrix whose diagonal elements are given by the elements of vector~$\bx$. This model is a particular case of~\eqref{eq:model_variab_general} that can model typical endmember variations, such as those caused by illumination variability due to the topography of the scene~\cite{drumetz2016blindUnmixingELMMvariability}. 
%
%
The optimization problem that needs to be solved using model~\eqref{eq:model_variab_elmm} is much less ill-posed than that generated using model~\eqref{eq:plmm_model_i} due to the reduced number of parameters to be estimated.  This simplicity is obtained at the price of \mbox{reduced generality.}

For both the PLMM and ELMM, the problem of estimating the fractional abundances and the spectral signatures of the endmembers was cast as a large scale, non-convex inverse problem, which was solved using variable splitting procedures~\cite{thouvenin2016hyperspectralPLMM,drumetz2016blindUnmixingELMMvariability}.
The computational cost of these solutions is very high, making them unsuited for processing large amounts of data. Furthermore, the introduction of \textit{a priori} information about the spatial regularity of the abundance maps, which is essential to reduce the ill-posedness of the inverse problem, results in a optimization problem that is not separable per pixel. This significantly increases the computational cost of the solution. Considering this limitation of the models described above, it is of interest to develop new mixture models that combine the generality of the endmember variability patterns that can be considered with the possibility of an efficient solution of the associated inverse problem. In the next section, we introduce a new mixture model that represents separately the image components at different scales using a data-dependent transformation learned from the observed hyperspectral image~$\bY$. 
This new multiscale representation can be employed to solve the unmixing problem with any parametric model to represent spectral variability that fits the form~\eqref{eq:param_mdl_generic}.
 The use of this new model results in a method that is able to provide more accurate solutions at a much lower computational cost than the existing methods.

\section{A Multiscale Spatial Mixture Model} \label{sec:multi_scale}

%


To constrain the set of possible solutions, we propose to separately represent the mixture process in two distinct image scales, namely, the coarse scale containing rough spatial structures, and the fine spatial scale containing small structures and details. By doing so, the conditions for spatial smoothness can be imposed on the relevant parameters in the much simpler coarse scale, and then be translated into the fine scale for further processing.

Denote the fractional abundance maps for all pixels by $\bA=[\ba_1,\ldots,\ba_N]$. We consider a transformation $\bW\in\mathbb{R}^{N\times S}$ based on relevant contextual inter-pixel information present in the observed image $\bY$ to be applied to both $\bY$ and $\bA$ to unveil the coarse image structures. The transformed matrices are given by
\begin{align} \label{eq:decomposition_calC_i}
	\bY_{\!\calC} = \bY\bW \,; 
    \quad \bA_{\!\calC} = \bA \bW \,,
\end{align}
where~$\bY_{\!\calC}=[\by_{\calC_1},\ldots,\by_{\calC_S}] \in \mathbb{R}^{L\times S}$ and~$\bA_{\!\calC}=[\ba_{\calC_1},\ldots,\ba_{\calC_S}]  \in \mathbb{R}^{P\times S}$ with $S \ll N$ are, respectively, the hyperspectral image and the abundance matrix in the coarse approximation scale, denoted by~$\calC$. Being signal dependent, the transformation $\bW$ is nonlinear, and does not bear a direct relationship with the frequency components of the image, though some general relationship exists.

The spatial details of the image are represented in the detail scale, denoted by~$\calD$, which is obtained by computing the complement to the transformation $\bW$. Mathematically,
\begin{align} \label{eq:decomposition_calD_i}
	\bY_{\!\calD} = \bY (\bI-\bW\bW^\ast) \,;
    \quad \bA_{\!\calD} = \bA (\bI-\bW\bW^\ast)
    \,,
\end{align}
where~$\bY_{\!\calD}=[\by_{\calD_1},\ldots,\by_{\calD_N}]\in \mathbb{R}^{L\times N}$ and~$\bA_{\!\calD}=[\ba_{\calD_1},\ldots,\ba_{\calD_N}] \in \mathbb{R}^{P\times N}$ are the input image and the abundance matrix in the detail scale. Matrix $\bW^\ast\in\mathbb{R}^{S\times N}$ is a conjugate transformation to $\bW$, and takes the images from the coarse domain $\calC$ back to the original image domain. $\bY_{\!\calD}$ and $\bA_{\!\calD}$ contain the fine scale details of~$\bY$ and~$\bA$ in the original image domain. The transformation $\bW$ captures the spatial correlation of the input image, whereas its complement $(\bI-\bW\bW^\ast)$ captures existing fine spatial variabilities. This way it is possible to introduce spatial correlation into the abundance map solutions by separately controlling the regularization strength in each of the scales~$\calC$ and~$\calD$. This is computationally much simpler than to use more complex penalties. By imposing a smaller penalty in the coarse scale~$\calC$ and a larger penalty in the details scale~$\calD$, we effectively favor smooth solutions to the optimization problem.

We can define a composite transformation as
\begin{align} \label{eq:TransfcalW}
	\calW = [\bW \quad \bI-\bW\bW^\ast] \,,
\end{align}
which decomposes the input image into the coarse approximation $\calC$ and its complement $\calD$. Note that the transformation is invertible, with a right inverse given by
\begin{align} \label{eq:transf_pinverse}
	\calW^\dagger = [\bW^\ast \quad \bI ]^\top
    \,.
\end{align}

Multiplying $\bY$ from the right by~$\calW$ and considering the generic mixing model for all pixels given in~\eqref{eq:model_variab_general_allpx} yields $\bY \calW = \big[ \bY_{\!\calC} \,\,\bY_{\!\calD} \big]$, with
\begin{align} \label{eq:model_decomposed_i}
\begin{split}
	\bY_{\!\calC} & = \big[\bM_1\ba_1 \ldots \bM_N\ba_N \big] \bW
    + \bE_{\!\calC}
    \\
    \bY_{\!\calD} & = \big[\bM_1\ba_1 \ldots \bM_N\ba_N \big] 
    (\bI-\bW\bW^\ast) + \bE_{\!\calD}
\end{split}
\end{align}
where $\bE_{\!\calC}=\bE\bW$ and $\bE_{\!\calD}=\bE(\bI-\bW\bW^\ast)$ represent the additive noise in the coarse and detail scales, respectively.

\begin{figure}[bth]
\centering
\begin{minipage}[t]{.35\linewidth}
  \centering
  \centerline{\includegraphics[width=2.5cm,trim={0 5cm 0 0},clip]{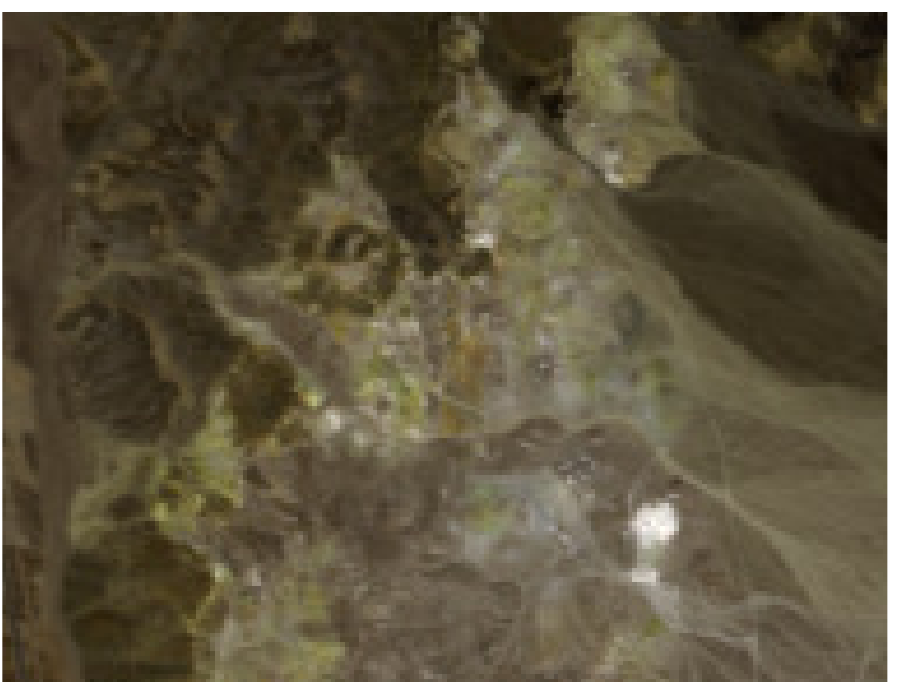}}
  \centerline{\small Image (bands 50, 80 and 100)}
\end{minipage}
\hspace{0.5cm}
\begin{minipage}[t]{0.35\linewidth}
  \centering
  \centerline{\includegraphics[width=2.5cm,trim={0 5cm 0 0},clip]{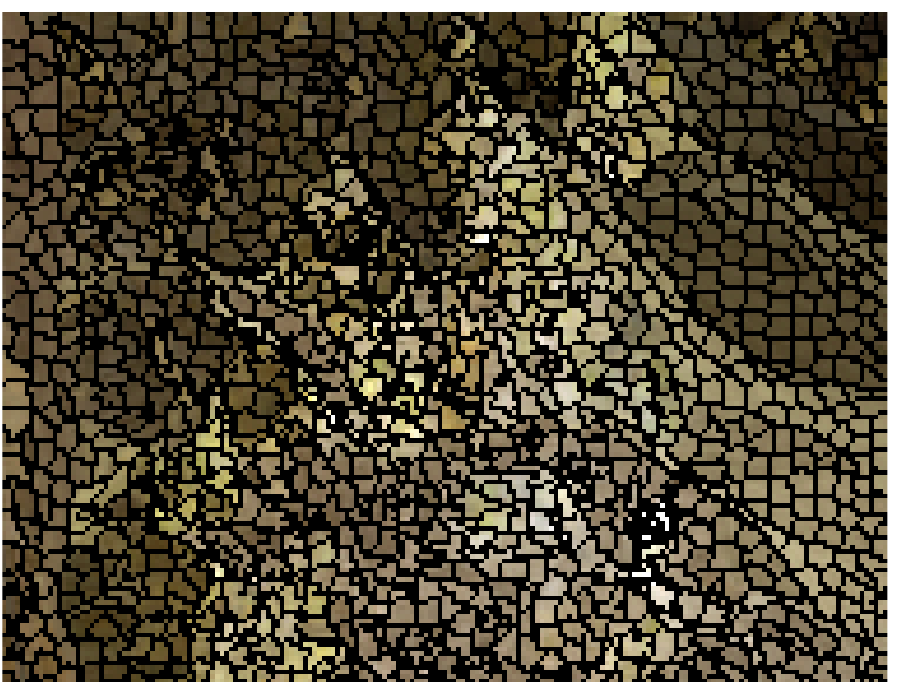}}
  \centerline{\small Superpixels}
\end{minipage}
\caption{Superpixel decomposition of a section of the Cuprite image using the algorithm in~\cite{achanta2012slicPAMI} with $S=5$ and $\gamma=0.005$.}
\label{eq:superpx_illustrative_ex}
\end{figure}


The choice of the multiscale transformation $\bW$ is important for the proposed methodology to achieve a good reconstruction accuracy. Desirable features for this transform are 1) to group image pixels that are spectrally similar and belong to spatially homogeneous regions, and 2) to respect image borders by not grouping pixels that belong to different image structures or features. Additionally, it must be fast to compute.

In~\cite{Borsoi2017_multiscale}, a superpixel decomposition of the image was considered for the transformation~$\bW$. Superpixels satisfy the aforementioned criteria, and have recently been widely applied to hyperspectral imaging tasks, including classification~\cite{jia2017superpixelHIclassification}, segmentation~\cite{saranathan2016superpixelsHIsegmentation}, endmember detection~\cite{thompson2010superpixelEMdetection}, and multiscale regularization~\cite{Borsoi2017_multiscale}.
Superpixel algorithms group image pixels into regions with contextually similar spatial information~\cite{achanta2012slicPAMI}, decomposing the image into a set of contiguous homogeneous regions. The sizes and regularity of the regions are controlled by adjusting a set of parameters. A particularly fast and efficient algorithm is the Simple Linear Iterative Clustering (SLIC) algorithm~\cite{achanta2012slicPAMI}, also considered in~\cite{Borsoi2017_multiscale}. The SLIC algorithm is an adaptation of the k-means algorithm that considers a reduced search space to lower the computational complexity, and a properly defined metric that balances spectral and spatial contributions. The superpixels (clusters) are initialized almost uniformly at low-gradient image neighborhoods to reduce the influence of noise. The number of clusters $S$ is determined as a function of the average superpixel size defined by the user. The clustering employs a normalized distance function that considers both spatial and spectral (color) similarities among  pixels. The relative contributions of spatial and spectral components are controlled by a regularity parameter $\gamma$. 
The parameter $\gamma$ can be increased to emphasize the spatial distance and obtain more compact (lower area to perimeter ratio) superpixels, or decreased to emphasize spectral distances and yield a tighter adherence to image borders (with more irregular shapes).
See the supplemental material in~\cite{Borsoi_multiscaleVar_2018} for more details.

The decomposition $\bY\bW$ of the image $\bY$ returns a set of superpixels. The value of each superpixel is equal to the average of all original pixel values inside that superpixel region.
The conjugate transform, $\bY_{\!\mathcal{C}}\bW^{\ast}$, takes each superpixel in $\bY_{\!\mathcal{C}}$ and attributes its value to all pixels of the uniform image sampling grid that lie inside its corresponding superpixel region.
The successive application of both transforms, $\bW\bW^{\ast}$ effectively consists in averaging all pixels inside each superpixel of the input image. 
The superpixel decomposition of the Cuprite image using the SLIC algorithm is illustrated \mbox{in Fig.~\ref{eq:superpx_illustrative_ex}.}

\section{The unmixing problem} \label{sec:unmixing_prob_general}

The spectral unmixing problem with spectral variability can be formulated as the minimization of the cost function
\begin{align} \label{eq:opt_prob_i}
\mathcal{J}(\mathbb{M},& \bTheta,\bA)  {}={}
    \frac{1}{2}
    \big\| \bY - \big[\bM_1\ba_1 \ldots \bM_N\ba_N ] \big\|_F^2  
    \nonumber\\
    & + \lambda_A {\mathcal{R}}(\bA) 
    + \frac{\lambda_M}{2} \sum_{n=1}^N \|\bM_n-f(\bM_0,\btheta_n)\|_F^2
    \nonumber\\
    & + \lambda_{\bTheta} \mathcal{R}(\bTheta) 
    \\ \nonumber
    & \text{subject to } \bA\geq\cb{0}, \,
    \cb{1}^\top \bA = \cb{1}^\top, 
    \nonumber \\ \nonumber 
    & \hspace{10ex}  \bM_n\geq\cb{0},\,\, n=1,\ldots,N.
\end{align}
where $\mathbb{M}$ is an ${L\times P\times N}$ tensor containing the endmember matrices, with entries given by $[\mathbb{M}]_{:,:,n}=\bM_n$ and $\bTheta=[\btheta_1,\ldots,\btheta_N]$ is a matrix containing the parameter vectors of the variability model for all pixels. Note that the generic parametric endmember model $\bM_n=f(\bM_0,\btheta_n)$ of~\eqref{eq:param_mdl_generic} was included in the cost function~\eqref{eq:opt_prob_i} in the form of an additive constraint. This decouples the problem of estimating the abundances from that of estimating the parametric endmember model, allowing the application of the multiscale formulation to other endmember models without loss of generality. Furthermore, this also gives more flexibility to the unmixing solution since the parameter $\lambda_M$ can be adjusted to either allow matrices $\bM_n$ to vary more freely or to strictly enforce the endmember variability model~\eqref{eq:param_mdl_generic}.

The regularization functionals $\mathcal{R}(\bA)$ and $\mathcal{R}(\bTheta)$ incorporate prior information about the spatial smoothness of the abundance and about the parameters of the spectral variability model.
The abundance maps constraint introduces spatial regularity indirectly through the transformation $\bW$. The constraint is given by
\begin{align}
	{\mathcal{R}}(\bA) & {}={}
    \frac{\rho}{2}\|\bA\bW\|^2_F + \frac{1}{2} \|\bA(\bI-\bW\bW^\ast)\|_F^2
    \nonumber\\
    & {}={} \frac{\rho}{2}\|\bA_{\calC}\|^2_F 
    + \frac{1}{2} \|\bA_{\calD}\|^2_F 
    \,
\end{align}
and consists of a quadratic penalization of the multiscale representation of the abundance maps, applied separately to the coarse and detail scales~$\calC$ and~$\calD$. Parameter $\rho$ allows the control of the relative weights of each scale in the abundance penalty. For instance, piecewise smooth abundance solutions to the optimization problem can be promoted by imposing a smaller penalty in the coarse scale~$\calC$ and a larger penalty in the details scale~$\calD$.

The constraint $\mathcal{R}(\bTheta)$ is selected according to the endmember variability model that is used, and might encode information such as the amount of spectral variability in a scene or spatial correlation in the variables $\btheta_n$. The parameters $\lambda_{A}$ and $\lambda_{\bTheta}$ control the balance between the different terms in the cost function.

In the following, we employ the ELMM model due to its parsimony and underlying physical motivation~\cite{drumetz2016blindUnmixingELMMvariability}. This results in the following concrete forms for $f$ and $\bTheta$:
\begin{align} \label{eq:var_mdl_elmm_sel}
\begin{split}
    f(\bM_0,\btheta_n) & {}\equiv{} \bM_0 \,\diag(\bpsi_n )
    \\
    \bTheta & {}\equiv{} \bPsi
\end{split}
\end{align}
where $\bPsi=[\bpsi_1,\ldots,\bpsi_N]$ is a matrix whose $n$-th column contains scaling factors~$\bpsi_n$ of the ELMM model~\eqref{eq:model_variab_elmm}. The scaling maps constraint $\mathcal{R}(\bTheta)$ is selected to introduce spatial smoothness to the endmember scaling factors, and is given by
\begin{align} \label{eq:var_reg_elmm_sel}
    \mathcal{R}(\bTheta) & {}\equiv{} \mathcal{R}(\bPsi)
    \nonumber \\
    & {}={} \|\mathcal{H}_h(\bPsi)\|_{F}^2 +  \|\mathcal{H}_v(\bPsi)\|_{F}^2 
    \,,
\end{align}
where $\mathcal{H}_h$ and $\mathcal{H}_v$ are linear operators that compute the vertical and horizontal gradients of a bi-dimensional signal, acting separately for each material. In the following, we make the variable substitutions outlined in~\eqref{eq:var_mdl_elmm_sel} and~\eqref{eq:var_reg_elmm_sel}, which turns the cost function in~\eqref{eq:opt_prob_i} into $\mathcal{J}(\mathbb{M},\bPsi,\bA)$.

The estimated abundance maps, endmember matrices and scaling factors can be obtained by minimizing~\eqref{eq:opt_prob_i} with respect to (w.r.t.) these variables, resulting in the following optimization problem
\begin{align} \label{eq:opt_prob_i_b}
    \widehat{\mathbb{M}}, \widehat{\bPsi}, \widehat{\!\bA} =  \mathop{\arg\min}_{\mathbb{M},\bPsi,\bA} 
    \,\,\,\mathcal{J}(\mathbb{M},\bPsi,\bA)
    \,.
\end{align}
This problem is non-convex and hard to solve directly due to the interdependence between $\mathbb{M}$, $\bPsi$ and $\bA$. Nevertheless, a local stationary point can be found using an Alternating Least Squares (ALS) strategy, which minimizes~\eqref{eq:opt_prob_i} successively with respect to one variable at a time~\cite{xu2013blockCoordinateDescent}.

The ALS approach allows us to break~\eqref{eq:opt_prob_i_b} into three simpler problems which are solved sequentially, consisting of:
\begin{align} \label{eq:als_sketch}
\hspace{-0.1cm}
\begin{split}
    & \!a) \, \text{minimize } \mathcal{J}(\mathbb{M}|\bA,\bPsi) \text{ w.r.t. } \mathbb{M} \text{ with } \bA \text{ and } \bPsi \text{ fixed}
    \\
    & \!b) \, \text{minimize } \mathcal{J}(\bPsi|\bA,\mathbb{M}) \text{ w.r.t. } \bPsi \text{ with } \bA \text{ and } \mathbb{M} \text{ fixed}
    \\
    & \!c) \, \text{minimize } \mathcal{J}(\!\bA|\,\mathbb{M},\bPsi) \text{ w.r.t. } \bA \text{ with } \mathbb{M} \text{ and } \bPsi \text{ fixed}
\end{split}
\end{align}
where $\mathcal{J}(\bB_1|\bB_2,\bB_3)$ denotes a cost function $\mathcal{J}$ in which $\bB_1$ is considered a variable and $\bB_2$, $\bB_3$ are fixed and \mbox{thus constants.}

Although this strategy yields a local minimum of the non-convex problem~\eqref{eq:opt_prob_i_b} by solving a sequence of convex optimization problems, it is still computationally intensive, specially due to the abundance estimation problem. This is because the spatial regularization term ${\mathcal{R}}(\bA)$ in~\eqref{eq:opt_prob_i} imposes interdependency among the different pixels of $\bA$, what also happens when the TV regularization is employed~\cite{drumetz2016blindUnmixingELMMvariability,shi2014surveySpatialRegUnmixing}.

Each of the optimization subproblems of the ALS strategy in~\eqref{eq:als_sketch} will be treated in detail in the next section.
Furthermore, in Section~\ref{sec:probA_reformulation} we will present a multiscale formulation that eliminates the interdependency of the abundance estimation problem between the different image pixels, allowing the solution to be computed faster and in parallel.



\section{Formulation and solutions to the optimization problems in~\eqref{eq:als_sketch}} \label{sec:algorithm}

We now detail the solution to each of the optimization problems in the ALS strategy outlined in~\eqref{eq:als_sketch}.
Although the solutions to the minimization problems w.r.t. $\mathbb{M}$ and $\bPsi$ are relatively straightforward and directly amenable to paralel or efficient implementations, optimizing~\eqref{eq:opt_prob_i} w.r.t. $\bA$ proves to be significantly more challenging due to the multiscale spatial regularization term~$\mathcal{R}(\bA)$.
Nevertheless, by making some approximations in Section~\ref{sec:probA_reformulation}, we will reformulate this optimization problem as a function of the multiscale representations $\bA_{\!\calC}$ and $\bA_{\!\calD}$ of the abundances. This will allow the extension of the ALS strategy to consider separate minimization steps w.r.t. $\bA_{\!\calC}$ and $\bA_{\!\calD}$, leading to a simple and parallelizable solution.
The complete algorithm including all optimization steps will be detailed in Section~\ref{sec:mua_sv_alg}.

\subsection{Optimizing with respect to $\mathbb{M}$ at the $i$-th iteration}

The cost function in this case is $\mathcal{J}(\mathbb{M}\,|\bA,\bPsi)$, where $\mathbb{M}$ is a variable and $\bA$ and $\bPsi$ are fixed at the solutions obtained in the previous iteration. Then, 
\begin{align} \label{eq:opt_subprob_M_i}
	& \mathcal{J}(\mathbb{M}\,|\bA,\bPsi)
	\nonumber \\ & 
	{}={} \frac{1}{2} \! \sum_{n=1}^N \Big(\|\by_n - \bM_n \ba_n\|^2_2 
    + \lambda_M \|\bM_n-\bM_0 \,\diag(\bpsi_n)\|_F^2 \Big)
    \nonumber \\
    & \hspace{5ex}\text{subject to } \bM_n\geq\cb{0}, \, n=1,\ldots, N
\end{align}
Similarly to~\cite{drumetz2016blindUnmixingELMMvariability}, we compute an approximate solution to minimize~\eqref{eq:opt_subprob_M_i} for each image pixel as
\begin{align} \label{eq:opt_subprob_M_ii}
	& \widehat{\!\bM}_n
	\\ & \hspace{1ex} \nonumber
	= \mathcal{P}_{\!+}\big(\big(\by_n\ba_n^\top + \lambda_M\bM_0\,\diag(\bpsi_n)\big)
    \big(\ba_n\ba_n^\top + \lambda_M \bI\big)^{-1}\big)
\end{align}
where $\mathcal{P}_{\!+}(\cdot)$ is an operator that projects each element of a matrix onto the nonnegative orthant by thresholding any negative element to zero.

\subsection{Optimizing with respect to $\bPsi$ at the $i$-th iteration}

The cost function in this case is $\mathcal{J}(\bPsi\,|\,\mathbb{M},\bA)$, where $\bPsi$ is a variable and $\bA$ and $\mathbb{M}$ are fixed at the solutions obtained in the previous iteration. Then,
\begin{align} \label{eq:opt_subprob_psi_i}
	\mathcal{J}(\bPsi | \mathbb{M},\!\bA) 
	\!=\! \frac{\lambda_M}{2} \! \sum_{n=1}^N \|\bM_n-\bM_0\bpsi_n\|_F^2
    + \lambda_{\bTheta} \mathcal{R}(\bPsi).
\end{align}
We follow the approach detailed in \cite[Eqs.~(20)-(23)]{drumetz2016blindUnmixingELMMvariability} to minimize~\eqref{eq:opt_subprob_psi_i}.

\subsection{Optimizing with respect to $\bA$ at the $i$-th iteration}\label{sec:A_opt}

The cost function in this case is $\mathcal{J}(\bA\,|\,\mathbb{M},\bPsi)$, where $\bA$ is a variable and $\bPsi$ and $\mathbb{M}$ are fixed at the solutions obtained in the previous iteration. Then,
\begin{align} \label{eq:opt_prob_a_minusOne}
    \mathcal{J}(\bA\,|\,\mathbb{M},\bPsi) & {}={}  
    \frac{1}{2} \big\| \bY - \big[\bM_1\ba_1 \ldots \bM_N\ba_N ] \big\|_F^2  
    \nonumber\\&
    + \frac{\rho\lambda_A}{2} \| \bA_{\!\calC} \|_F^2 + \frac{\lambda_A}{2} \| \bA_{\!\calD} \|_F^2 
    \\ \nonumber
    & \text{subject to } \bA\geq\cb{0}, \, \cb{1}^\top \bA = \cb{1}^\top
    \\ \nonumber 
    & \hspace{9.5ex} \bA_{\!\calC}=\bA\bW, \,  \bA_{\!\calD}=\bA(\bI-\bW\bW^*)
\end{align}
Using the multiscale transformation $\calW$ to write~\eqref{eq:opt_prob_a_minusOne} as a function of the observed hyperspectral images $\bY_{\!\calC}$ and $\bY_{\!\calD}$ represented at the coarse and detail scales yields
\begin{align} \label{eq:opt_prob_a_zero}
	\mathcal{J}(\bA &\,|\,\mathbb{M},\bPsi) {}={} \frac{1}{2} 
    \big\|\bY_{\!\calC}\bW^\ast - \big[\bM_1\ba_1 \ldots \bM_N\ba_N ] \bW\bW^\ast \big\|_F^2    
    \nonumber \\
    & + \frac{1}{2} \big\|\bY_{\!\calD} - \big[\bM_1\ba_1 \ldots \bM_N\ba_N ](\bI-\bW\bW^\ast)\big\|_F^2
    \nonumber\\
    & + \tr\Big\{ \big(\bY_{\!\calC}\bW^\ast - \big[\bM_1\ba_1 \ldots \bM_N\ba_N ] \bW\bW^\ast \big)^\top
    \nonumber \\ & \hspace{0.75cm} 
    \cdot \big( \bY_{\!\calD} - \big[\bM_1\ba_1 \ldots \bM_N\ba_N ](\bI-\bW\bW^\ast)\big)\Big\}
    \nonumber \\
    & + \frac{\rho\lambda_A}{2} \| \bA_{\!\calC} \|_F^2 
    + \frac{\lambda_A}{2} \| \bA_{\!\calD} \|_F^2 
    \\ \nonumber
    & \text{subject to } \bA\geq \cb{0}, \,\, \cb{1}^\top \bA = \cb{1}^\top
    \\ \nonumber 
    & \hspace{9.5ex} \bA_{\!\calC}=\bA\bW, \,  \bA_{\!\calD}=\bA(\bI-\bW\bW^*)
    \nonumber
\end{align}
where $\tr(\cdot)$ is the matrix trace operator.

Cost function~\eqref{eq:opt_prob_a_zero} is neither separable with respect to the abundance matrices $\bA_{\!\calC}$ and $\bA_{\!\calD}$ in the coarse and detail scales, nor with respect to the image pixels.
This can severely impact the required computational load and the convergence time to a meaningful result. To mitigate this issue, in the following section we propose to use few reasonable approximations to turn the minimization of~\eqref{eq:opt_prob_a_zero} into an optimization problem separable in $\bA_{\!\calC}$ and $\bA_{\!\calD}$. This will remove the interdependency between the different image pixels, and allow the extension of the ALS strategy to consider the optimization w.r.t. $\bA_{\!\calC}$ and $\bA_{\!\calD}$ successively, instead of w.r.t. $\bA$.


%

\section{Modification and solution to the optimization problem~w.r.t.~$\bA$} \label{sec:probA_reformulation}


Initially, we note that the cost function~\eqref{eq:opt_prob_a_zero} does not depends on the endmember variability model $f(\bM_0,\btheta_n)$. Hence, the derivations presented in this section are not limited to the ELMM, and can be equally applied to other models without loss of generality.

\subsection{Residuals inner product}\label{sec:residuals}
To proceed, we first denote by $RE_{\calC}$ and $RE_{\calD}$ the residuals/reconstruction errors in each image scale $\calC$ and $\calD$, where $RE_{\calC}$ and $RE_{\calD}$ are given by
\begin{align}
\begin{split}
	RE_{\calC} & {}={} \bY_{\!\calC}\bW^\ast - \big[\bM_1\ba_1 \ldots \bM_N\ba_N \big] \bW\bW^\ast
    \\
    RE_{\calD} & {}={} \bY_{\!\calD} - \big[\bM_1\ba_1 \ldots \bM_N\ba_N \big](\bI-\bW\bW^\ast)
    \,.
\end{split}
\end{align}
It follows from the above definition that the third term in the cost function~\eqref{eq:opt_prob_a_zero} consists of the inner product $\langle RE_{\calC},RE_{\calD}\rangle$ between the residuals/reconstruction errors at the coarse and detail scales. This inner product, however, usually contributes a small value to the cost function, and can be neglected under the following assumption:

\smallskip
\noindent\textsf{A1} - \textbf{Zero-mean, uncorrelated residuals:} We assume that for $\bA$ a critical point of~\eqref{eq:opt_prob_a_zero}, $RE_{\calC}$ and $RE_{\calD}$ are spatially zero-mean and uncorrelated across scales. This is reasonable if the observation/mixing model given by the ELMM in~\eqref{eq:model_variab_elmm} represents the data with reasonable accuracy, in which case the main contribution towards the residual error comes from the observation noise~$\be_n$, which is white and spatially uncorrelated.
If \textsf{A1} is satisfied, then the term $\langle RE_{\calC},RE_{\calD}\rangle$ can be neglected when compared to the first two terms without significantly altering the critical point.

\smallskip

Although neglecting the third term of~\eqref{eq:opt_prob_a_zero} simplifies the optimization problem, the first two terms still encompass intricate relationships between the abundances at different pixels due to the action of the multiscale transformation $\bW$.
Furthermore, the optimization problem still involves terms depending on both $\bA$ and the pair $(\bA_{\!\calC}$, $\bA_{\!\calD})$, which are related through $\calW$, and thus cannot be easily solved  in this form.

In order to proceed, we make the following assumption:

\smallskip
\noindent\textsf{A2} - \textbf{Spatially smooth endmember signatures:}  We assume that the pixel-by-pixel endmember signatures~$\bM_n$ are similar in small, compact spatial neighborhoods. More precisely, if~$\mathcal{N}$ is a set of pixels comprising a compact spatial neighborhood, we assume that the endmember signature of any pixel in~$\mathcal{N}$ does not deviate significantly from the average signature, so that the quantity
\[\bigg\| \bM_j - \frac{1}{|\mathcal{N}|} \sum_{n\in\mathcal{N}} \bM_n \bigg\|_F\]
is small for all $j\in\mathcal{N}$, where $|\mathcal{N}|$ is the cardinality of~$\mathcal{N}$.
%

\smallskip

We show in the following that this assumption leads to the separation of the optimization w.r.t. $\bA$ in~\eqref{eq:als_sketch} into two optimization steps, one w.r.t. $\bA_{\!\calC}$, and the other w.r.t. $\bA_{\!\calD}$. For numerical verification of the reasonability of \textsf{A1} and \textsf{A2}, see the supplemental material, also available in~\cite{Borsoi_multiscaleVar_2018}.

\subsection{Approximate Mixture Model}

Consider~\eqref{eq:opt_prob_a_zero} after neglecting its third term. Both $\bW$ (in the first term) and $\bI-\bW\bW^\ast$ (in the second term) act upon all the products $\bM_n\ba_n$, instead of just upon $\ba_n$, for $n = 1, \dots, N$. This precludes the separation of~\eqref{eq:opt_prob_a_zero} in a sum of non-negative functions exclusively dependent on $\bA_\calC$ or $\bA_\calD$, which could be independently minimized. However, combining \textsf{A2} and the fact that the transformation $\bW$ groups pixels that are in spatially adjacent regions, we now propose an approximate separable mixing model.

%


We initially express each pixel $\by_{\calC_i}$ and $\by_{\calD_i}$ of~\eqref{eq:model_decomposed_i} as
\begin{align} \label{eq:model_decomposed_ii_c}
	\by_{\calC_i} & = \sum_{j=1}^N  W_{j,i} \, \bM_j \ba_j + \be_{\calC_i}
\end{align}
and
\begin{align} \label{eq:model_decomposed_ii_d}
    \by_{\calD_i} & = 
    \bM_i\ba_i - \sum_{j=1}^S \sum_{\ell=1}^N  W^\ast_{j,i} \, W_{\ell,j} \, \bM_{\ell} \ba_{\ell}
    + \be_{\calD_i}
\end{align}
where $W_{j,i}$ and $W^\ast_{j,i}$ are the $(j,i)$-th elements of $\bW$ and $\bW^\ast$, respectively, and $\be_{\calC_i}$ and $\be_{\calD_i}$ denote the $i$-th columns of $\bE_{\calC}$ and $\bE_{\calD}$.
Then, using \textsf{A2} and the fact that $\bW$ is a localized decomposition, we approximate every endmember matrix $\bM_j$ in~\eqref{eq:model_decomposed_ii_c} by 
\begin{equation}
 \bM_j \approx \bM_{\!\calC_i} = \sum_{\ell=1}^N \frac{\mathbbm{1}_{W_{\ell,i}}}{|\emph{supp}_{\ell}(W_{\ell,i})|} \bM_{\ell}
\end{equation}
where $\mathbbm{1}_{W_{j,i}}$ is the indicator function of $W_{j,i}$ (i.e. $\mathbbm{1}_{W_{j,i}}=1$ if $W_{j,i}\neq0$ and $\mathbbm{1}_{W_{j,i}}=0$ otherwise), and $|\emph{supp}_{\ell}(f)|$ denotes the cardinality of the support of~$f$ as a function of~$\ell$.

Equivalently, we approximate every matrix $\bM_{\ell}$ in~\eqref{eq:model_decomposed_ii_d} by
\begin{equation}\label{eq:mci_star}
 \bM_{\ell} \approx \bM_{\!\calC_i^\ast} = \sum_{n=1}^S\sum_{m=1}^N \frac{\mathbbm{1}_{W_{n,i}^\ast} \mathbbm{1}_{W_{m,n}}}{|\emph{supp}_{n,m}(W_{n,i}^\ast W_{m,n})|} \bM_m
\end{equation}
where $|\emph{supp}_{m,n}(f)|$ denotes the cardinality of the support of~$f$ as a function of both $m$ and $n$. Thus,~\eqref{eq:model_decomposed_ii_c} and~\eqref{eq:model_decomposed_ii_d} can be approximated as (details in Appendix~\ref{app:model}) 
\begin{align} \label{eq:model_decomposed_iii_c}
	\by_{\calC_i} & \approx \bM_{\!\calC_i} \ba_{\!\calC_i} + \be_{\calC_i}
\end{align}
and
\begin{align} \label{eq:model_decomposed_iii_d}
    \by_{\calD_i} & \approx \bM_i\ba_{\!\calD_i} + \bM_{\!\calD_i} \big[\bA_{\calC} \bW^{\ast}\big]_i + \be_{\calD_i}
\end{align}
where $[\cdot]_i$ denotes the $i-$th column of a matrix, and $\bM_{\!\calD_i}  = \bM_i - \bM_{\!\calC_i^\ast}$ reflects the variability of $\bM_i$ with respect to $\bM_{\!\calC_i^\ast}$, the average endmember matrix of its neighborhood. According to \textsf{A2}, $\bM_{\!\calD_i}\approx \cb{0}$. Note that, since the transformation $\bW$ only groups together pixels that lie inside a single superpixel, we average $\ba_{n}$ and $\bM_n$ only in small spatial neighborhoods where their variability is small.

%

Selecting $\bW$ and $\bW^*$ according to the superpixels decomposition, we have that:
\begin{itemize}
\item $\bM_{\!\calC_i}$ is the average of all $\bM_j$ inside the $i$-th superpixel.
\item $\bM_{\!\calC_i^\ast}$ is the average of all $\bM_j$ inside the superpixel that contains the $i$-th pixel. Thus, if pixel $i$ belongs to the $k$-th superpixel, $\bM_{\!\calC_i^\ast}$ is the average of all $\bM_j$ inside the $k$-th superpixel.
\end{itemize}
Note that $\bW^\ast$ is also a localized transform, as it attributes the superpixel value to all pixels in the original  domain that lie inside that superpixel, which encompasses a compact spatial neighborhood.


Writting~\eqref{eq:model_decomposed_iii_c} and~\eqref{eq:model_decomposed_iii_d} for all pixels, we write \eqref{eq:model_decomposed_ii_c} and~\eqref{eq:model_decomposed_ii_d} in the matrix form as:
\begin{align} \label{eq:model_decomposed_iv}
\begin{split}
	\bY_{\!\calC} {}={} & \big[\bM_{\calC_1}\ba_{\calC_1},\ldots,\bM_{\calC_S}\ba_{\calC_S}\big] 
    + \widetilde{\!\bE}_{\calC}
    \\
    \bY_{\!\calD} {}={} & \big[\bM_{\!\calD_1} \big[\bA_{\!\calC} \bW^{\ast}\big]_1,\ldots,\bM_{\!\calD_N} \big[\bA_{\!\calC} \bW^{\ast}\big]_N\big]
    \\
    & + \big[\bM_1\ba_{\calD_1},\ldots,\bM_N\ba_{\calD_N}\big] 
    + \widetilde{\!\bE}_{\calD}
\end{split}
\end{align}
where $\widetilde{\!\bE}_{\calC}$ and $\widetilde{\!\bE}_{\calD}$ include additive noise and modeling errors.


\subsubsection{Abundance constraints}\label{sec:constraints}

The two constraints in~\eqref{eq:opt_prob_a_zero} are functions of $\bA$, and thus must be considered in the optimization with respect to $\bA_{\!\calC}$ and $\bA_{\!\calD}$.  Assuming $\calW$ in \eqref{eq:TransfcalW} to be of full row rank, the sum-to-one constraint can be expressed as 
\begin{align} \label{eq:sum_to_one_transf}
    &\mathbf{1}^\top\bA\calW = \mathbf{1}^\top\calW 
    \nonumber \\
    & \Longleftrightarrow 
    \mathbf{1}^\top\bA_{\!\calC} 
    = \mathbf{1}^\top\bW
    \,,\quad \mathbf{1}^\top\bA_{\!\calD} 
    = \mathbf{1}^\top (\bI-\bW\bW^\ast).
\end{align}


Considering the positivity constraint we have
\begin{align} \label{eq:abundances_pos_constr}
    & \bA\geq\cb{0} \Rightarrow \bA\calW\calW^\dagger \geq\cb{0} 
    \nonumber \\
    & \Longleftrightarrow
    [\bA_{\!\calC} \,\,\, \bA_{\!\calD}]\calW^\dagger \geq \cb{0}
    \Longleftrightarrow \bA_{\!\calC} \bW^\ast + \bA_{\!\calD} \geq \cb{0} \,.
\end{align}

If $\bW^\ast \ge \cb{0}$, which is true if $\bW$ is selected as the superpixel decomposition, we can further state that
\begin{align} \label{eq:SimplifiedConstraint}
	\bA_{\!\calC} \bW^\ast \geq\cb{0} \Longleftrightarrow \bA_{\!\calC}\geq\cb{0}
\end{align}
what simplifies the constraint by removing possible interdependencies between different pixels, and makes the problem separable for all pixels in the coarse scale $\calC$.

\subsubsection{The updated optimization problem}

Using the results obtained in Sections~\ref{sec:residuals} to \ref{sec:constraints}, minimizing~\eqref{eq:opt_prob_a_zero} with respect to $\bA$ can be restated as determining $\bA_{\!\calC}$ and $\bA_{\!\calD}$ that minimize
\begin{align} \label{eq:sec_opt_A_approx_gl}
	\widetilde{\mathcal{J}}(\bA_{\!\calC}&,\bA_{\!\calD}|\,\mathbb{M},\bPsi) 
	\nonumber \\ & 
	{}={} \frac{1}{2} 
    \Big\|\bY_{\!\calC}\bW^\ast - \big[\bM_{\calC_1}\ba_{\calC_1},\ldots,\bM_{\calC_S}\ba_{\calC_S}\big] \bW^\ast \Big\|_F^2 
    \nonumber \\
    & + \frac{1}{2} \Big\|\bY_{\!\calD} - \big[\bM_1\ba_{\calD_1},\ldots,\bM_N\ba_{\calD_N}\big]
    \nonumber \\
    & \qquad - \big[\bM_{\!\calD_1} \big[\bA_{\!\calC} \bW^{\ast}\big]_1,\ldots,\bM_{\!\calD_N} \big[\bA_{\!\calC} \bW^{\ast}\big]_N\big] \Big\|_F^2
    \nonumber \\
    & + \frac{\rho\lambda_A}{2} \| \bA_{\!\calC} \|_F^2 
    + \frac{\lambda_A}{2} \| \bA_{\!\calD} \|_F^2 
    \nonumber \\
    & \text{subject to } \bA_{\!\calC}\bW^\ast+\bA_{\!\calD}\geq \cb{0}, 
    \,\, \cb{1}^\top \bA_{\!\calC} = \cb{1}^\top\bW ,\,\,
    \nonumber \\
    & \hspace{10ex}  \cb{1}^\top\bA_{\!\calD} = \cb{1}^\top(\bI-\bW\bW^\ast).
\end{align}
Optimization problem~\eqref{eq:sec_opt_A_approx_gl} is amenable to an efficient solution, as detailed in the following section.

\subsection{Solution to the optimization problem~\eqref{eq:sec_opt_A_approx_gl}}

This section details the proposed solution of the optimization problem~\eqref{eq:sec_opt_A_approx_gl} w.r.t. $\bA_{\!\calC}$ and $\bA_{\!\calD}$.

\smallskip
\subsubsection{Optimizing with respect to $\bA_\calC$ at the $i$-th iteration} \label{sec:als_opt_ac}

The cost function in this case is $\widetilde{\mathcal{J}}(\bA_{\!\calC}|\bA_{\!\calD},\mathbb{M},\bPsi)$, where $\bA_{\!\calC}$ is a variable and $\bA_{\!\calD}$, $\mathbb{M}$ and $\bPsi$ are fixed at the solutions obtained in the previous iteration.
Note that this problem is still not separable with respect to each pixel in $\bA_{\!\calC}$ since the second term of~\eqref{eq:sec_opt_A_approx_gl} includes products between $\bA_{\!\calC}$ and $\bW^*$. However, this cost function can be simplified to yield a separable problem by making the following considerations using assumption \textsf{A2}:
\begin{enumerate}
 \item \textsf{A2} implies that the entries of $\bM_{\!\calD_i}$ are small when compared to those of $\bM_n$;
 \item \textsf{A2} also implies that the entries of $\bY_{\!\calC}\bW^\ast$ are usually much larger than the entries of $\bY_{\!\calD}$.
 %
\end{enumerate}
These considerations imply that the contribution of the terms $\bM_{\!\calD_i} \big[\bA_{\!\calC} \bW^{\ast}\big]$ in the second term of~\eqref{eq:sec_opt_A_approx_gl} can be neglected when compared to $\bY_{\!\calC}\bW^\ast$. Using this approximation and~\eqref{eq:SimplifiedConstraint}, the optimization with respect to $\bA_{\!\calC}$ can be stated as the minimization of
\begin{align}\label{eq:opt_AC_1}
	\overline{\mathcal{J}}(\bA_{\!\calC}&|\bA_{\!\calD},\mathbb{M},\bPsi) 
	\nonumber \\ 
	& {}={} \frac{1}{2} 
    \big\|\bY_{\!\calC}\bW^\ast - \big[\bM_{\calC_1}\ba_{\calC_1},\ldots,\bM_{\calC_S}\ba_{\calC_S}\big] \bW^\ast \big\|_F^2 
    \nonumber \\
    & + \frac{\rho\lambda_A}{2} \| \bA_{\!\calC} \|_F^2 
    \nonumber \\
    & \text{subject to } \bA_{\!\calC}\geq \cb{0}, 
    \,\, \cb{1}^\top \bA_{\!\calC} = \cb{1}^\top\bW 
\end{align}

For $\bW$ based on the superpixel decomposition, $\bW^\ast$ assigns to each pixel in the original image domain the value of the superpixel to which it belongs. 
Using this property, the cost function~\eqref{eq:opt_AC_1} simplifies to
\begin{align} \label{eq:SimplifiedACOptimization}
	\overline{\mathcal{J}} (\bA_{\!\calC}&|\bA_{\!\calD},\mathbb{M},\bPsi) 
	\\ & \!\!\! {}={}
    \frac{1}{2} \sum_{n=1}^S \Omega_s^2(n)  \Big(
    \|\by_{\calC_n} - \bM_{\!\calC_n}\ba_{\calC_n}\|_2^2
    + \frac{\widetilde{\rho}(n)\lambda_A}{2} \|\ba_{\calC_n}\|_2^2 \Big)
    \nonumber \\
    & \text{subject to} \,\,\, \ba_{\calC_n}\geq\cb{0},  \,
    \cb{1}^\top\ba_{\calC_n} = \cb{1}^\top [{\bW}]_n ,\,\, n=1,\ldots,S
    \nonumber 
\end{align}
where $[{\bW}]_n$ is the $n$-th column of $\bW$, $\Omega_s(n)$ is the number of pixels contained in the $n$-th superpixel and $\widetilde{\rho}(n)=\rho\Omega_s^{-2}(n)$, $n=1,\ldots,S$ is a superpixel-dependent regularization parameter that controls the balance between both terms in the cost function for each superpixel.

For simplicity, in the following we replace $\widetilde{\rho}(n)$ by a weighting term $\widetilde{\rho}_0=\rho S^2/N^2$ that is constant for all superpixels. This further simplifies the optimization problem since~$S$ is specified a priori by the user. Furthermore, since the optimization is independent for each pixel, we can also move the $\Omega_s^2(n)$ factor outside the summation in~\eqref{eq:SimplifiedACOptimization} without changing the critical point of the cost function.






Doing this results in the following cost function that can be minimized individually for each pixel:
\begin{align} \label{eq:simplified_cost_on_Ac}
    \widehat{\mathcal{J}} (\bA_{\!\calC}&|\bA_{\!\calD},\mathbb{M},\bPsi) 
	\\ & {}={}
    \frac{N^2}{2S^2} \sum_{n=1}^S  \Big( \|\by_{\calC_n} - \bM_{\!\calC_n}\ba_{\calC_n}\|_2^2
    + \frac{\widetilde{\rho}_0\lambda_A}{2} \|\ba_{\calC_n}\|_2^2 \Big)
    \nonumber \\
    & \text{subject to} \,\,\, \ba_{\calC_n}\geq\cb{0},  \,
    \cb{1}^\top\ba_{\calC_n} = \cb{1}^\top [{\bW}]_n ,\,\,n=1,\ldots,S.
    \nonumber
\end{align}
Note that~\eqref{eq:simplified_cost_on_Ac} is equivalent to a a standard FCLS problem, which can be solved efficiently.



\smallskip
\subsubsection{Optimizing with respect to $\bA_\calD$ at the $i$-th iteration} \label{sec:als_opt_ad}

The cost function in this case is $\widetilde{\mathcal{J}}(\bA_{\!\calD}|\bA_{\!\calC},\mathbb{M},\bPsi)$, where $\bA_{\!\calD}$ is a variable and $\bA_{\!\calC}$, $\mathbb{M}$ and $\bPsi$ are fixed at the solutions obtained in the previous iteration. Then, considering only the terms and constraints in~\eqref{eq:sec_opt_A_approx_gl} that depend on $\bA_{\!\calD}$ yields
\begin{align} \label{eq:opt_cf_cald_iii_prev}
	\widetilde{\mathcal{J}}(\bA_{\!\calD}&|\bA_{\!\calC},\mathbb{M},\bPsi) 
	\nonumber \\ & {}={} 
	\frac{1}{2} \Big\|\bY_{\!\calD} - \big[\bM_1\ba_{\calD_1},\ldots,\bM_N\ba_{\calD_N}\big]
    \nonumber \\
    & \qquad - \big[\bM_{\!\calD_1} \big[\bA_{\!\calC} \bW^{\ast}\big]_1,\ldots,\bM_{\!\calD_N} \big[\bA_{\!\calC} \bW^{\ast}\big]_N\big] \Big\|_F^2
    \nonumber \\
    & + \frac{\lambda_A}{2} \| \bA_{\!\calD} \|_F^2 
     \\
    & \text{subject to } \bA_{\!\calC}\bW^\ast+\bA_{\!\calD}\geq 0, 
    \nonumber \\
    & \hspace{10ex}  \cb{1}^\top\bA_{\!\calD} = \cb{1}^\top(\bI-\bW\bW^\ast). \nonumber
\end{align}

Since matrix $\bA_{\calC}$ is fixed, this problem can be decomposed for each pixel. This results in the minimization of the following cost function:
\begin{align} \label{eq:opt_cf_cald_iii}
    \widetilde{\mathcal{J}}(\bA_{\!\calD}&|\bA_{\!\calC},\mathbb{M},\bPsi)
    \nonumber \\ & {}={}
    \frac{1}{2} \sum_{n=1}^N \Big( \big\|\by_{\calD_n} - \bM_n\ba_{\calD_n} 
    - \bM_{\!\calD_n} \big[\bA_{\!\calC}\bW^\ast\big]_n \big\|_2^2 
    \nonumber \\
    & + \lambda_A \|\ba_{\calD_n}\|_2^2 \Big)
     \\
    & \text{subject to} \,\,\, \big[\bA_{\!\calC}\bW^{\ast}\big]_n + \ba_{\calD_n}\geq0 \nonumber
    \\
    & \hspace{10ex} \mathbf{1}^\top\ba_{\calD_n} = \mathbf{1}^\top \big[\bI-\bW\bW^\ast\big]_n \nonumber
    \\ 
    & \hspace{10ex} n=1,\ldots,N \nonumber 
\end{align}
where matrices $\bM_{\!\calD_n}$ are given in~\eqref{eq:model_decomposed_iii_c} and~\eqref{eq:model_decomposed_iii_d}. Note that this cost function is again equivalent to a standard FCLS problem, which can be solved efficiently.

\section{The MUA-SV unmixing algorithm} \label{sec:mua_sv_alg}

Considering the solutions to the optimization subproblems derived in the previous sections, the global unmixing procedure can be directly derived by setting the fixed variables of each subproblem with the estimates obtained from the previous iteration. The MUA-SV algorithm is presented in Algorithm~\ref{alg:global_opt}.


\begin{algorithm} [bth]
\small
\SetKwInOut{Input}{Input}
\SetKwInOut{Output}{Output}
\caption{Global MUA-SV algorithm~\label{alg:global_opt}}
\Input{Image $\bY$, parameters $\lambda_M$, $\lambda_A$, $\lambda_{\Psi}$, $\rho$ and matrices $\bA^{(0)}$, $\bPsi^{(0)}$ and $\bM_0$.}
\Output{Estimated matrices $\widehat{\!\bA}$, $\widehat{\bPsi}$ and tensor $\widehat{\mathbb{M}}$.}
Compute the superpixel decomposition of the hyperspectral image $\bY$ and the corresponding transformation matrices $\bW$, $\bW^\ast$, $\calW$ and $\calW^\dagger$ using the SLIC algorithm~\cite{achanta2012slicPAMI}\; 
Compute the decomposition of~$\bY$ into approximation and detail domains $\bY_{\!\mathcal{C}}$, and $\bY_{\!\mathcal{D}}$ using~\eqref{eq:decomposition_calC_i} and~\eqref{eq:decomposition_calD_i} \;
Set $\bA_{\!\calD}^{(0)}=\bA^{(0)}(\bI-\bW\bW^*)$ \;
Set $i=1$ \;
\While{stopping criterion is not satisfied}{
$\mathbb{M}^{(i)} = \underset{\mathbb{M}}{\arg\min} \,\,\,\,  \mathcal{J}(\mathbb{M}|\bA^{(i-1)},\bPsi^{(i-1)})$ \;
$\bA_{\!\calC}^{(i)} = \underset{\bA_{\!\calC}}{\arg\min} \,\,\,\, \widehat{\mathcal{J}}(\bA_{\!\calC}|\bA_{\!\calD}^{(i-1)},\mathbb{M}^{(i)},\bPsi^{(i-1)})$ \;
$\bA_{\!\calD}^{(i)} = \underset{\bA_{\!\calD}}{\arg\min} \,\,\,\,  \widetilde{\mathcal{J}}(\bA_{\!\calD}|\bA_{\!\calC}^{(i)},\mathbb{M}^{(i)},\bPsi^{(i-1)})$ \;
$\bA^{(i)}=\big[\bA_{\!\calC}^{(i)}\,\,\bA_{\!\calD}^{(i)}\big]\calW^\dagger$ \;
$\bPsi^{(i)} = \underset{\bPsi}{\arg\min} \,\,\,\, \mathcal{J}(\bPsi|\,\mathbb{M}^{(i)},\bA^{(i)})$ \;
$i=i+1$ \;
}
\KwRet $\widehat{\!\bA}=\bA^{(i-1)}$,~ $\widehat{\mathbb{M}}=\mathbb{M}^{(i-1)}$,~ $\widehat{\bPsi}=\bPsi^{(i-1)}$  \;
\end{algorithm}

\section{Results} \label{sec:results}

In this section, we compare the unmixing performances achieved using the proposed MUA-SV algorithm, the Fully Constrained Least Squares (FCLS), the Scaled Constrained Least Squares (SCLS), the PLMM-based solution~\cite{thouvenin2016hyperspectralPLMM} and the ELMM-based solution~\cite{drumetz2016blindUnmixingELMMvariability}, the latter two designed to tackle spectral variability.
The SCLS algorithm is a particular case of the ELMM model that employs the same scaling factors $\bpsi_n$ for all endmembers in each pixel (i.e. $\bM_n=\psi_n\bM_0$, where $\psi_n\in\mathbb{R}_+$)~\cite{Nascimento2005doesICAplaysRole}. It is a low complexity algorithm that can be used as a baseline method to account for spectral variability.

For all simulations, the reference endmember signatures~$\bM_0$ were extracted from the observed image using the Vertex Component Analysis (VCA) algorithm~\cite{nascimento2005vca}.
The abundance maps were initialized with the SCLS result for all algorithms. The scaling factors $\bPsi$ for ELMM and MUA-SV were initialized with ones. The matrix $\bM$ for the PLMM was initialized with the results from the VCA.
The alternating least squares loop in Algorithm~\ref{alg:global_opt} is terminated when the norm of the relative variation of the three variables between two successive iterations is smaller than $\epsilon_A=\epsilon_{\Psi}=\epsilon_M=2\times10^{-3}$.


Experiments were performed for three synthetic and two real data sets. For the synthetic data, the regularization parameters were selected for each algorithm to provide the best abundance estimation performance.
The complete set of parameters, comprising the SLIC ($S$ and $\gamma$) and the regularization parameters ($\rho$, $\lambda_M$, $\lambda_A$, and $\lambda_\psi$), were searched in appropriate intervals. For instance, $\gamma\in\{0.001,\, 0.0025,\, 0.005,\, 0.01,\, 0.025,\, 0.05\}$, $S$ assumed an integer value in the interval $[2,9]$, $\rho$ was selected so that $\rho S^2/N^2 \in\{0.001,\, 0.01,\, 0.025,\, 0.05,\, 0.1,\, 0.15,\, \allowbreak 0.2,\, 0.25,\, 0.35,\, 0.5\}$, while $\lambda_M$, $\lambda_A$, and $\lambda_\psi$ were searched in the range $[5\times10^{-4},100]$, with 12 points sampled uniformly.

The algorithms were implemented on a desktop computer equipped with an Intel~I7~4.2~Ghz processor with~4 cores and 16~Gb RAM. ELMM, PLMM and SLIC were implemented using the codes made available by the respective authors.
We did not employ parallelism when implementing the MUA-SV algorithm, so as to reduce the influence of the hardware platform when evaluating the performance gains achieved through the proposed simplifications. If parallelism is employed, the execution times can be even smaller.

\begin{figure*}
    \centering
    \begin{subfigure}[b]{0.3\textwidth}
        \includegraphics[width=5.5cm]{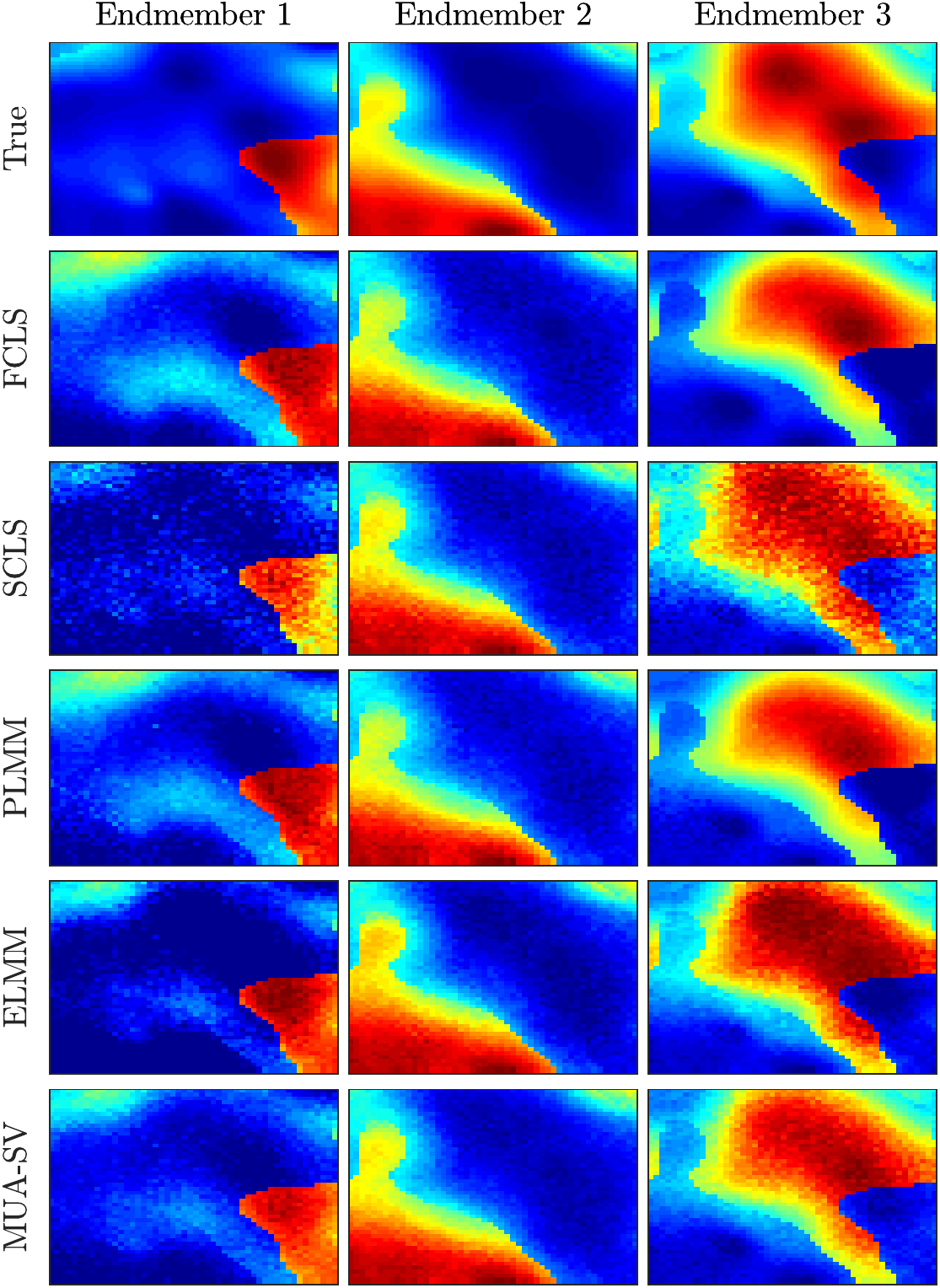}
        \caption{DC1.}
        \label{fig:abundances_DC1}
    \end{subfigure}
    ~~
    \begin{subfigure}[b]{0.3\textwidth}
        \includegraphics[width=5.5cm]{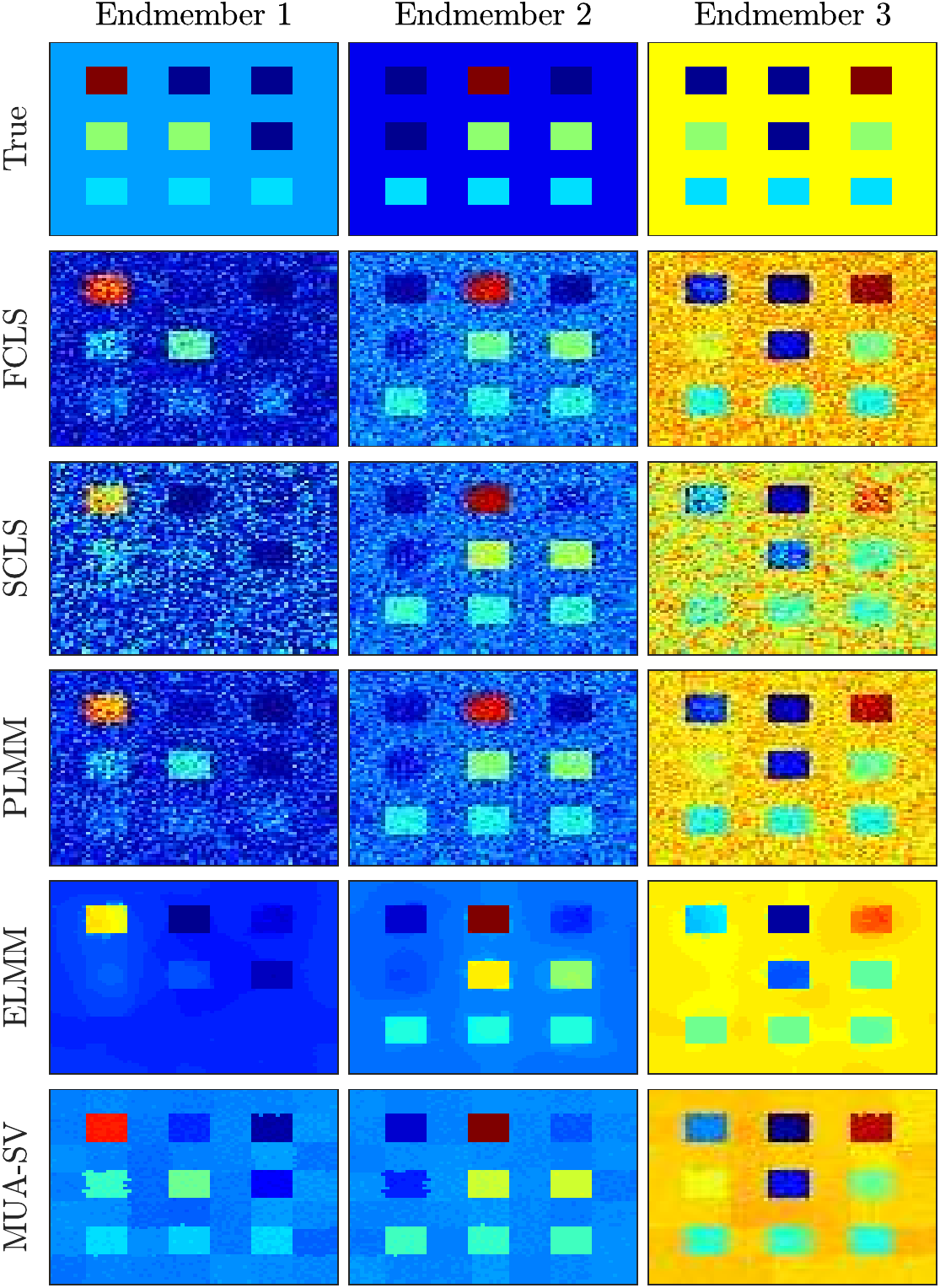}
        \caption{DC2.}
        \label{fig:abundances_DC2}
    \end{subfigure}
    ~~
    \begin{subfigure}[b]{0.3\textwidth}
        \includegraphics[width=5.5cm]{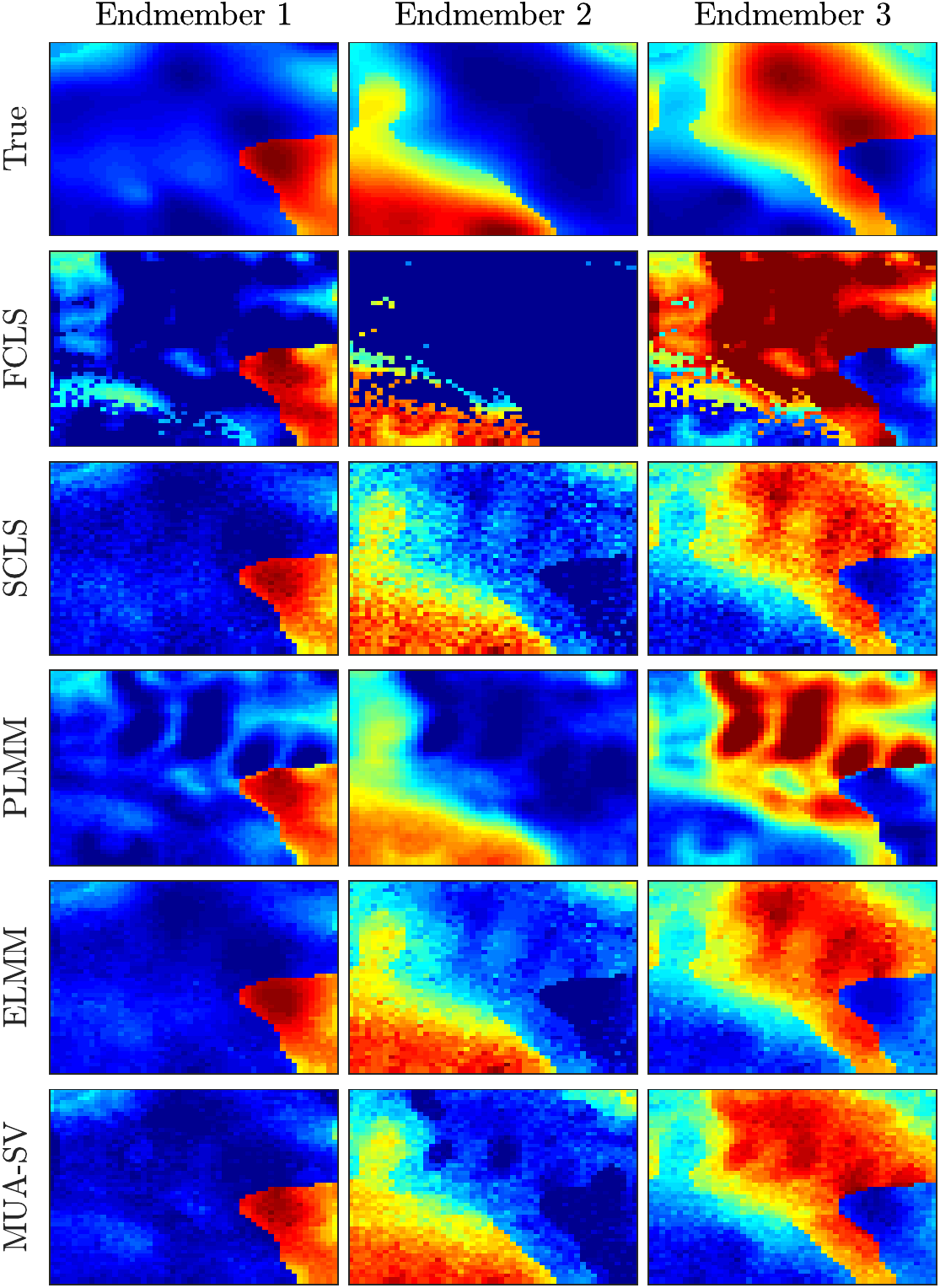}
        \caption{DC3.}
        \label{fig:abundances_DC3}
    \end{subfigure}
    \caption{True and reconstructed abundance maps for the synthetic data cubes for SNR=30~dB.}
    \label{fig:animals}
\end{figure*}

\subsection{Synthetic data sets}

Three synthetic data sets were built. The first data cube (DC1) was built from the ELMM model to verify how MUA-SV performs when the actual endmembers closely follows the adopted model.
The second data cube (DC2) was built using the more challenging additive perturbation model of~\cite{thouvenin2016hyperspectralPLMM}. 
The third data cube (DC3) was based on a realistic simulation of endmember variability caused by illumination conditions following the Hapke's model~\cite{Hapke1981}.

The data cube DC1 contains $50\times50$ pixels and three materials selected randomly from the USGS library and used as the reference endmember matrix~$\bM_0$, with~224 spectral bands.
The abundance maps are piecewise smooth images generated by sampling from a Gaussian Random field
\footnote{Generated using the code in http://www.ehu.es/ccwintco/index.php/ Hyperspectral\_Imagery\_Synthesis\_tools\_for\_MATLAB}~\cite{kozintsev1999computationsGaussianFields}, and are depicted in Fig.~\ref{fig:abundances_DC1}.
Spectral variability was added to the reference endmembers using the same model as in~\cite{drumetz2016blindUnmixingELMMvariability}, where the endmember instances for each pixel were generated by applying a constant scaling factor to the reference endmembers with amplitude limited to the interval $[0.75,1.25]$. Finally, a white Gaussian noise with a 25~dB SNR was added to the already scaled endmembers.
The true scaling factors applied to each endmember were generated using a Gaussian Random field, and thus exhibit spatial correlation.

The data cube DC2 contains $70\times70$ pixels and three materials, also randomly selected from the USGS spectral library to compose matrix~$\bM_0$ with~224 spectral bands.
The abundance maps (shown in Fig.~\ref{fig:abundances_DC2}) are composed by square regions distributed uniformly over a background, containing pure pixels (first row) and mixtures of two and three endmembers (second and third rows). The background pixels are mixtures of the same three endmembers, with abundances~$0.2744$,~$0.1055$ and~$0.62$. 
Spectral variability was added following the model proposed in~\cite{thouvenin2016hyperspectralPLMM}, which considered a per-pixel variability given by random piecewise linear functions to scale individually the spectrum of each endmember by a factor in the interval $[0.8,1.2]$.
Such a variability model does not match the ELMM, as it yields different variabilities across  the spectral bands, and is not designed to produce spatial correlation. Nevertheless, it provides a good ground for comparison with more flexible models such as the PLMM.

\begin{figure}
	\centering
    \includegraphics[width=5cm]{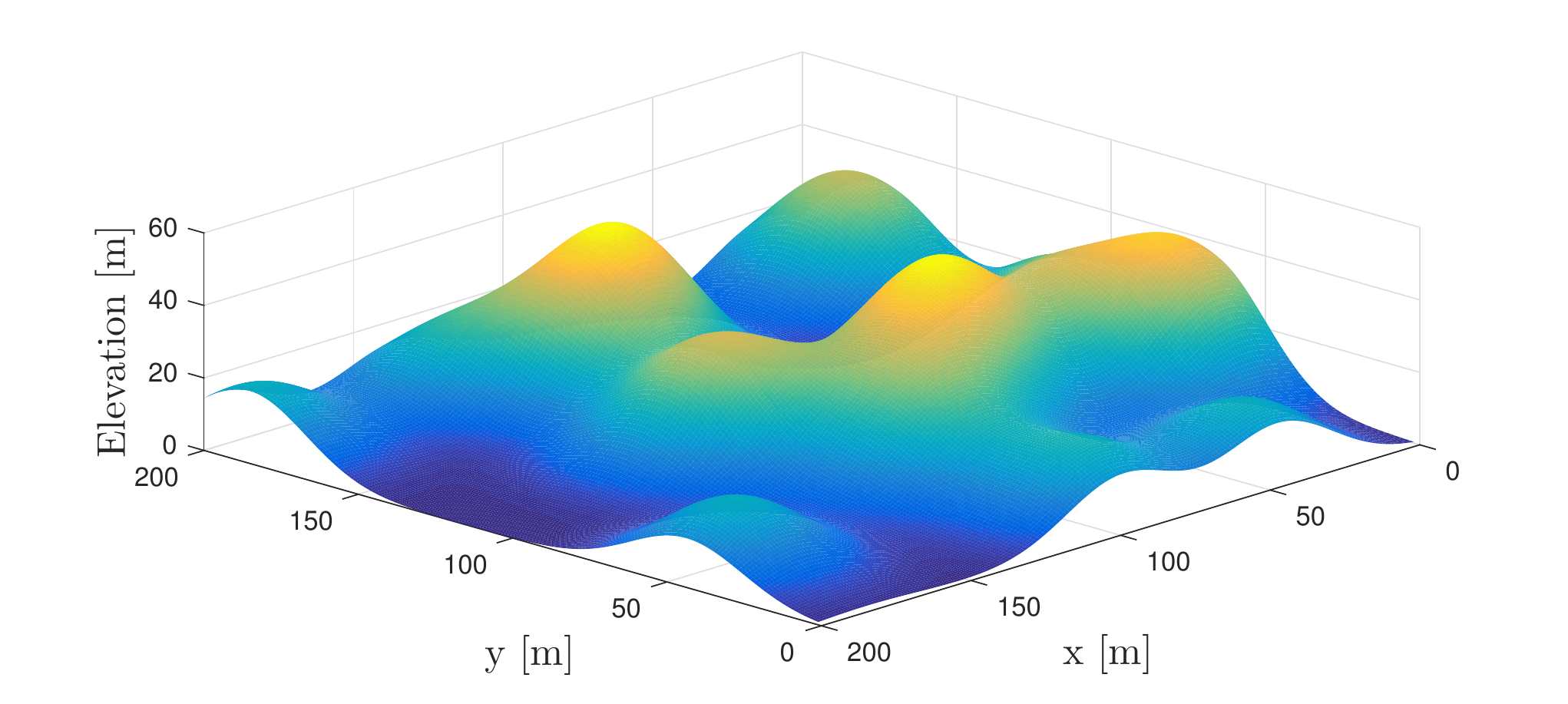}
    \vspace{-0.2cm}
	\caption{Discrete terrain model used with the Hapke model in the data cube DC3, provided by~\cite{drumetz2016blindUnmixingELMMvariability}.}
    \label{fig:terrain_hapke_ex}
\end{figure}

The data cube DC3 contains $50\times50$ pixels and three materials, and is based on a simulation originally presented in~\cite{drumetz2016blindUnmixingELMMvariability}\footnote{Most of the data for this simulation was generously provided by Lucas Drumetz and his collaborators.}.
This data cube is devised to realistically represent the spectral variability introduced due to changes in the illumination conditions caused by the topography of the scene, and is generated according to a physical model proposed by Hapke~\cite{Hapke1981}.
Hapke's model is able to represent the reflectance of a material as a function of its single scattering albedo, photometric parameters and geometric characteristics of the scene, namely, the incidence, emergence and azimuth angles during acquisition~\cite{drumetz2016blindUnmixingELMMvariability,Hapke1981}.
Thus, pixel dependent reflectance signatures for each endmember can be obtained given its single scattering albedo and the scene topography.

In this example, the scene was composed of three materials, namely, basalt, palagonite and tephra, which are frequently present on small bodies of the Solar System, and contained 16 spectral bands.
Afterwards, a digital terrain model simulating a hilly region was generated, which is shown in Fig.~\ref{fig:terrain_hapke_ex}, and from this model the acquisition angles associated with each pixel were derived (as a function of the scene topography) by considering the angle between the sun and the horizontal plane as $18^\circ$, and the sensor to be placed vertically downward.
Finally, the pixel dependent endmember signatures for the scene were generated from the single scattering albedo of the materials, and from the geometric characteristics of the scene using Hapke's model.
The abundance maps used for DC2 were the same used for DC1, as shown in Fig.~\ref{fig:abundances_DC3}.

The resulting hyperspectral images for all data cubes were generated from the pixel-dependent endmember signatures and abundance maps following the LMM, and were later contaminated by white Gaussian noise, with signal-to-noise ratios (SNR) of 20, 30, and 40~dB.
%
%
The regularization parameters for all algorithms and all examples were selected using a grid search procedure in order to provide best abundance estimation performance, and are presented in the supplemental material and in~\cite{Borsoi_multiscaleVar_2018}.

The unmixing accuracy metrics used are the abundances mean squared error (MSE)
\begin{equation}
	\text{MSE}_{\bA} = {\frac{1}{NP}\|\bA - \widehat{\!\bA}\|_F^2}
    \,,
\end{equation}
the mean squared error of the estimated spectra
\begin{equation}
	\text{MSE}_{\bbM} = {\frac{1}{NLP} \sum_{n=1}^N \|\bM_n - \widehat{\!\bM}_n\|_F^2}
    \,,
\end{equation}
and the mean squared reconstruction error
\begin{equation}
	\text{MSE}_{\bY} = \frac{1}{NL} \sum_{n=1}^N \|\by_n - \widehat{\!\bM}_n\widehat{\ba}_n\|^2
    \,.
\end{equation}

We also evaluate the estimates of the endmember signatures using the average Spectral Angle Mapper (SAM), defined by
\begin{equation}
  \text{SAM}_{\bbM} = \frac{1}{N}\sum_{n=1}^{N}\sum_{k=1}^{P}\arccos\left(\frac{\bm_{k,n}^\top\widehat{\bm}_{k,n}}{\|\bm_{k,n}\|\|\widehat{\bm}_{k,n}\|}\right).
\end{equation}
where $\bm_{k,n}$ and $\widehat{\bm}_{k,n}$ are the $k$-th columns of $\bM_n$ and $\widehat{\!\bM}_n$, respectively.

The quantitative results achieved by all algorithms are displayed in Table~\ref{tab:quantitative_results} for all tested SNR values.
The reconstructed abundance maps for the three data cubes and an SNR of 30~dB are shown in Figs.~\ref{fig:abundances_DC1},~\ref{fig:abundances_DC2} and~\ref{fig:abundances_DC3} for a qualitative comparison.
The computational complexity of the algorithms was evaluated through their execution times, which are shown in Table~\ref{tab:alg_exec_time}.

\begin{table*} [!ht]
\scriptsize
\caption{Quantitative results of all algorithms for data cubes DC1, DC2 and DC3 (with parameters selected to yield best abundance estimates). All values are multiplied by $10^3$.}
\centering
\renewcommand{\arraystretch}{1.2}
\begin{tabular}{c||c|c|c|c|c||c|c|c|c||c|c|c|cc}
\hline\hline
& & \multicolumn{4}{c||}{DC1 data cube} & \multicolumn{4}{|c||}{DC2 data cube} & \multicolumn{4}{c}{DC3 data cube} \\
\hline\hline
SNR	&	Method	&	$\text{MSE}_{\!\bA}$	&	$\text{MSE}_{\mathbb{M}}$	&	$\text{SAM}_{\!\mathbb{M}}$	&	$\text{MSE}_{\bY}$			&	$\text{MSE}_{\!\bA}$	&	$\text{MSE}_{\!\mathbb{M}}$	&	$\text{SAM}_{\!\mathbb{M}}$	&	$\text{MSE}_{\bY}$		&	$\text{MSE}_{\!\bA}$	&	$\text{MSE}_{\!\mathbb{M}}$	&	$\text{SAM}_{\!\mathbb{M}}$	&	$\text{MSE}_{\bY}$		\\
\hline\hline																																	
\multirow{5}{*}{20~dB}																																	
&	FCLS	&	21.97	&	$\times$	&	$\times$	&	6.91				&	66.47	&	$\times$	&	$\times$	&	6.45				&	74.14	&	$\times$	&	$\times$	&	2.63		\\
&	SCLS	&	28.79	&	6.87	&	190.50	&	6.86				&	73.35	&	4.07	&	171.01	&	6.20				&	73.18	&	3.02	&	214.56	&	0.50		\\
&	PLMM	&	24.64	&	5.42	&	188.80	&	3.50				&	85.65	&	3.19	&	174.35	&	3.33				&	39.07	&	\textbf{1.44}	&	\textbf{122.66}	&	0.39		\\
&	ELMM	&	17.81	&	5.34	&	\textbf{186.70}	&	5.59				&	65.11	&	\textbf{3.09}	&	\textbf{170.85}	&	6.69				&	59.54	&	2.80	&	317.44	&	\textbf{0.0001}		\\
&	MUA-SV	&	\textbf{12.90}	&	\textbf{5.24}	&	212.20	&	\textbf{1.56}				&	\textbf{29.80}	&	3.36	&	185.67	&	\textbf{3.28}				&	\textbf{28.11}	&	1.84	&	308.57	&	0.0002		\\
\hline																																	
\multirow{5}{*}{30~dB}																																	
&	FCLS	&	28.10	&	$\times$	&	$\times$	&	1.76				&	60.28	&	$\times$	&	$\times$	&	0.93				&	172.31	&	$\times$	&	$\times$	&	1.41		\\
&	SCLS	&	12.37	&	4.53	&	187.60	&	1.63				&	62.23	&	3.84	&	\textbf{161.20}	&	0.71				&	21.41	&	2.42	&	68.73	&	0.05	\\	
&	PLMM	&	19.61	&	4.88	&	173.00	&	0.86				&	49.38	&	3.95	&	162.54	&	0.41				&	38.00	&	\textbf{1.53}	&	\textbf{68.53}	&	0.10	\\	
&	ELMM	&	10.71	&	3.70	&	170.20	&	0.59				&	40.16	&	3.05	&	177.91	&	\textbf{0.001}				&	18.47	&	1.73	&	101.51	&	\textbf{0.00002}		\\
&	MUA-SV	&	\textbf{7.07}	&	\textbf{3.46}	&	\textbf{166.90}	&	\textbf{0.35}				&	\textbf{24.30}	&	\textbf{2.83}	&	161.52	&	0.33				&	\textbf{14.70}	&	1.75	&	68.62	&	0.07		\\
\hline																																	
\multirow{5}{*}{40~dB}																																	
&	FCLS	&	20.04	&	$\times$	&	$\times$	&	1.23				&	71.37	&	$\times$	&	$\times$	&	0.44				&	256.20	&	$\times$	&	$\times$	&	1.39		\\
&	SCLS	&	7.38	&	3.88	&	186.30	&	1.10				&	69.48	&	3.52	&	160.10	&	0.17				&	8.98	&	2.40	&	30.90	&	\textbf{0.01}		\\
&	PLMM	&	13.44	&	3.64	&	170.30	&	0.56				&	44.73	&	3.02	&	\textbf{140.74}	&	0.11				&	34.38	&	1.47	&	74.15	&	0.08	\\	
&	ELMM	&	5.36	&	\textbf{2.51}	&	\textbf{149.70}	&	\textbf{0.02}				&	46.83	&	\textbf{2.63}	&	159.21	&	\textbf{0.0002}				&	8.12	&	\textbf{1.28}	&	43.14	&	\textbf{0.01}	\\	
&	MUA-SV	&	\textbf{3.98}	&	2.52	&	149.90	&	\textbf{0.02}				&	\textbf{26.01}	&	2.97	&	155.96	&	0.31				&	\textbf{7.94}	&	1.81	&	\textbf{30.66}	&	0.02	\\			
\hline\hline
\end{tabular}
\label{tab:quantitative_results}
\end{table*}

\begin{table}[!htbp]
\footnotesize
\caption{Execution time (in seconds) of the unmixing algorithms, averaged for all SNR values considered}
\centering
\renewcommand{\arraystretch}{1.2}
\begin{tabular}{c||c|c|c|c|c}
\hline\hline
 & FCLS & SCLS & ELMM & PLMM & MUA-SV \\
\hline\hline
DC1 & 0.14\,s & 0.42\,s & 14.76\,s & 16.17\,s & 2.57\,s \\
\hline
DC2 & 0.27\,s & 0.83\,s & 37.52\,s & 149.91\,s & 18.29\,s \\
\hline
DC3 & 0.17\,s & 0.35\,s & 15.82\,s & 63.07\,s & 9.59\,s \\
\hline
Houston & 0.82\,s & 2.31\,s & 174.53\,s & 484.02\,s & 36.29\,s \\
\hline
Cuprite & 6.63\,s & 15.61\,s & 527.89\,s & 7998.02\,s & 95.54\,s \\
\hline
\end{tabular}
\label{tab:alg_exec_time}
\end{table}

\begin{table}[!htbp]
\footnotesize
\caption{Reconstruction errors ($\text{MSE}_{\bY}$) for the Houston and Cuprite data sets (all values are multiplied by $10^3$).}
\centering
\renewcommand{\arraystretch}{1.2}
\begin{tabular}{c||c|c|c|c|c}
\hline\hline
 & FCLS & SCLS & ELMM & PLMM & MUA-SV \\
\hline\hline
Houston & 2.283 & 0.037 & 0.010 & 0.190 & 0.014 \\
\hline
Cuprite & 0.050 & 0.044 & 0.040 & 0.079 & 0.050 \\
\hline
\end{tabular}
\label{tab:reconstr_err_real_img}
\end{table}

\subsubsection{Discussion}

Table~\ref{tab:quantitative_results} shows a significantly better $\text{MSE}_{\bA}$ performance of MUA-SV for all three data cubes and SNR values when compared with the other algorithms. This indicates that MUA-SV effectively exploits the spatial properties of the abundance maps, even when the actual spectral variability does not follow exactly the model in~\eqref{eq:model_variab_elmm}.


Figs.~\ref{fig:abundances_DC1},~\ref{fig:abundances_DC2} and~\ref{fig:abundances_DC3} show the true and reconstructed abundance maps for all algorithms and 30~dB SNR.
As expected, models accounting for spectral variability tend to yield better reconstruction quality than FCLS, with EELMM yielding piecewise smooth solutions. In general, the solution provided by MUA-SV approaches better the ground-truth, in that it estimates the intensity of the abundance maps with better accuracy than the other algorithms. This can be most clearly seen for the results for DC2 (Fig.~\ref{fig:abundances_DC2}), where the regions with pure pixels are better represented by MUA-SV.

Regarding the spectral performances, as measured by the $\text{MSE}_{\mathbb{M}}$ and $\text{SAM}_{\mathbb{M}}$, the results varied among the algorithms, with no method performing uniformly better than the others.  There is also a significant discrepancy between the Euclidean metric and the spectral angle in many examples, highlighting the different characteristics of the two metrics.

The ELMM model yielded the smallest reconstruction error $\text{MSE}_{\bY}$ in most cases (6), followed by MUA-SV (4 cases).
However, the connection between the reconstruction error~$\text{MSE}_{\bY}$ and the abundance estimation performance~$\text{MSE}_{\bA}$ of the unmixing methods that address spectral variability is not clear, as can be attested from Table~\ref{tab:quantitative_results}.

The execution times shown in Table~\ref{tab:alg_exec_time} indicate that MUA-SV is 2.2 times faster than ELMM and 7.5 times faster than PLMM, a significant gain in computational efficiency. This difference is more accentuated when processing larger datasets, as will be verified in the following.

\subsection{Sensitivity analysis}
\label{sec:sensitivity}

\begin{figure} [!htbp]
\centering
\hspace{-1.05cm}
\begin{minipage}[b]{.48\linewidth}
  \centering
  \centerline{\includegraphics[width=1\linewidth]{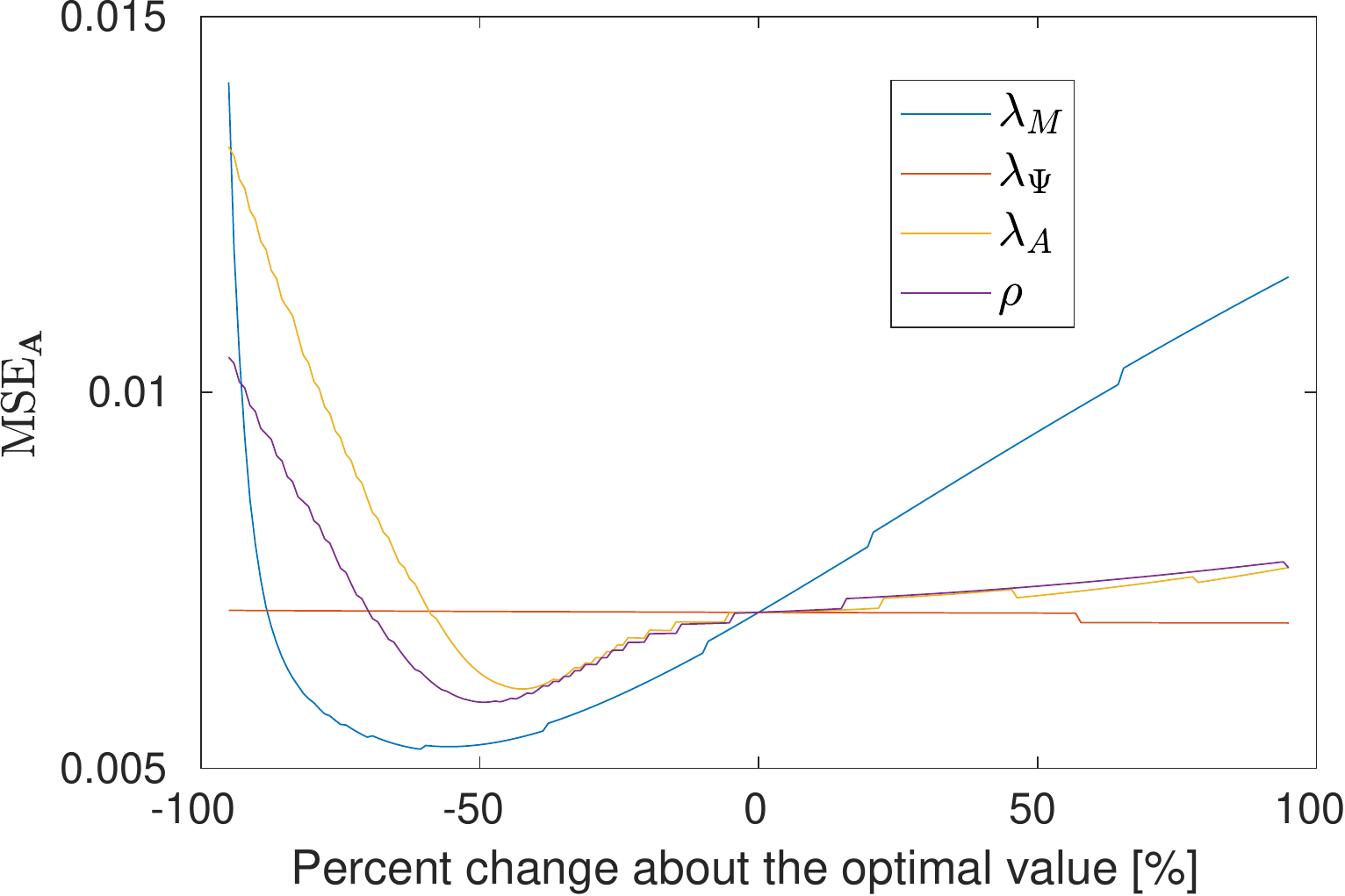}}
\end{minipage}
\hspace{-0.25cm}
\begin{minipage}[b]{.52\linewidth}
  \centering
  \centerline{\raisebox{0.4cm}{\includegraphics[width=1\linewidth]{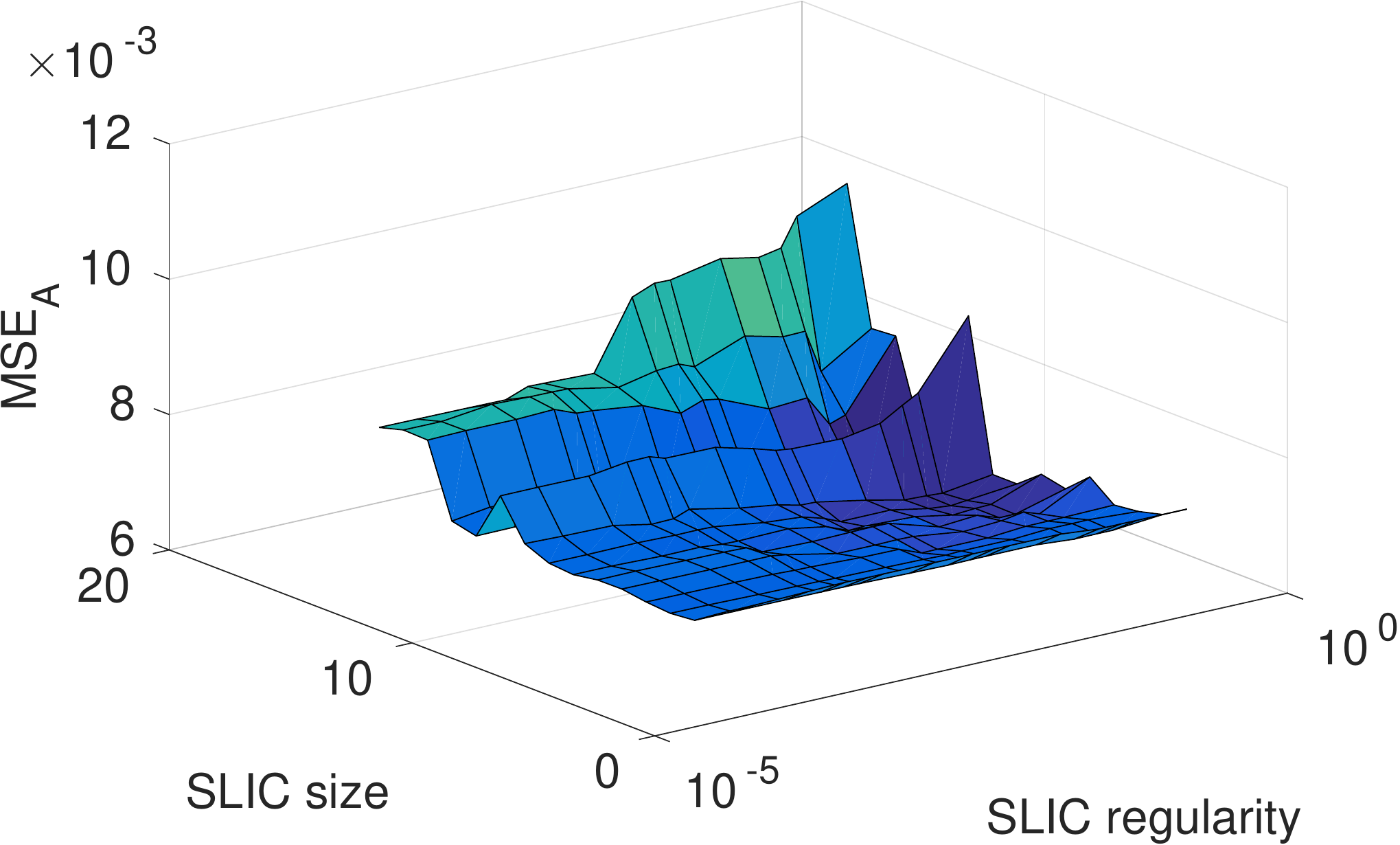}}}
\end{minipage}
\hspace{-1cm}
\caption{$\text{MSE}_{\bA}$ variation due to relative changes in each parameter value about its optimal value (left) and $\text{MSE}_{\bA}$ as a function of SLIC parameters~$\sqrt{N/S}$ and~$\gamma$ (right).}
\label{fig:sensitivity_i}
\end{figure}

To evaluate the sensitivity of MUA-SV $\text{MSE}_{\!\bA}$ to variations in the algorithm parameters\footnote{For conciseness, we present only the results for the DC1 data cube with a 30~dB SNR. The results for the other data cubes and SNRs are described in the supplemental material, also available in~\cite{Borsoi_multiscaleVar_2018}, and corroborate the conclusions presented in this section.},
we initially set all regularization parameters ($\lambda_M$, $\lambda_A$, $\lambda_\Psi$ and $\rho$) equal to their optimal values\footnote{Note that the operating point of  MUA-SV is not optimal for this case due to the relatively coarse grid employed in the parameter search procedure.}.
Then, we varied one parameter at a time within a range from $-95\%$ to $+95\%$ of its optimal value. Fig.~\ref{fig:sensitivity_i} presents the $\text{MSE}_{\!\bA}$ values obtained by varying each parameter. It can be seen that small variations about the optimal values do not affect the $\text{MSE}_{\!\bA}$ significantly, and that the maximum values obtained for the whole parameter ranges tested are still lower than those achieved by the other algorithms.

%
To evaluate the sensitivity of the MUA-SV results to variations in the SLIC parameters, we plotted the resulting $\text{MSE}_{\bA}$ as a function of~$\sqrt{N/S}$ and~$\gamma$, with the algorithm parameters $\lambda_M$, $\lambda_A$, $\lambda_\Psi$ and $\rho$ fixed at their optimal values. The results are shown in Fig.~\ref{fig:sensitivity_i}.
It is seen that the $\text{MSE}_{\bA}$ performance does not deviate significantly from its optimal value unless the superpixel size~$\sqrt{N/S}$ becomes too large. This is expected since very large superpixels may contain semantically different pixels, hindering the capability of the transform~$\bW$ to adequately capture coarse scale information. Furthermore, large values of~$\sqrt{N/S}$ may violate hypothesis \textsf{A2}, which has been used thoroughly in the derivation of the MUA-SV algorithm, and thus represent a bad design choice.

\begin{figure}[!ht]
\centering
\includegraphics[width=7.5cm]{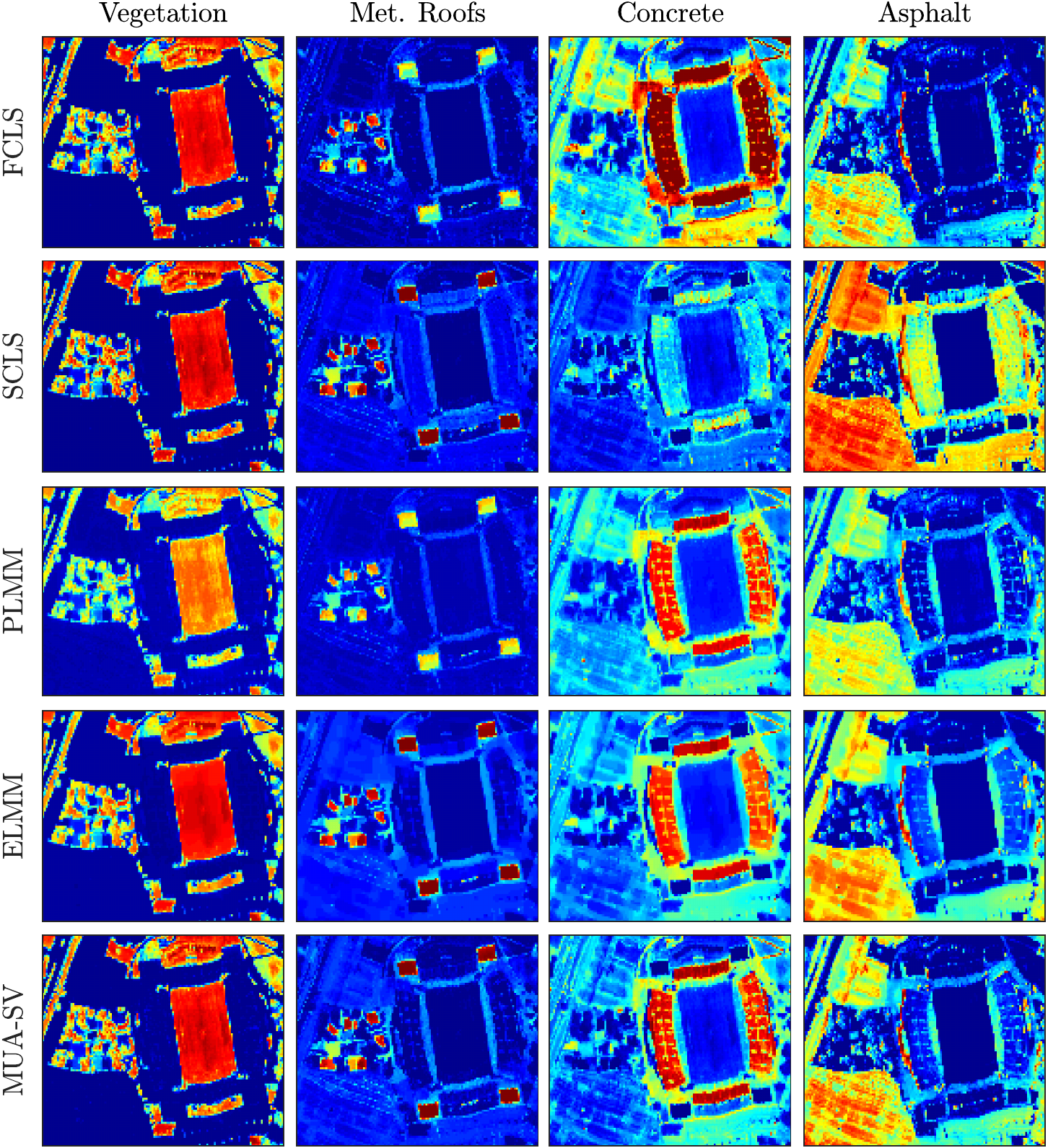}
\caption{Reconstructed fractional abundance maps for the Houston data set.}
\label{fig:abundances_houston}
\end{figure}

\begin{figure}[!ht]
\centering
\includegraphics[width=7.5cm]{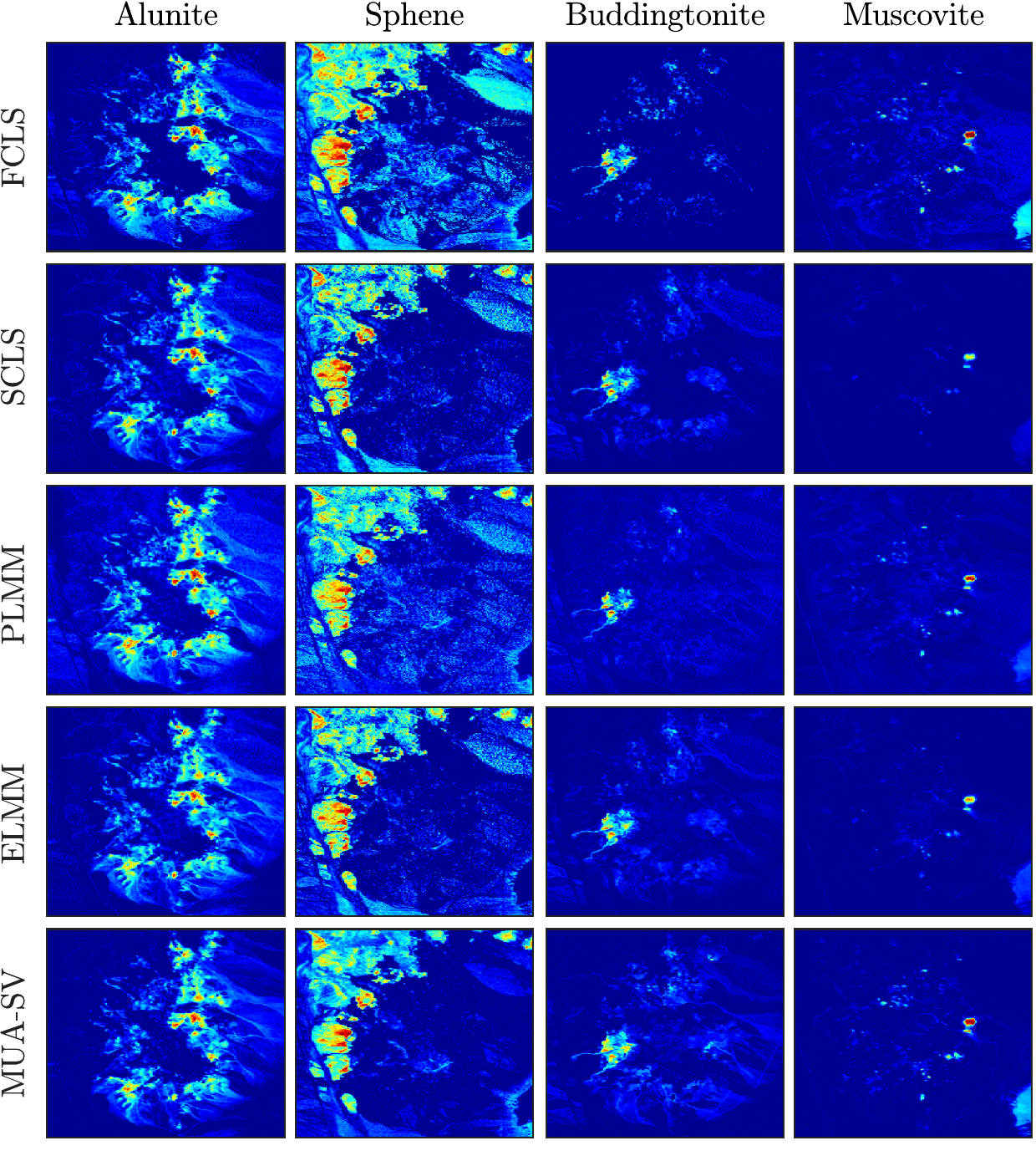}
\caption{Reconstructed fractional abundance maps for the Cuprite data set.}
\label{fig:abundances_cuprite}
\end{figure}

\subsection{Simulations with real images}

In this experiment, we consider two data sets obtained from real hyperspectral images. 
The first data set is comprised of a 152$\times$108 pixels subset of the Houston hyperspectral image, with 144 spectral bands. The second data set is a 250$\times$191 pixels subregion of the Cuprite image, with 188 spectral bands. Spectral bands presenting water absorption and low SNR were removed from both images.
The parameters of the algorithms are shown in the supplemental material and in~\cite{Borsoi_multiscaleVar_2018}. They were selected empirically for the proposed method, and set identically to those reported in~\cite{drumetz2016blindUnmixingELMMvariability} for the ELMM and PLMM.
The number of endmemebrs was selected as $P=4$ for the Houston data set, and as $P=14$ for the Cuprite data set, following the observations in~\cite{drumetz2016blindUnmixingELMMvariability}. The endmembers were extracted using the VCA algorithm~\cite{nascimento2005vca}.
%

Since the true abundance maps are unavailable for those hyperspectral images, we make a qualitative assessment of the recovered abundance maps based on knowledge of materials present in prominent fashion in those scenes.
The reconstructed abundance maps for the Houston data set are depicted in Fig.~\ref{fig:abundances_houston}. The four materials which are prominently present in this dataset are vegetation, red metallic roofs, concrete stands, and asphalt.
It can be seen that ELMM and MUA-SV yield the best results for the overall abundances of all materials, with smaller proportion indeterminacy in regions known to have mostly pure materials such as the football field, the square metallic roofs and the concrete stands in the stadium.
However, MUA-SV provides better results, more clearly observed in the purer areas such as the concrete stands of the stadium, which appear to be more mixed with the asphalt abundances in the ELMM results. This evidences the better performance of the MUA-SV algorithm.


The reconstructed abundance maps for the Alunite, Sphene, Buddingtonite and Muscovite materials of the Cuprite data set are depicted in Fig.~\ref{fig:abundances_cuprite}. Although all methods provide abundance maps which generally agree with previous knowledge about their distribution in this image~\cite{nascimento2005vca}, the MUA-SV results show abundances for all endmembers in Fig.~\ref{fig:abundances_cuprite} that are more homogeneous and clearly delineated in the regions where the materials are present. Moreover, these results show significantly smaller contributions due to outliers in the background regions of the abundance maps.

The reconstruction errors for all algorithms and both data sets are shown in Table~\ref{tab:reconstr_err_real_img}. For the Houston data, the ELMM and MUA-SV results are very close and significantly smaller than those of the other methods, what agrees with their better representation of the abundance maps.
For the Cuprite data, the errors are small and comparable for all algorithms, except for a slightly larger PLMM error. This goes in line with the fact that the abundance maps generally agree with the known distribution of these materials in the scene.
However, reconstruction error results should be taken with proper care, as observed in the examples using synthetic data. Their correlation with the quality of abundance estimation is far from straightforward.

The execution times for all methods, shown in Table~\ref{tab:alg_exec_time}, illustrate again the significantly smaller computational load of MUA-SV when compared to other methods addressing spectral variability, as it performed, on average, 5.3 times faster than ELMM and 64.3 times faster than PLMM.

\section{Conclusions}
\label{sec:conclusions}

In this paper we proposed a new data-dependent multiscale model for spectral unmixing accounting for spectral variability of the endmembers. 
Using a multiscale transformation based on the superpixel decomposition, spatial contextual information was incorporated into the unmixing problem through the decomposition of the observation model into two models in different domains, one capturing coarse image structures and another representing fine scale details. This facilitated the characterization of spatial regularity.
Under reasonable assumptions, the proposed method yields a fast iterative algorithm, in which the abundance estimation problem is solved only once in each scale.
Simulation results with both synthetic and real data show that the proposed MUA-SV algorithm outperforms other methods addressing spectral variability, both in accuracy of the reconstructed abundance maps and in computational complexity.

\appendices

\section{Derivation of the approximated mixing model}\label{app:model}


Given the coarse pixel model in~\eqref{eq:model_decomposed_ii_c} can be approximated using hypothesis \textsf{A2} as 
\begin{align} \label{app_eq:model_decomposed_iii_c}
	\by_{\calC_i} & \approx \sum_{\ell=1}^N \frac{\mathbbm{1}_{W_{\ell,i}}}{|\emph{supp}_{\ell}(W_{\ell,i})|} \bM_{\ell} \sum_{j=1}^N W_{j,i} \,\ba_i 
    + \be_{\calC_i}
    \nonumber \\
    & = \sum_{\ell=1}^N \frac{\mathbbm{1}_{W_{\ell,i}}}{|\emph{supp}_{\ell}(W_{\ell,i})|} \bM_{\ell}  \,\, \ba_{\!\calC_i} 
    + \be_{\calC_i}
    \nonumber \\
    & = \bM_{\!\calC_i} \ba_{\!\calC_i} + \be_{\calC_i}
\end{align}
where $\ba_{\!\calC_i} = \sum_{j=1}^N W_{j,i}\ba_i$. The detail model in~\eqref{eq:model_decomposed_ii_d} can be approximated as
\begin{align} \label{app_eq:model_decomposed_iii_d}
    \by_{\calD_i} & = \bM_i\ba_i - \sum_{j=1}^S \sum_{\ell=1}^N  W^\ast_{j,i} \, W_{\ell,j} \, \bM_{\ell}\, \ba_{\ell} + \be_{\calD_i}
    \nonumber \\
    & \approx \bM_i\ba_i - \bigg(\sum_{n=1}^S\sum_{m=1}^N \frac{\mathbbm{1}_{W_{n,i}^\ast} \mathbbm{1}_{W_{m,n}}}{|\emph{supp}_{n,m}(W_{n,i}^\ast W_{m,n})|} \bM_m\bigg)
    \nonumber \\
    & \qquad \cdot \sum_{j=1}^S \sum_{\ell=1}^N  W^\ast_{j,i} \, W_{\ell,j} \, \ba_{\ell} + \be_{\calD_i}
    \nonumber \\
    & = \bM_i\ba_i - \bM_{\!\calC_i^\ast} \,\sum_{j=1}^S \sum_{\ell=1}^N  W^\ast_{j,i} \, W_{\ell,j} \, \ba_{\ell} + \be_{\calD_i}
\end{align}
and straightforward computations leads to
\begin{align} 
\by_{\calD_i} & \approx \bM_i\ba_i - \bM_{\!\calC_i^\ast} \,\sum_{j=1}^S \sum_{\ell=1}^N  W^\ast_{j,i} \, W_{\ell,j} \, \ba_{\ell} + \be_{\calD_i} \nonumber \\
    & = \bM_i(\ba_{\!\calD_i}+\sum_{j=1}^S  W^\ast_{j,i} \, \ba_{\!\calC_j})
    \nonumber \\
    & \qquad - \bM_{\!\calC_i^\ast} \,\sum_{j=1}^S \sum_{\ell=1}^N  W^\ast_{j,i} \, W_{\ell,j} \, \ba_{\ell} + \be_{\calD_i}
    \nonumber \\
    & = \bM_i\bigg(\ba_{\!\calD_i}+\sum_{j=1}^S  W^\ast_{j,i} \, \ba_{\!\calC_j}\bigg) - \bM_{\!\calC_i^\ast} \,\sum_{j=1}^S W^\ast_{j,i} \, \ba_{\!\calC_j} + \be_{\calD_i}
    \nonumber \\
    & = \bM_i\ba_{\!\calD_i} + \bigg(\bM_i - \bM_{\!\calC_i^\ast}\bigg) \sum_{j=1}^S  W^\ast_{j,i} \, \ba_{\!\calC_j} + \be_{\calD_i}
    \nonumber \\
    & = \bM_i\ba_{\!\calD_i} + \bigg(\bM_i - \bM_{\!\calC_i^\ast}\bigg) \big[\bA_{\!\calC} \bW^{\ast}\big]_i + \be_{\calD_i}
    \nonumber \\
    & = \bM_i\ba_{\!\calD_i} + \bM_{\!\calD_i} \big[\bA_{\!\calC} \bW^{\ast}\big]_i + \be_{\calD_i}.
\end{align}
where $\ba_{\!\calD_i} = \ba_i-\sum_{j=1}^S W_{j,i}^*\ba_{\!\calC_j}$.

\bibliographystyle{IEEEtran}
\bibliography{references,references_revpaper}

\vspace{-1cm}

\begin{IEEEbiography}[{\includegraphics[width=1in,height=1.25in,clip,keepaspectratio]{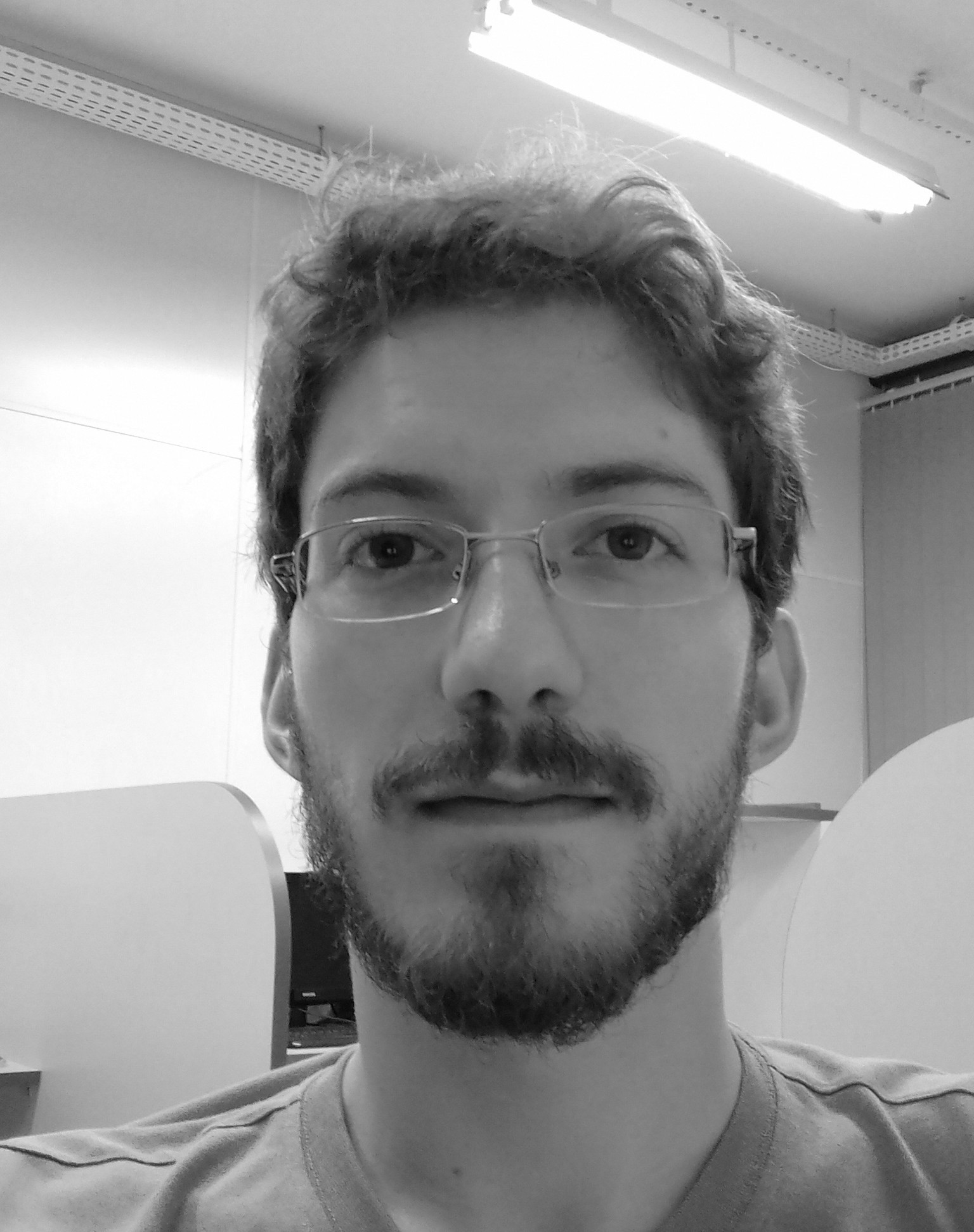}}]{Ricardo Augusto Borsoi (S'18)} 
received the MSc degree in electrical engineering from Federal University of Santa Catarina (UFSC), Florian\'opolis, Brazil, in 2016. He is currently working towards his doctoral degree at Universit\'e C\^ote d'Azur (OCA) and at UFSC. His research interests include image processing, tensor decomposition, and hyperspectral image analysis.
\end{IEEEbiography}

\begin{IEEEbiography}[{\includegraphics[width=1in,height=1.25in,clip,keepaspectratio]{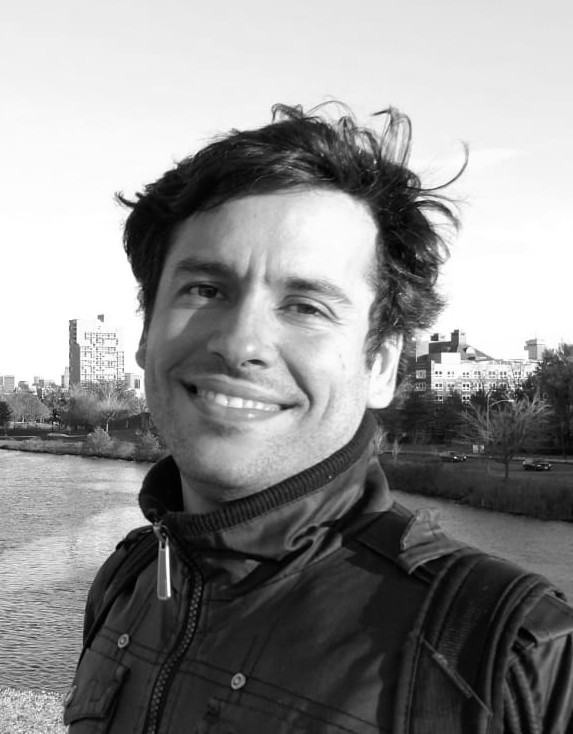}}]{Tales Imbiriba (S'14, M'17)}   
received his Doctorate degree from the Department of Electrical Engineering (DEE) of the Federal University of Santa Catarina (UFSC), Florian\'opolis, Brazil, in 2016. He served as a Postdoctoral Researcher at the DEE--UFSC and is currently a Postdoctoral Researcher at the ECE dept. of the Northeastern University, Boston, MA, USA. 
His research interests include audio and image processing, pattern recognition, kernel methods, adaptive filtering, and Bayesian Inference.
\end{IEEEbiography}

\vspace{-1cm}
\begin{IEEEbiography}[{\includegraphics[width=1in,height=1.25in,clip,keepaspectratio]{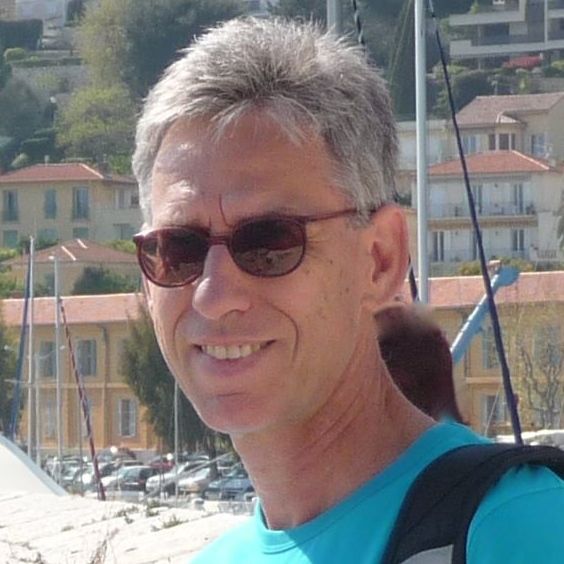}}]{Jos\'e Carlos M. Bermudez (S'78,M'85,SM'02)}
received the B.E.E. degree from the Federal University of Rio de Janeiro (UFRJ), Rio de Janeiro, Brazil, the M.Sc. degree in electrical engineering from COPPE/UFRJ, and the Ph.D. degree in electrical engineering from Concordia University, Montreal, Canada, in 1978, 1981, and 1985, respectively.
  He joined the Department of Electrical Engineering, Federal University of Santa Catarina (UFSC), Florianopolis, Brazil, in 1985. He is currently a Professor of Electrical Engineering at UFSC and a Professor at Catholic University of Pelotas (UCPel), Pelotas, Brazil. He has held the position of Visiting Researcher several times for periods of one month at the Institut National Polytechnique de Toulouse, France, and at Université Nice Sophia-Antipolis, France. He spent sabbatical years at the Department of Electrical Engineering and Computer Science, University of California, Irvine (UCI), USA, in 1994, and at the Institut National Polytechnique de Toulouse, France, in 2012. 
  His recent research interests are in statistical signal processing, including linear and nonlinear adaptive filtering, image processing, hyperspectral image processing and machine learning.
  Prof. Bermudez served as an Associate Editor of the IEEE TRANSACTIONS ON SIGNAL PROCESSING in the area of adaptive filtering from 1994 to 1996 and from 1999 to 2001. He also served as an Associate Editor of the EURASIP Journal of Advances on Signal Processing from 2006 to 2010, and as a Senior Area Editor of the IEEE TRANSACTIONS ON SIGNAL PROCESSING from 2015 to 2019.  He is the Chair of the Signal Processing Theory and Methods Technical Committee of the IEEE Signal Processing Society. Prof. Bermudez is a Senior Member of the IEEE.
\end{IEEEbiography}

\onecolumn
\centerline{{\huge Supplemental Material: A Data Dependent Multiscale Model}} \medskip
\centerline{{\huge for Hyperspectral Unmixing With Spectral Variability}}


\section{SLIC Superpixels for HIs}

The SLIC superpixel decomposition consists of an extension of the k-means algorithm, with properly initialized cluster centers and a suitable distance function, defined as~[S1]
\[D_{SLIC} = \sqrt{d_{spectral}^2 + \gamma^2 d_{spatial}^2 S/N} \]
where $d_{spatial}$ and $d_{spectral}$ are the spatial and spectral distances, respectively.
Although the SLIC algorithm was initially designed to work with color (3 bands) images, it can be extended to HIs straightforwardly by considering $d_{spectral}$ to be the Euclidean distance between reflectance vectors (HI pixels) and adjusting the normalization factor~$\gamma$ accordingly.

The superpixel transform requires the number of clusters $S$ and their regularity $\gamma$ as parameters to compute the transformation $\bY\bW$. Nevertheless, we found that it is often easier to design the transform using the parameter $\sqrt{N/S}$ instead of $S$ since it is invariant to the image size and corresponds to the average sampling interval in the irregular domain. This quantity changes only slightly on a relatively short interval between the different simulations.

\section{Numerical verification of the simplifying hypothesis}

Although hypothesis \textsf{A1} and \textsf{A2} impose some limitation to the MUA-SV algorithm, they are reasonable and are satisfied in many practical circumstances. Below, we present a more thorough analysis of each of these hypotheses.

\medskip
Hypothesis \textsf{A1} consists of assuming that the inner product $\langle RE_{\calC},RE_{\calD}\rangle$ between the residuals/reconstruction errors $RE_{\calC}$ and $RE_{\calD}$ in the coarse and detail image scales is comparatively small, when compared to the first two terms of the cost function~(22). To illustrate the validity of this claim, we compare here the values of $\langle RE_{\calC},RE_{\calD}\rangle$ with those of the first two terms of the cost function, given by $\|RE_{\calC}\|_F^2$ and $\|RE_{\calD}\|_F^2$, for some practical examples. We considered the result of unmixing DC1, DC2 and DC3 with an SNR of 30~dB presented in Section~VIII using the ELMM model. The results are presented below in Table~\ref{tab:hypothesis_A1_ver}.
\begin{table}[h]
\renewcommand{\arraystretch}{1.3}
\centering
\caption{Comparison between the residuals inner product and the first two terms of the cost function}
\begin{tabular}{c||c|r}
    & $\|RE_{\calC}\|_F^2+\|RE_{\calD}\|_F^2$ & $\langle RE_{\calC},RE_{\calD}\rangle$
    \\\hline\hline
    DC1 & 328.35 & -1.316$\times10^{-15}$ \\
    DC2 & 0.5605 &  2.845$\times10^{-16}$ \\
    DC3 & 7.105$\times10^{-4}$ & 1.948$\times10^{-19}$
\end{tabular}
\label{tab:hypothesis_A1_ver}
\end{table}
It can be seen that the quadratic norms exceed this inner product in value by several orders of magnitude. Thus, the latter can be reasonably neglected, i.e. $\langle RE_{\calC},RE_{\calD}\rangle\approx0$.

\medskip
Hypothesis \textsf{A2} basically states that the endmember signatures for each pixel $\bM_n$ do not deviate much from the average endmember signature in its neighborhood, i.e. $\bM_n$ is similar to $\frac{1}{|\mathcal{N}_n|}\sum_{j\in\mathcal{N}_n}\bM_j$ where $\mathcal{N}_n$ contains indexes of pixels that are spatially close to pixel~$n$. This is an assumption about the underlying physical model that is reasonable in practical scenarios. To illustrate this, we consider two experiments, one based on the Hapke model and another based on real data.

For instance, using synthetic data generated using the Hapke model~[S2] we can represent spectral variability due to topographic variations of the scene. Consider the discrete terrain model and reference endmember signatures presented in Figure~\ref{fig:refEms_terrain_hapke_A2} below, extracted from~[S3]. From this data and using the Hapke model, one can generate a set of pixel dependent endmember signatures which can be used to evaluate the spatial characteristics of spectral variability. For simplicity, we measure the similarity between the reference and the pixel dependent endmember signatures using both the Euclidean distance and the spectral angle, for all materials. The results are shown in Figure~\ref{fig:hypothesis_A2_ver_ii} below, where it can be seen that there these deviations show significant spatial correlation.

\begin{figure}[bh]
	\centering
	\includegraphics[width=6cm]{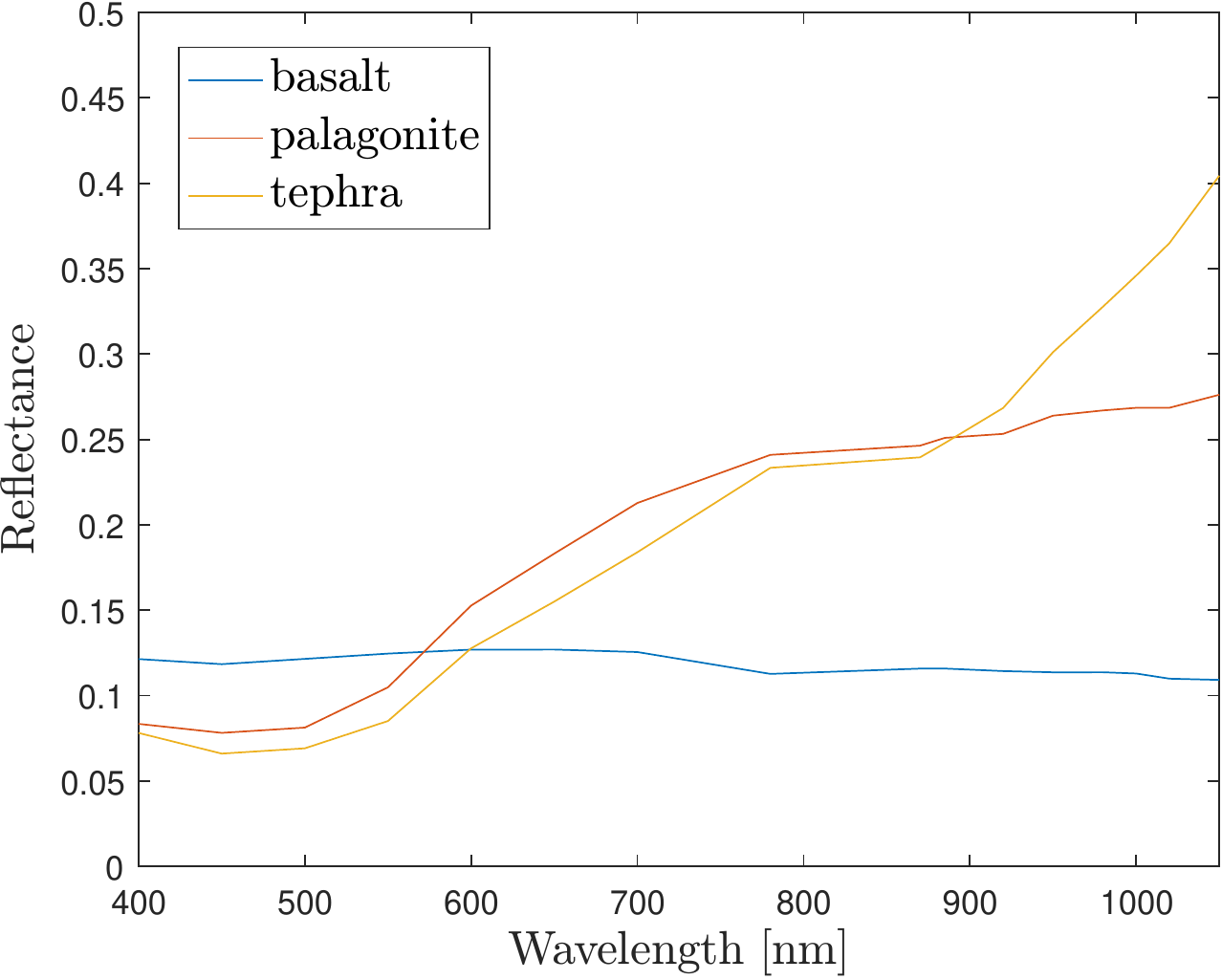}
    \includegraphics[width=7.5cm,height=5cm]{figures/terrain-crop}
    \vspace{-0.2cm}
	\caption{Reference endmember signatures (left) and discrete terrain model (right) used with the Hapke model in the data cube DC3 to generate the pixel-dependent endmember signatures (data provided by~\cite{drumetz2016blindUnmixingELMMvariability}).}
    \label{fig:refEms_terrain_hapke_A2}
\end{figure}
\begin{figure}
    \centering
    \includegraphics[width=10cm]{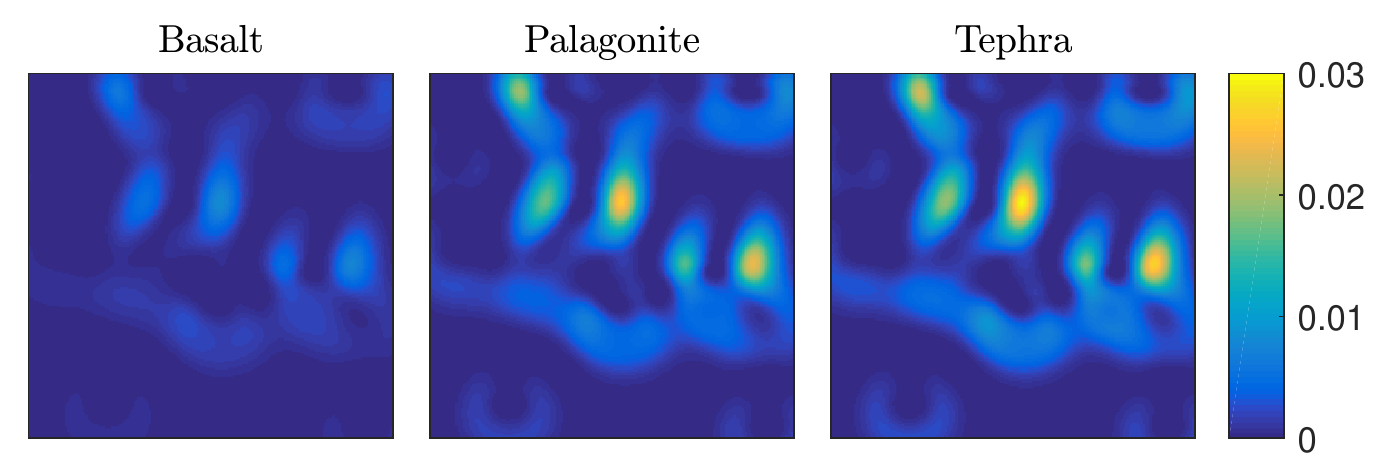} \\[0.2cm]
    \includegraphics[width=10cm]{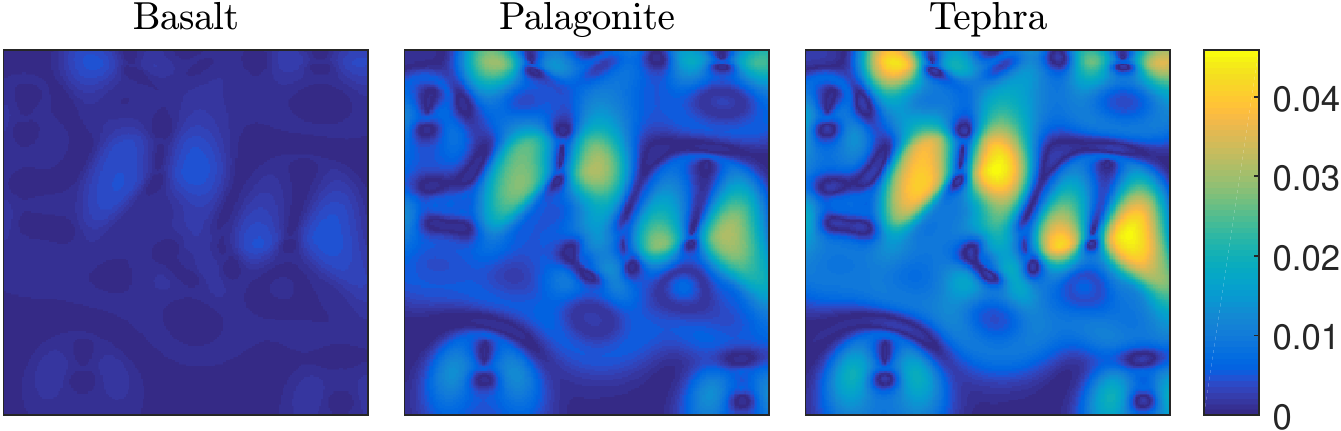}
    \caption{Measures of endmember spatial variability in the Hapke model. Top row: Euclidean distance between the soil spectral signature of each pixel and the reference signature. Bottom row: Spectral angle between the soil spectral signature of each pixel and the reference signature.}
    \label{fig:hypothesis_A2_ver_ii}
\end{figure}

Spectral variability occurring due to intrinsic variations of the material spectra (e.g. soil or vegetation) can also show significant spatial correlation~(see~[S4]), since endmember spectra usually depends on physical quantities that are correlated in space.
Many experimental studies support this claim, including geostatistical works evaluating the spatial distribution and variability of soil's physico-chemical properties (e.g. for grass crop terrain~(see~[S5]), calcareous soils~(see~[S6]), rice fields~(see~[S7]) and tobacco plantations~(see~[S8]), and also measurements of mineral spectra due to the presence of spatially correlated grain sizes and impurity concentrations~(see~[S9],[S10]).

\begin{figure}
    \centering
    \begin{minipage}{.5\textwidth} \centering
    \includegraphics[height=5cm]{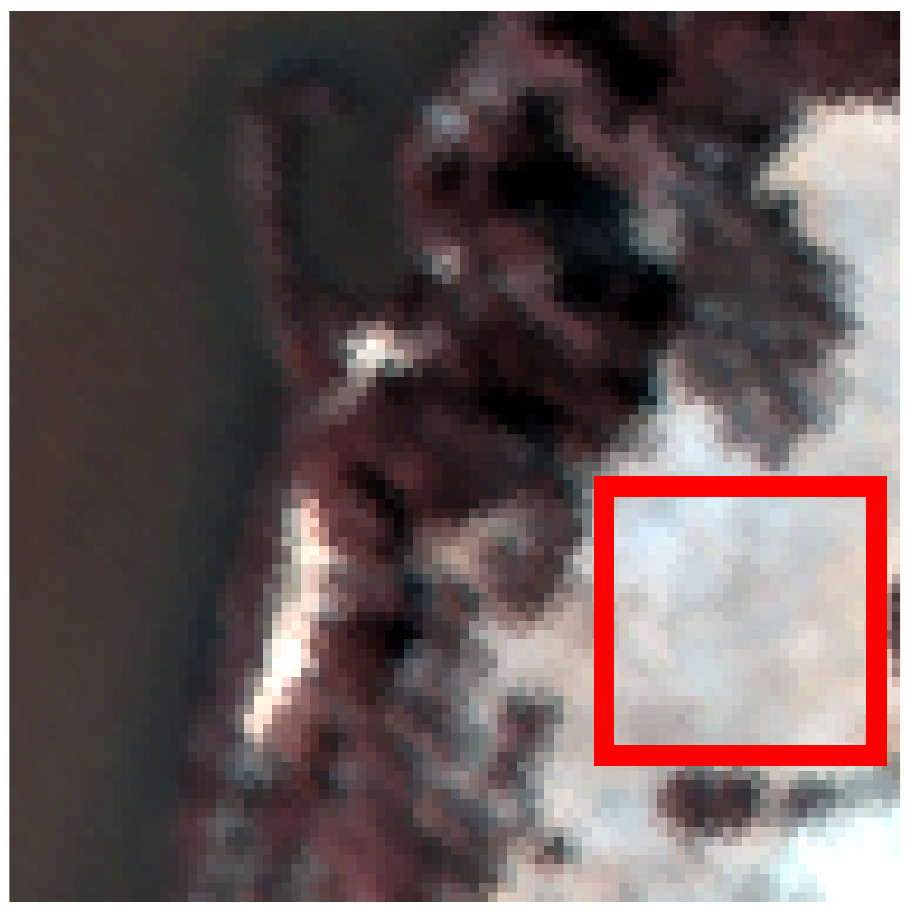} \\ (a)
    \end{minipage}\\
    ~
    \begin{minipage}{.4\textwidth} \centering
    \includegraphics[height=4.75cm]{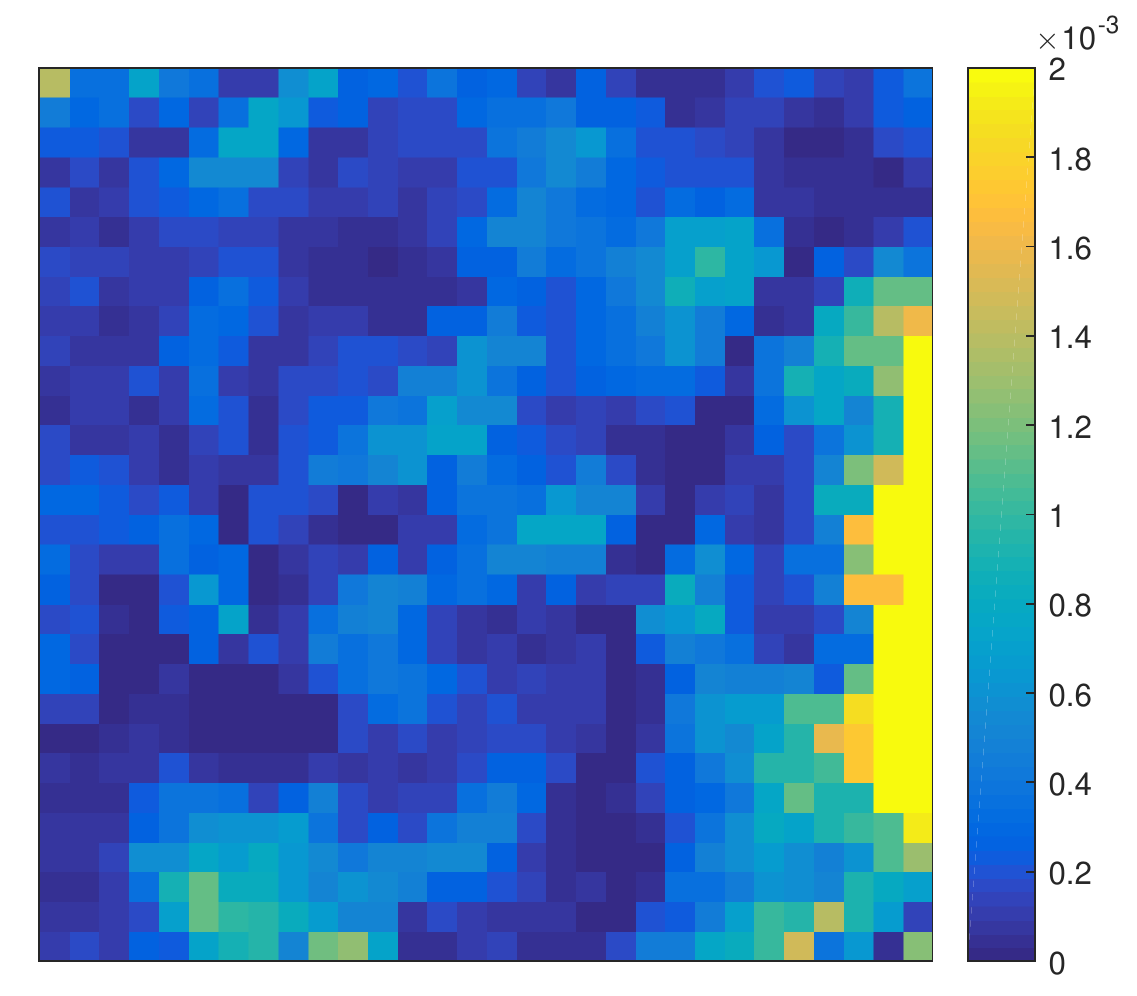} \\ (b)
    \end{minipage}
    ~
    \begin{minipage}{.4\textwidth} \centering
    \includegraphics[height=4.75cm]{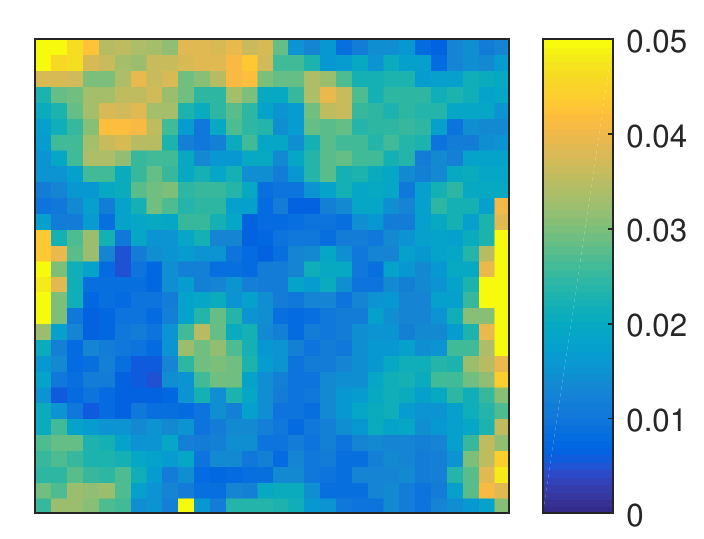} \\ (c)
    \end{minipage}
    \caption{(a) Samson hyperspectral image with a subimage containing soil highlighted. (b) Euclidean distance between the soil spectral signature of each pixel and their average value. (c) Spectral angle between the soil spectral signature of each pixel and their average value.}
    \label{fig:hypothesis_A2_ver_iii}
\end{figure}

To illustrate this effect, we performed an experiment considering real data using the Samsom image. We considered a subregion containing pure pixels of the soil material, shown in Figure~\ref{fig:hypothesis_A2_ver_iii}-(a) below. We considered these pixels as pixel dependent endmember signatures and evaluated the similarity between them and the average endmember spectra for all these pixels. The results, shown in Figures~\ref{fig:hypothesis_A2_ver_iii}-(b) and~\ref{fig:hypothesis_A2_ver_iii}-(c), are similar to the Hapke data, and illustrate that the variability shows considerable spatial correlation.

\section{Estimated scaling factors for the ELMM and MUA-SV algorithms}

\begin{figure}
        \centering
        \begin{minipage}{.5\textwidth} \centering
        \includegraphics[height=5cm,width=9cm]{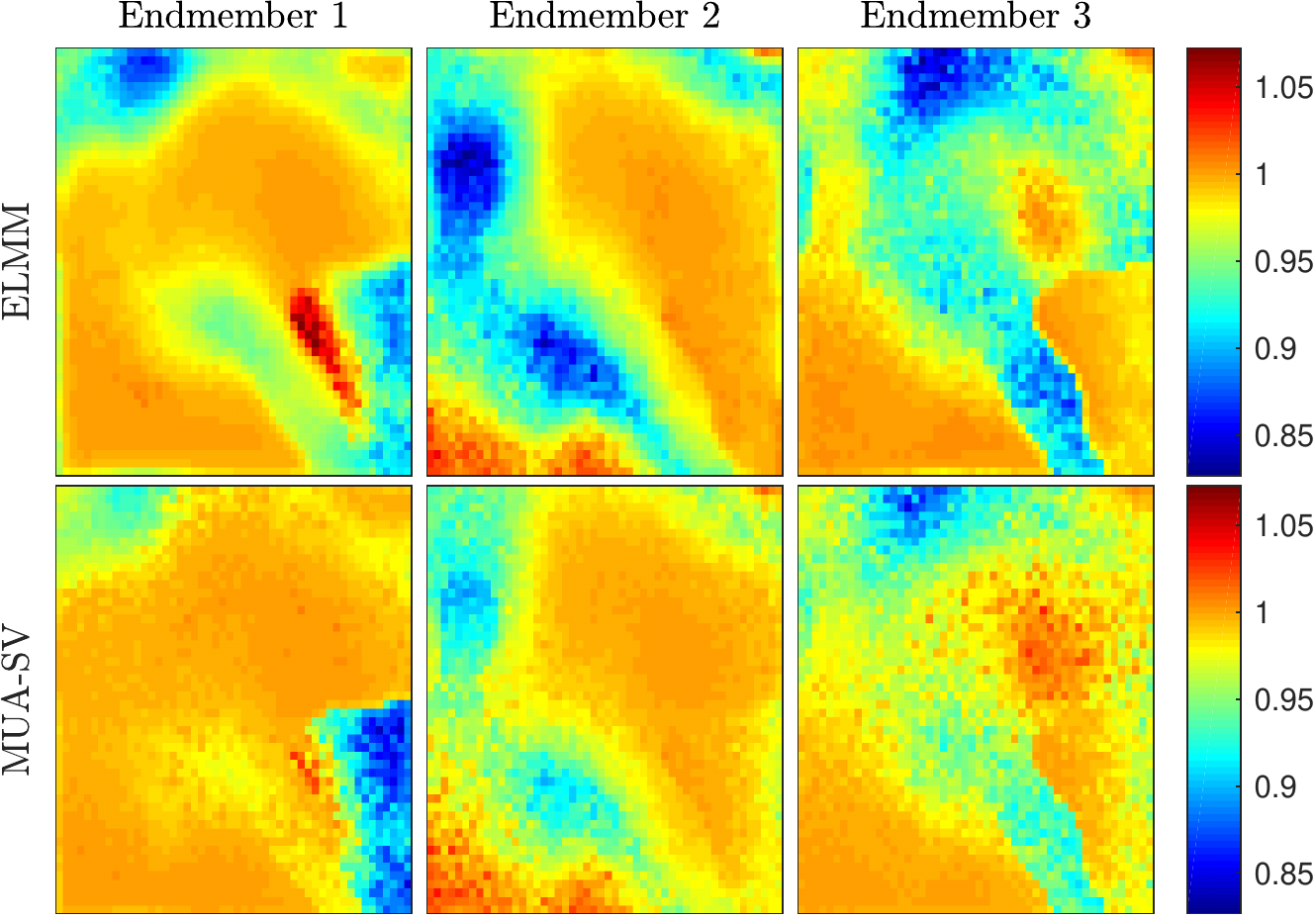} \\ (a)
        \end{minipage}\\
        \begin{minipage}{.5\textwidth} \centering
        \includegraphics[height=5cm,width=9cm]{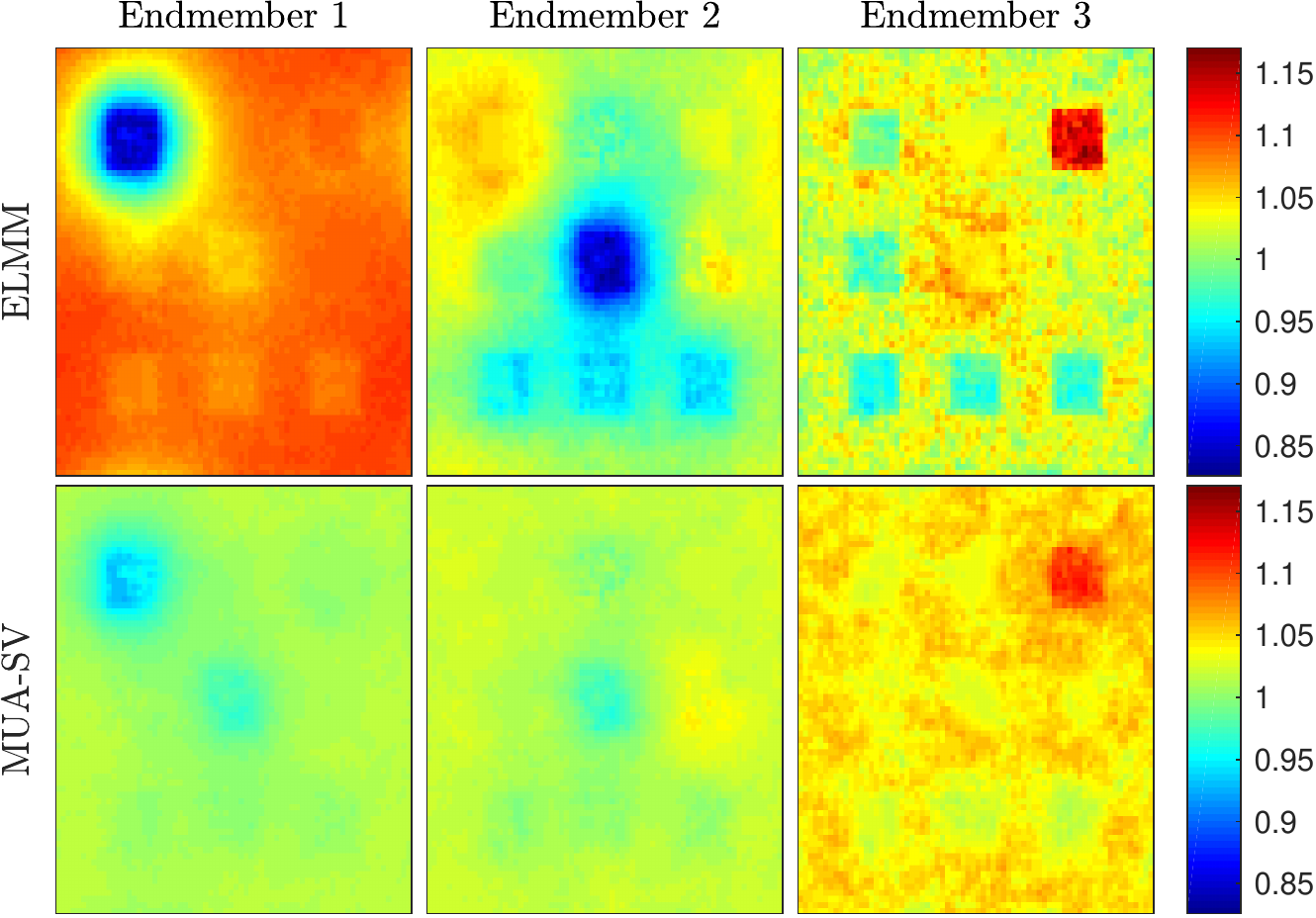} \\ (b)
        \end{minipage}\\
        \begin{minipage}{.5\textwidth} \centering
        \includegraphics[height=5cm,width=9cm]{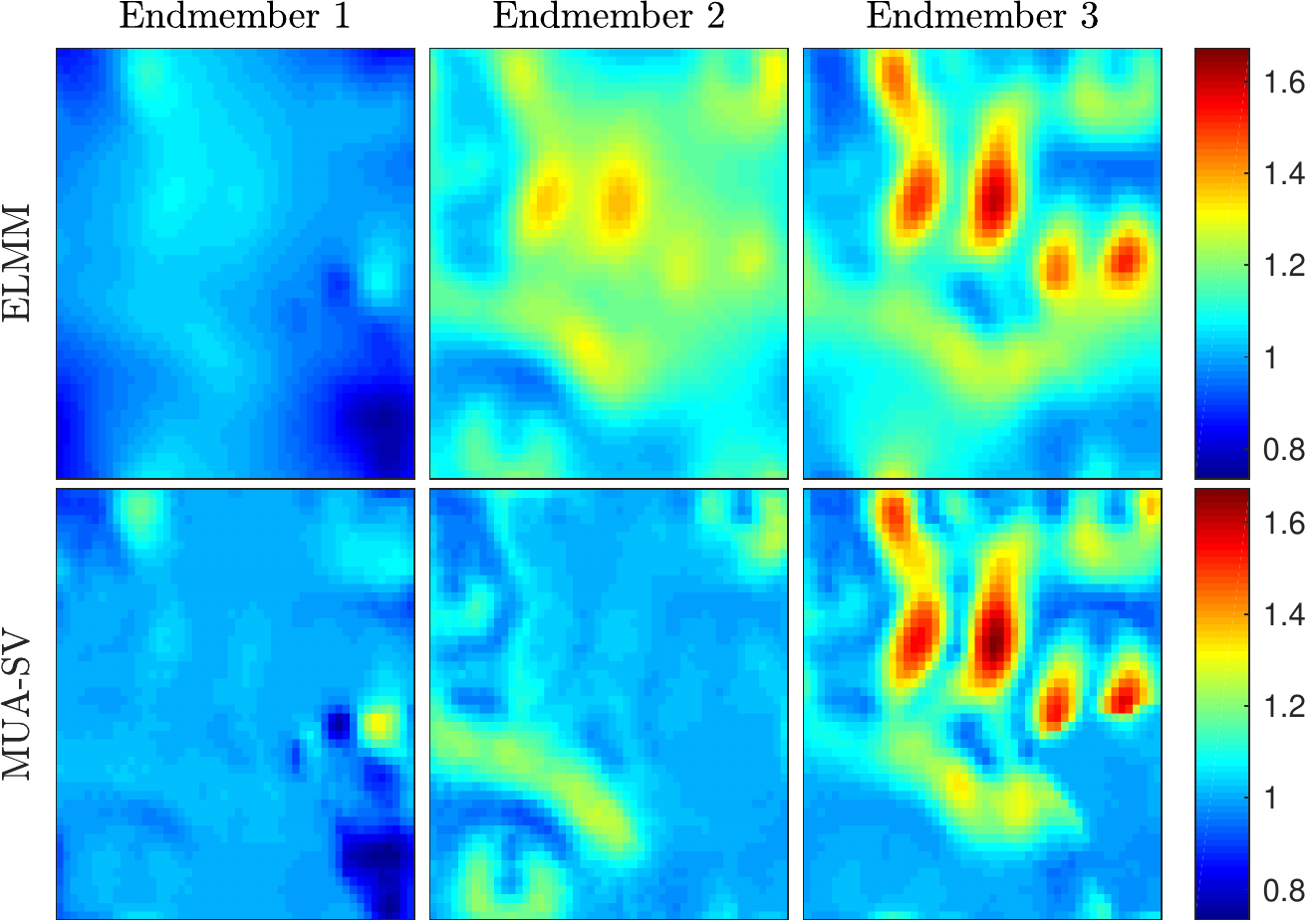} \\ (c)
        \end{minipage}
        \caption{Comparison between the scaling factors of the ELMM and MUA-SV algorithms for (a) DC1, (b) DC2 and (c) DC3, for an SNR of 30dB.}
        \label{fig:scaling_facts_example}
    \end{figure}

\begin{figure}
        \centering
        \includegraphics[height=7cm,width=9cm]{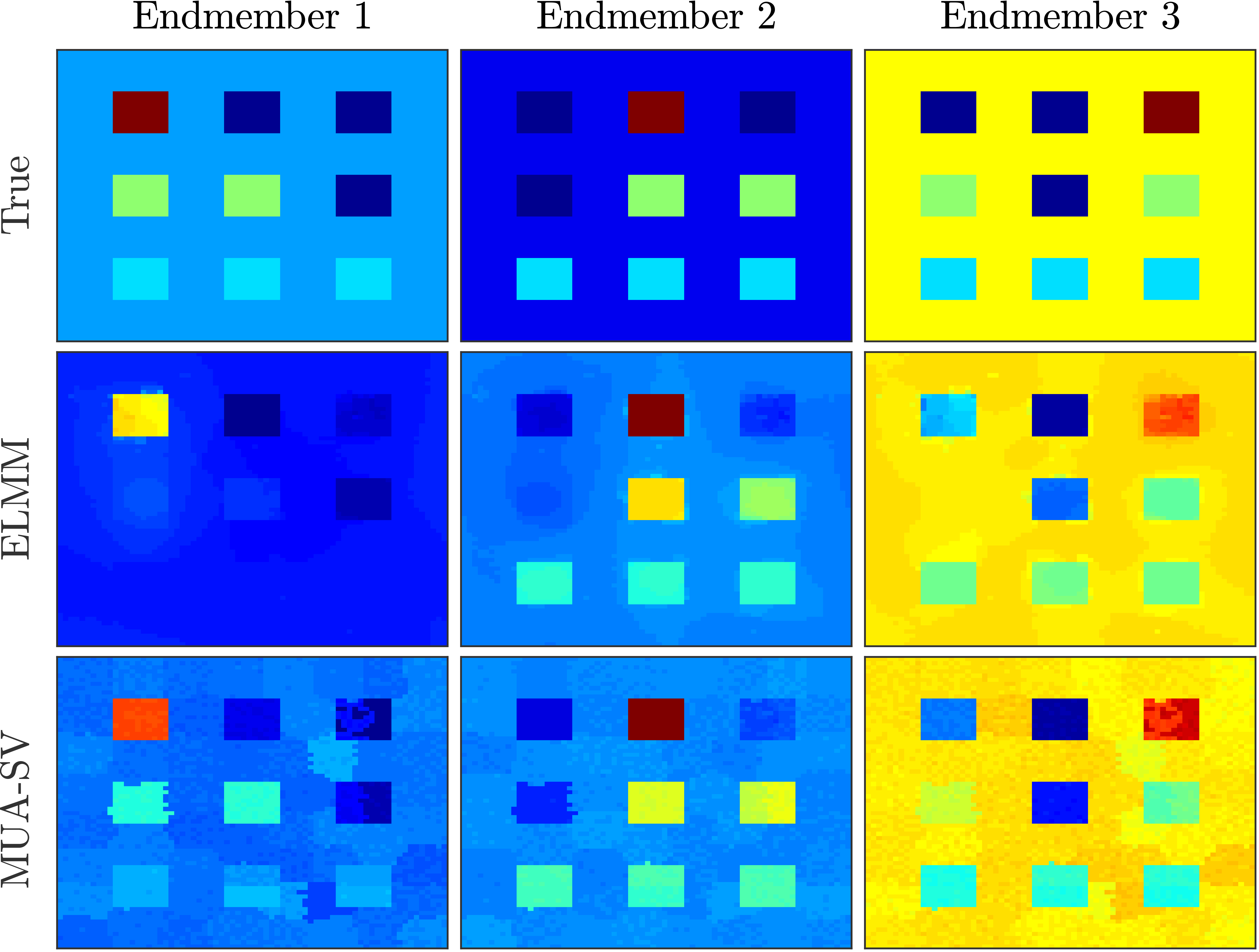} 
        \caption{Abundance maps estimated by the ELMM and MUA-SV algorithms for DC2 with an SNR of 30dB.}
        \label{fig:scaling_facts_example_abundances}
    \end{figure}

We have plotted the scaling factors $\psi_n$, $n=1,\ldots,N$ for ELMM and for the MUA-SV algorithms, for data cubes DC1, DC2 and DC3 and an SNR of 30~dB. They are shown in Figure~\ref{fig:scaling_facts_example}. It can be seen that the overall spatial variations of the scaling factors generally occur in the same regions for both algorithms, except for some endmembers such as EMs 1 and 2 for DC2 and EM2 in DC3. This difference is most easily relatable to the abundance estimation in the case of EM 1 in DC2, where the abundances estimated by the ELMM (shown in Figure~\ref{fig:scaling_facts_example_abundances}) deviate significantly from the ground truth in a pattern that is very similar to the estimated scaling factors, with a different overall scaling and a significantly smaller amplitude in the upper-left square.
    

\section{Parameter Selection}


For the synthetic data, the parameters were selected by exhaustive search within the range of values used by the respective authors in the original papers, aiming at achieving the minimum MSE for the reconstructed abundances. They are depicted in Table~\ref{tab:alg_param_dc1_dc2_optA} for all data cubes and all SNRs. For the real data, the parameters for the MUA-SV were selected in order to produce coherent abundance maps. For the other methods the parameters were extracted from~[S3]. All parameters used with real data simulations are displayed in Table~\ref{tab:alg_param_realData}.

\begin{table*} [!ht]
\footnotesize
\caption{Parameters of the algorithms used for unmixing data cubes DC1, DC2 and DC3 (selected in order to provide the best abundance estimation performance).}
\begin{center}
\renewcommand{\arraystretch}{1.2}
\begin{tabular}{c||c|c}
\hline\hline
\multicolumn{3}{c}{DC1 data cube}\\
\hline\hline
SNR & Method & Parameters\\
\hline\hline
\multirow{5}{*}{20dB} 
& FCLS & $\times$ \\
& SCLS & $\times$ \\
& PLMM & $\alpha=0.01$, $\beta=1000$, $\gamma=1.5$\\
& ELMM & $\lambda_M=5$, $\lambda_{\bPsi}=0.0005$, $\lambda_A=0.5$ \\
& MUA-SV & $\lambda_M=0.5$, $\lambda_{\bPsi}=10$, $\lambda_A=1$, $\rho\,S^2/N^2=0.5$, $\sqrt{N/S}=5$, $\gamma=0.005$ \\
\hline
\multirow{5}{*}{30dB} 
& FCLS & $\times$ \\
& SCLS & $\times$ \\
& PLMM & $\alpha=0.01$, $\beta=10^3$, $\gamma=1.5$\\
& ELMM & $\lambda_M=1$, $\lambda_{\bPsi}=0.5$, $\lambda_A=0.01$ \\
& MUA-SV & $\lambda_M=0.5$, $\lambda_{\bPsi}=0.5$, $\lambda_A=1$, $\rho\,S^2/N^2=0.1$, $\sqrt{N/S}=3$, $\gamma=0.001$ \\
\hline
\multirow{5}{*}{40dB} 
& FCLS & $\times$ \\
& SCLS & $\times$ \\
& PLMM & $\alpha=0.01$, $\beta=1000$, $\gamma=1.5$\\
& ELMM & $\lambda_M=0.1$, $\lambda_{\bPsi}=0.5$, $\lambda_A=0.005$ \\
& MUA-SV & $\lambda_M=0.1$, $\lambda_{\bPsi}=1$, $\lambda_A=0.5$, $\rho\,S^2/N^2=0.1$, $\sqrt{N/S}=3$, $\gamma=0.01$ \\
\hline\hline
\multicolumn{3}{c}{DC2 data cube}\\
\hline\hline
SNR & Method & Parameters\\
\hline\hline
\multirow{5}{*}{20dB} 
& FCLS & $\times$ \\
& SCLS & $\times$ \\
& PLMM & $\alpha=0.000005$, $\beta=1000$, $\gamma=1.5$\\
& ELMM & $\lambda_M=50$, $\lambda_{\bPsi}=0.5$, $\lambda_A=0.5$ \\
& MUA-SV & $\lambda_M=1$, $\lambda_{\bPsi}=50$, $\lambda_A=100$, $\rho\,S^2/N^2=0.005$, $\sqrt{N/S}=9$, $\gamma=0.01$ \\
\hline
\multirow{5}{*}{30dB} 
& FCLS & $\times$ \\
& SCLS & $\times$ \\
& PLMM & $\alpha=0.01$, $\beta=0.05$, $\gamma=1.5$\\
& ELMM & $\lambda_M=0.005$, $\lambda_{\bPsi}=0.5$, $\lambda_A=0.05$ \\
& MUA-SV & $\lambda_M=0.5$, $\lambda_{\bPsi}=50$, $\lambda_A=50$, $\rho\,S^2/N^2=0.005$, $\sqrt{N/S}=9$, $\gamma=0.01$ \\
\hline
\multirow{5}{*}{40dB} 
& FCLS & $\times$ \\
& SCLS & $\times$ \\
& PLMM & $\alpha=0.00005$, $\beta=10$, $\gamma=1.5$\\
& ELMM & $\lambda_M=0.005$, $\lambda_{\bPsi}=0.5$, $\lambda_A=0.05$ \\
& MUA-SV & $\lambda_M=1$, $\lambda_{\bPsi}=50$, $\lambda_A=100$, $\rho\,S^2/N^2=0.005$, $\sqrt{N/S}=9$, $\gamma=0.01$ \\
\hline\hline
\multicolumn{3}{c}{DC3 data cube}\\
\hline\hline
SNR & Method  & Parameters\\
\hline\hline
\multirow{5}{*}{20dB} 
& FCLS & $\times$ \\
& SCLS & $\times$ \\
& PLMM & $\alpha=0.01$, $\beta=50$, $\gamma=1.5$\\
& ELMM & $\lambda_M=0.005$, $\lambda_{\bPsi}=0.0005$, $\lambda_A=0.01$ \\
& MUA-SV & $\lambda_M=0.01$, $\lambda_{\bPsi}=0.05$, $\lambda_A=0.005$, $\rho\,S^2/N^2=0.001$, $\sqrt{N/S}=4$, $\gamma=0.05$ \\
\hline
\multirow{5}{*}{30dB} 
& FCLS & $\times$ \\
& SCLS & $\times$ \\
& PLMM & $\alpha=0.01$, $\beta=50$, $\gamma=1$\\
& ELMM & $\lambda_M=0.005$, $\lambda_{\bPsi}=0.01$, $\lambda_A=0.01$ \\
& MUA-SV & $\lambda_M=5$, $\lambda_{\bPsi}=0.05$, $\lambda_A=0.01$, $\rho\,S^2/N^2=0.001$, $\sqrt{N/S}=4$, $\gamma=0.01$ \\
\hline
\multirow{5}{*}{40dB} 
& FCLS & $\times$ \\
& SCLS & $\times$ \\
& PLMM & $\alpha=0.01$, $\beta=50$, $\gamma=1$\\
& ELMM & $\lambda_M=0.1$, $\lambda_{\bPsi}=0.5$, $\lambda_A=0.0005$ \\
& MUA-SV & $\lambda_M=5$, $\lambda_{\bPsi}=0.01$, $\lambda_A=0.01$, $\rho/\sqrt{N/S}^4=0.001$, $\sqrt{N/S}=2$, $\gamma=0.0005$ \\
\hline\hline
\end{tabular}
\end{center}
\label{tab:alg_param_dc1_dc2_optA}
\end{table*}

\begin{table*} [!ht]
\footnotesize
\caption{Parameters of the algorithms used for unmixing Houston and Cuprite data cubes.}
\begin{center}
\renewcommand{\arraystretch}{1.2}
\begin{tabular}{c||c|c}
\hline\hline
Dataset & Method & Parameters\\
\hline\hline
\multirow{5}{*}{Houston} 
& FCLS & $\times$ \\
& SCLS & $\times$ \\
& PLMM & $\alpha=0.0014$, $\beta=500$, $\gamma=1$\\
& ELMM & $\lambda_M=0.4$, $\lambda_{\bPsi}=0.001$, $\lambda_A=0.005$ \\
& MUA-SV & $\lambda_M=0.5$, $\lambda_{\bPsi}=0.001$, $\lambda_A=0.001$, $\rho\,S^2/N^2=0.35$, $\sqrt{N/S}=5$, $\gamma=0.001$ \\
\hline\hline
\multirow{5}{*}{Cuprite} 
& FCLS & $\times$ \\
& SCLS & $\times$ \\
& PLMM & $\alpha=0.00031$, $\beta=500$, $\gamma=1$\\
& ELMM & $\lambda_M=0.4$, $\lambda_{\bPsi}=0.005$, $\lambda_A=0.005$ \\
& MUA-SV & $\lambda_M=5$, $\lambda_{\bPsi}=0.01$, $\lambda_A=0.05$, $\rho\,S^2/N^2=0.01$, $\sqrt{N/S}=6$, $\gamma=0.001$ \\
\hline\hline
\end{tabular}
\end{center}
\label{tab:alg_param_realData}
\end{table*}

\section{Sensitivity Analysis}

\begin{figure} [!htbp]
\centering
\hspace{-1.0cm}
\begin{minipage}[b]{.4\linewidth}
  \centering
  \centerline{\includegraphics[width=1\linewidth]{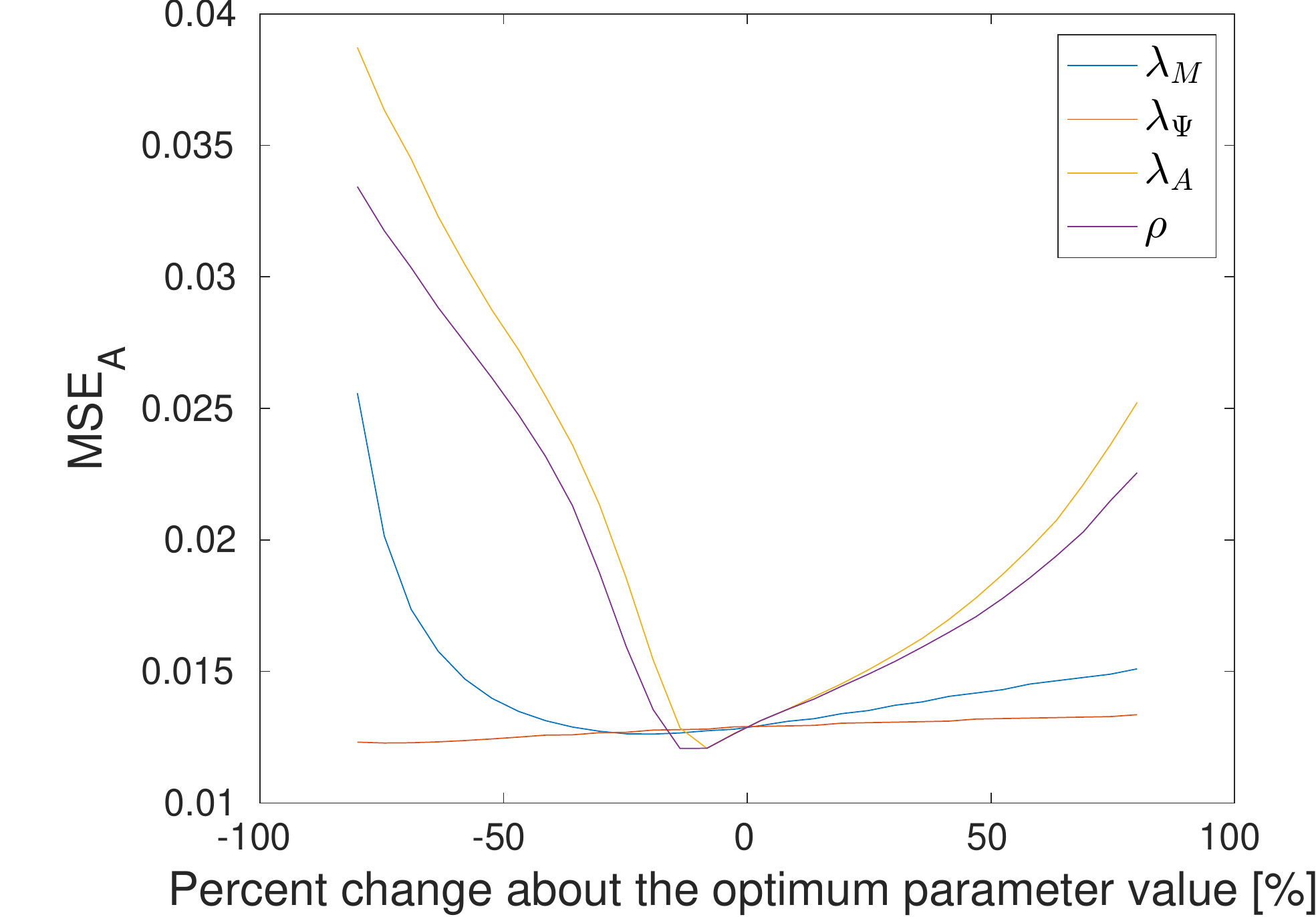}}
\end{minipage}
\begin{minipage}[b]{.43\linewidth}
  \centering
  \centerline{\raisebox{0.85cm}{\includegraphics[width=1\linewidth]{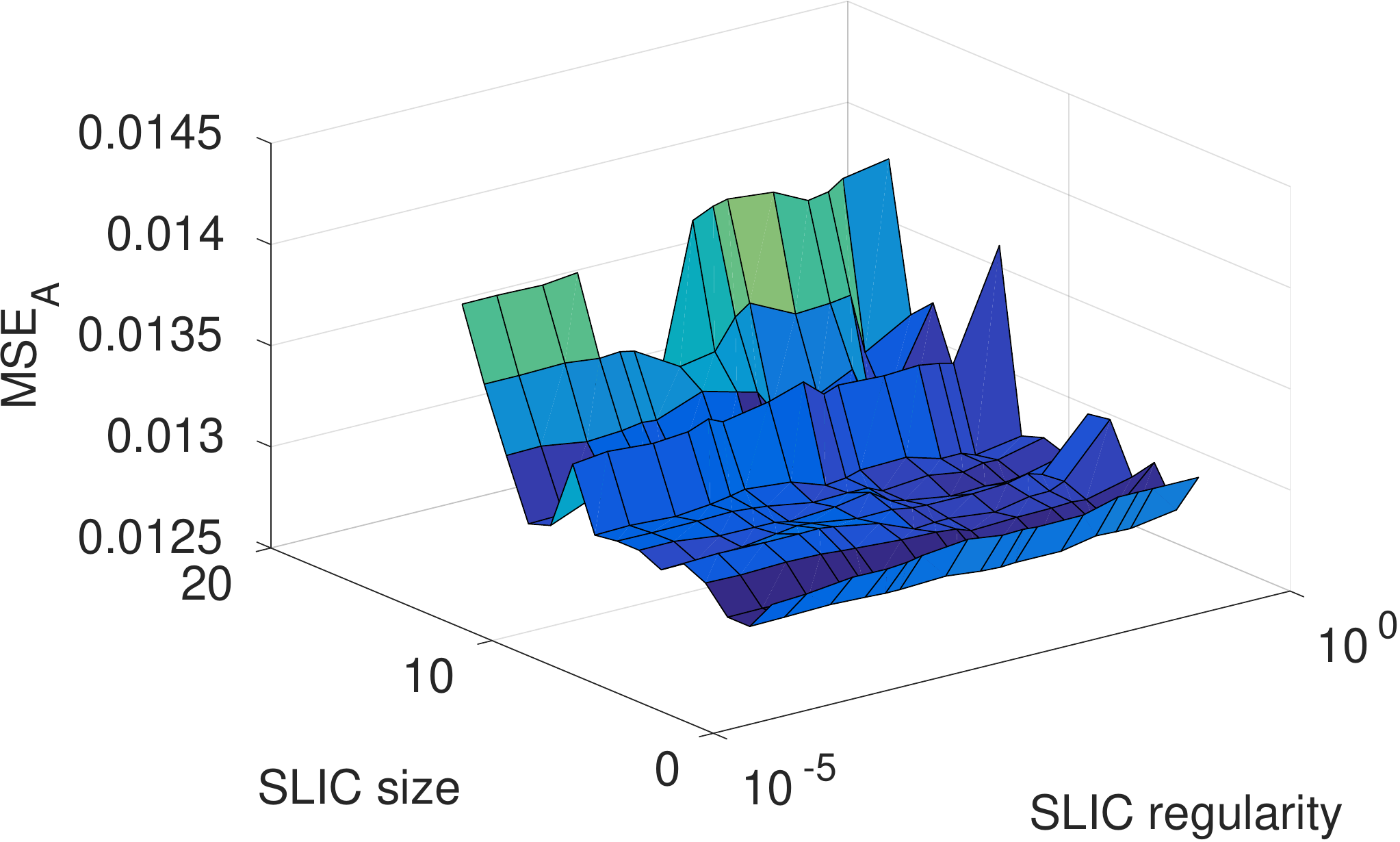}}}
\end{minipage}
\hspace{-0.75cm}
\caption{MSE variation due to relative changes in each parameter value about its optimal value (left) and MSE as a function of SLIC parameters~$\sqrt{N/S}$ and~$\gamma$ (right) for data cube DC1 with an SNR of 20dB.}
\label{fig:supp_sensitivity_1}
\end{figure}

\begin{figure} [!htbp]
\centering
\hspace{-1.0cm}
\begin{minipage}[b]{.4\linewidth}
  \centering
  \centerline{\includegraphics[width=1\linewidth]{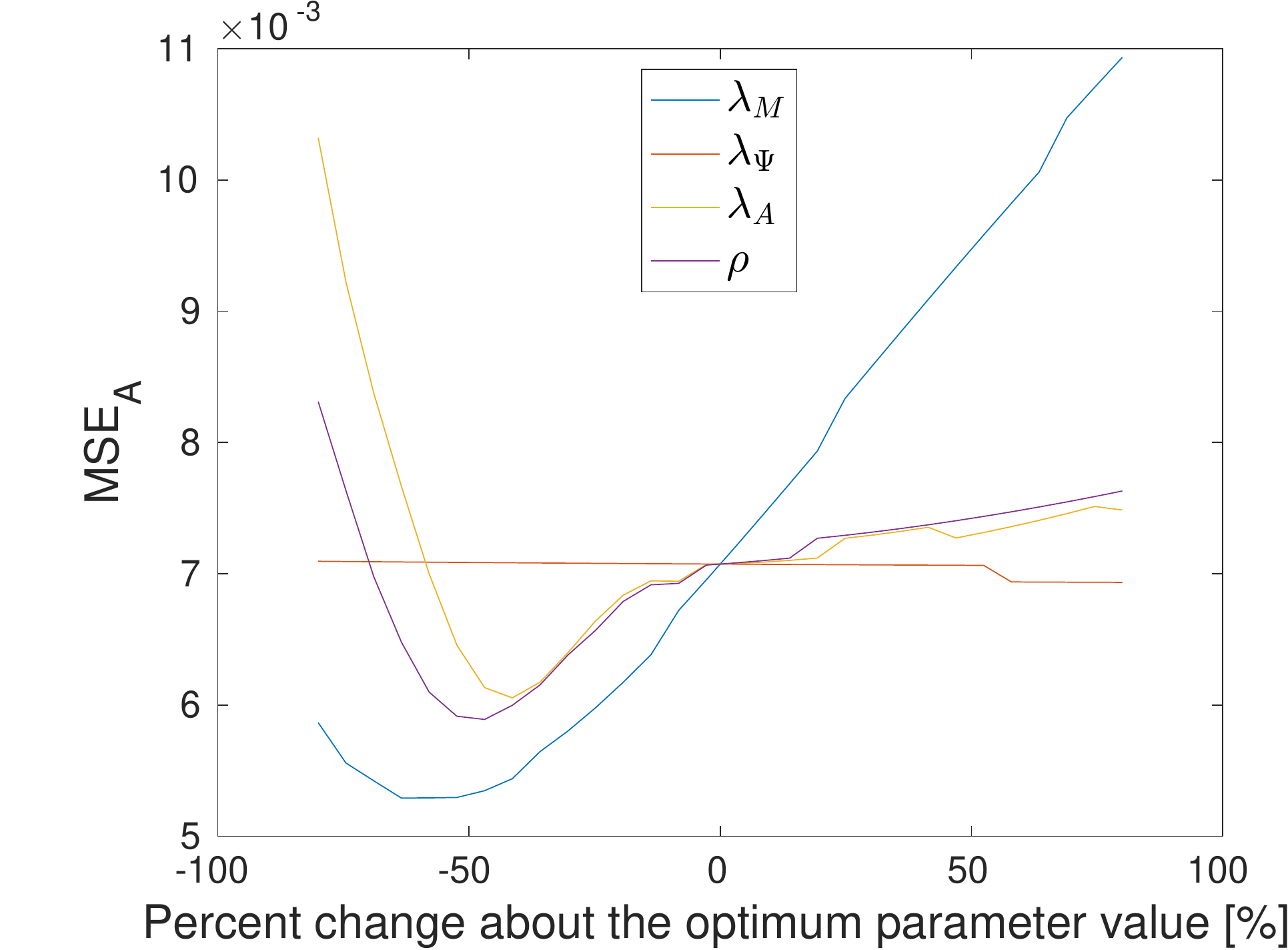}}
\end{minipage}
\begin{minipage}[b]{.43\linewidth}
  \centering
  \centerline{\raisebox{0.85cm}{\includegraphics[width=1\linewidth]{figures/sensitivity/SLIC_DC1_SNR30}}}
\end{minipage}
\hspace{-0.75cm}
\caption{MSE variation due to relative changes in each parameter value about its optimal value (left) and MSE as a function of SLIC parameters~$\sqrt{N/S}$ and~$\gamma$ (right) for data cube DC1 with an SNR of 30dB.}
\label{fig:supp_sensitivity_2}
\end{figure}

\begin{figure} [!htbp]
\centering
\hspace{-1.0cm}
\begin{minipage}[b]{.4\linewidth}
  \centering
  \centerline{\includegraphics[width=1\linewidth]{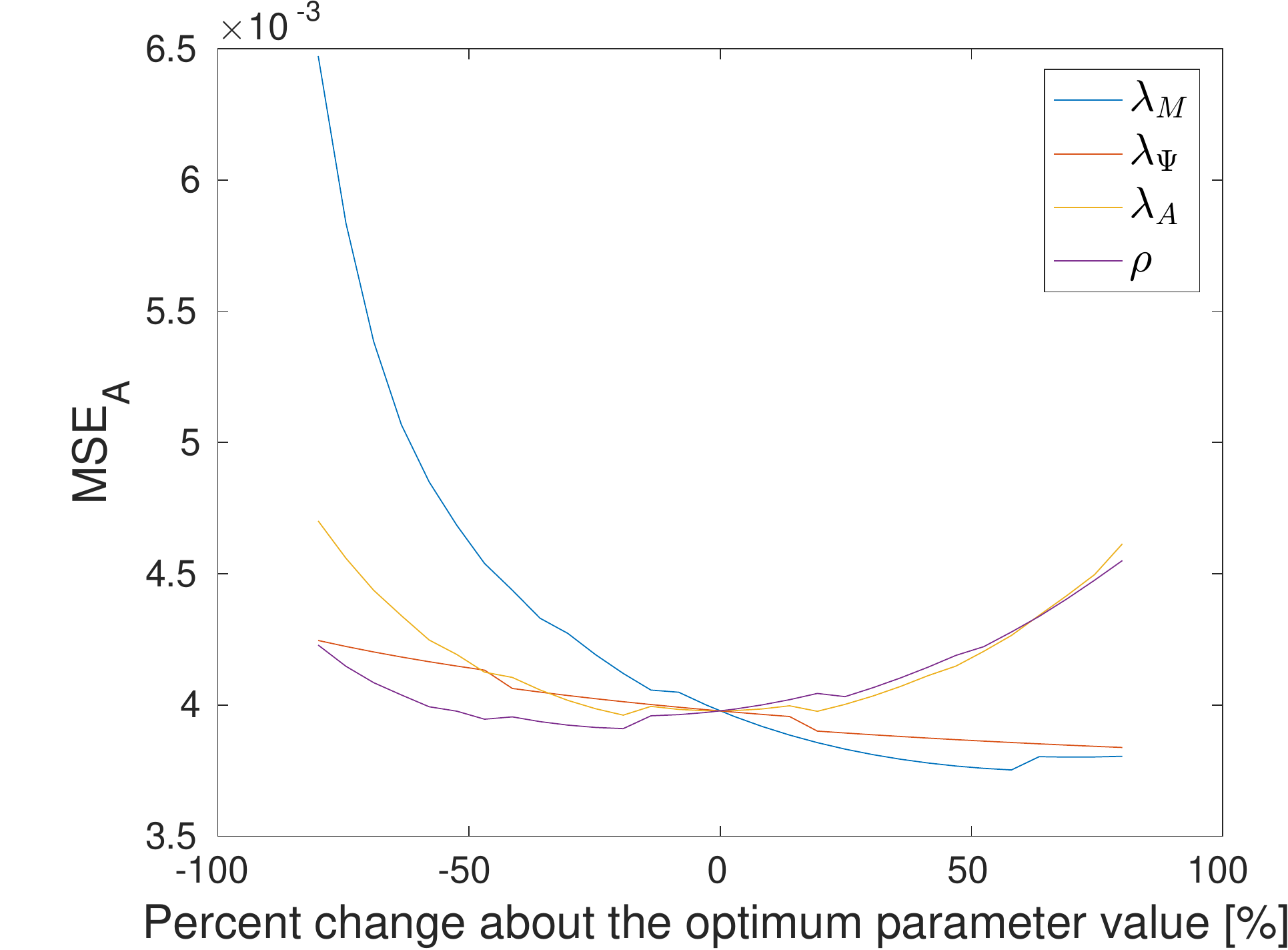}}
\end{minipage}
\begin{minipage}[b]{.43\linewidth}
  \centering
  \centerline{\raisebox{0.85cm}{\includegraphics[width=1\linewidth]{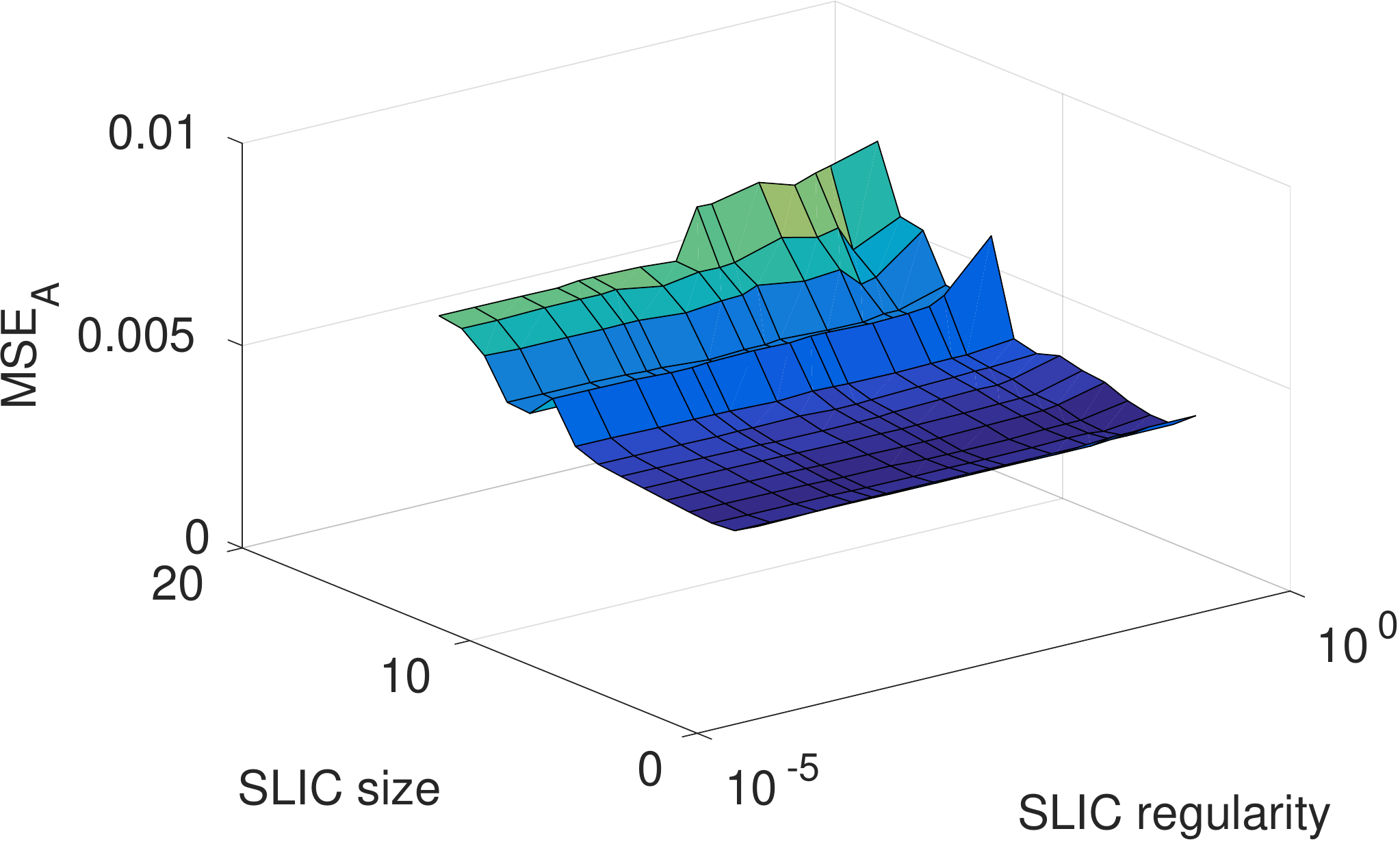}}}
\end{minipage}
\hspace{-0.75cm}
\caption{MSE variation due to relative changes in each parameter value about its optimal value (left) and MSE as a function of SLIC parameters~$\sqrt{N/S}$ and~$\gamma$ (right) for data cube DC1 with an SNR of 40dB.}
\label{fig:supp_sensitivity_3}
\end{figure}

\begin{figure} [!htbp]
\centering
\hspace{-1.0cm}
\begin{minipage}[b]{.4\linewidth}
  \centering
  \centerline{\includegraphics[width=1\linewidth]{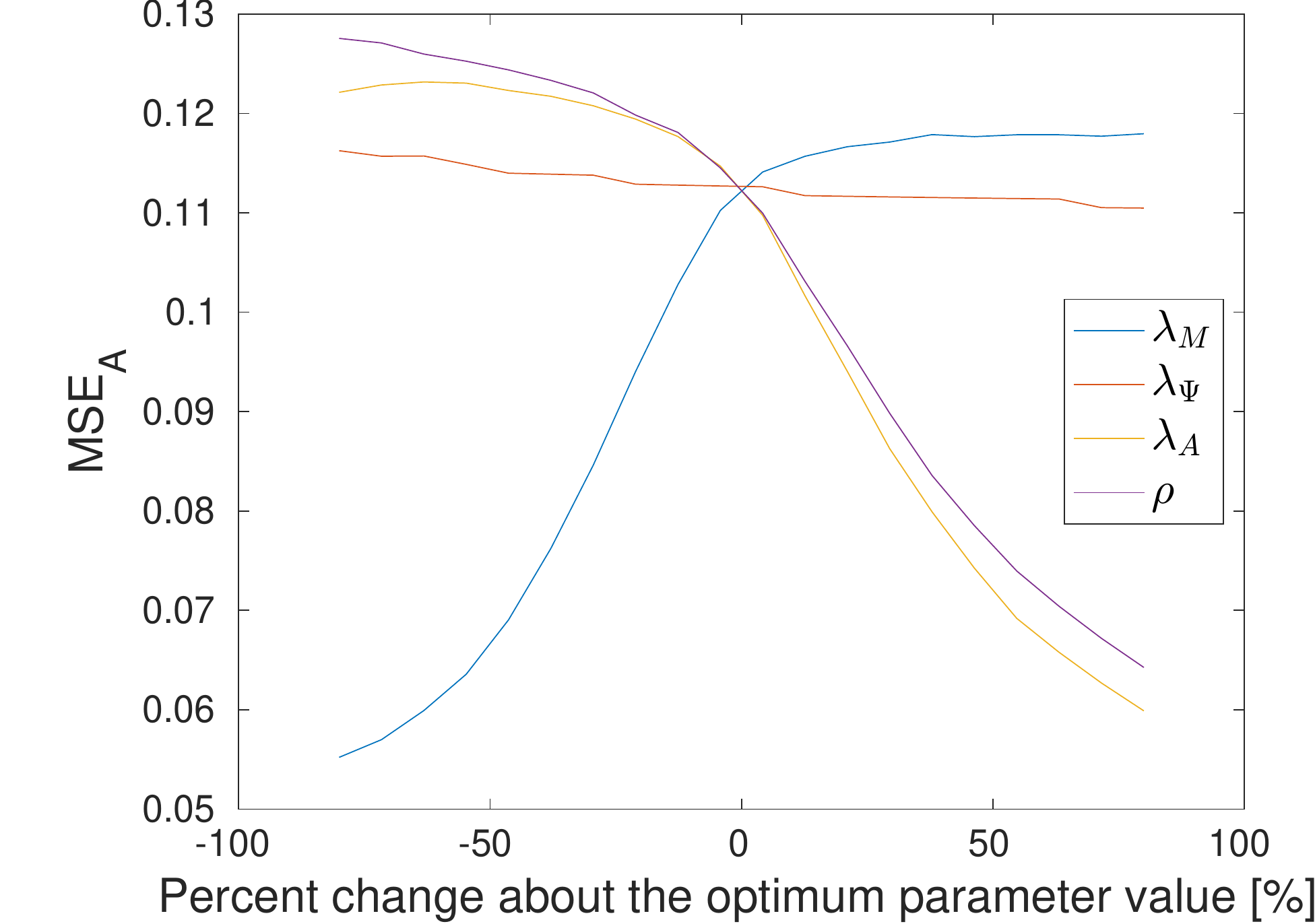}}
\end{minipage}
\begin{minipage}[b]{.43\linewidth}
  \centering
  \centerline{\raisebox{0.85cm}{\includegraphics[width=1\linewidth]{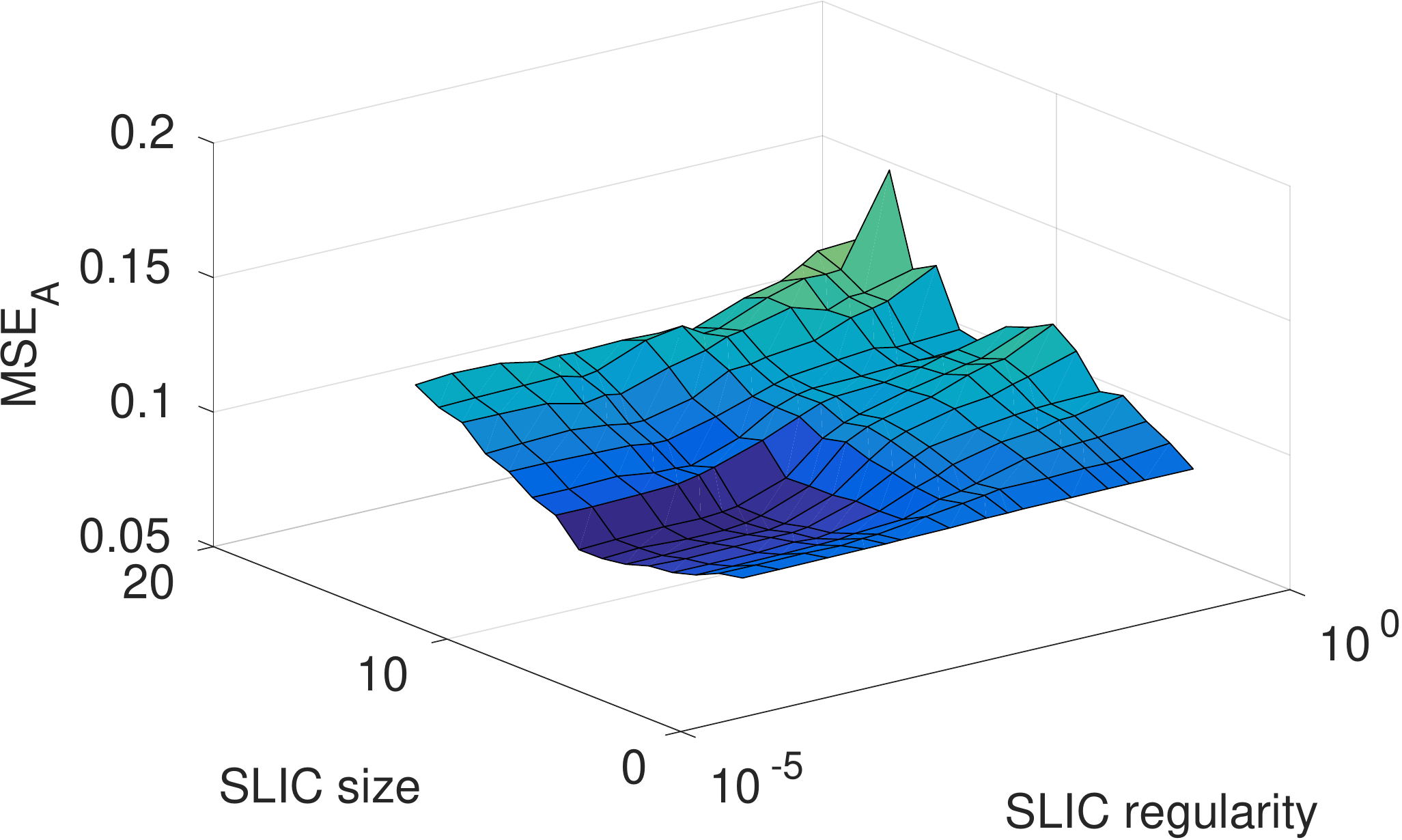}}}
\end{minipage}
\hspace{-0.75cm}
\caption{MSE variation due to relative changes in each parameter value about its optimal value (left) and MSE as a function of SLIC parameters~$\sqrt{N/S}$ and~$\gamma$ (right) for data cube DC2 with an SNR of 20dB.}
\label{fig:supp_sensitivity_4}
\end{figure}

\begin{figure} [!htbp]
\centering
\hspace{-1.0cm}
\begin{minipage}[b]{.4\linewidth}
  \centering
  \centerline{\includegraphics[width=1\linewidth]{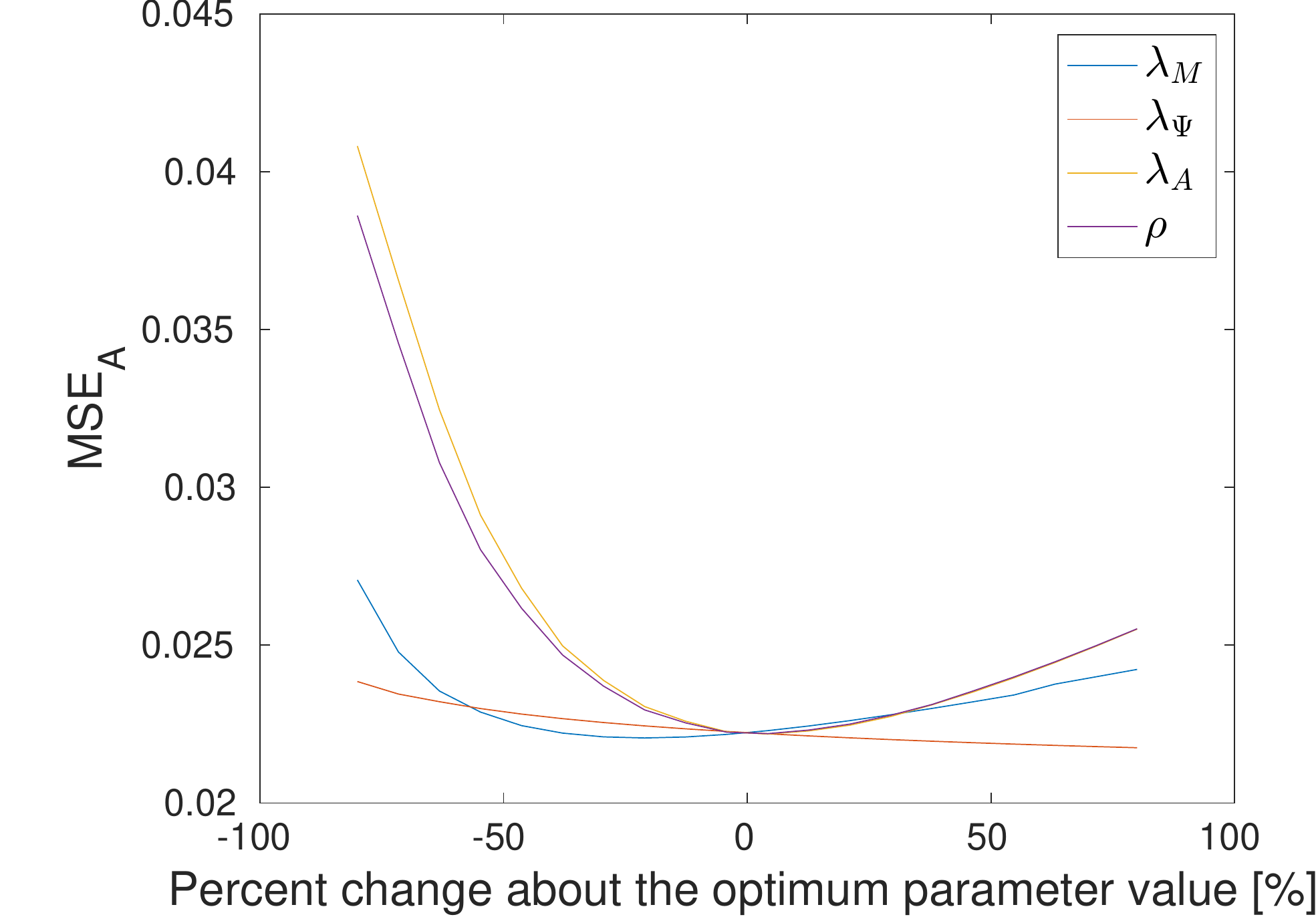}}
\end{minipage}
\begin{minipage}[b]{.43\linewidth}
  \centering
  \centerline{\raisebox{0.85cm}{\includegraphics[width=1\linewidth]{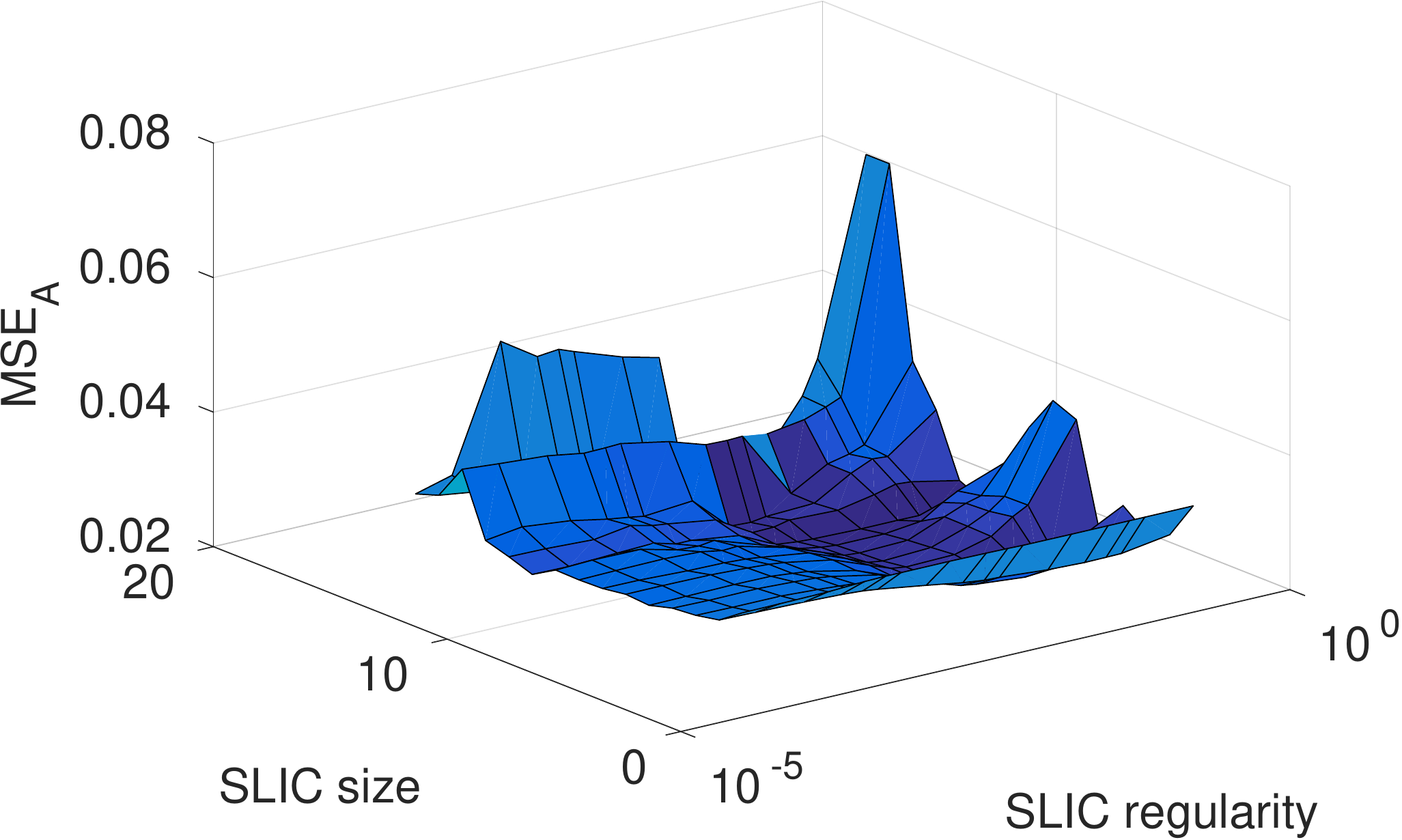}}}
\end{minipage}
\hspace{-0.75cm}
\caption{MSE variation due to relative changes in each parameter value about its optimal value (left) and MSE as a function of SLIC parameters~$\sqrt{N/S}$ and~$\gamma$ (right) for data cube DC2 with an SNR of 30dB.}
\label{fig:supp_sensitivity_5}
\end{figure}

\begin{figure} [!htbp]
\centering
\hspace{-1.0cm}
\begin{minipage}[b]{.4\linewidth}
  \centering
  \centerline{\includegraphics[width=1\linewidth]{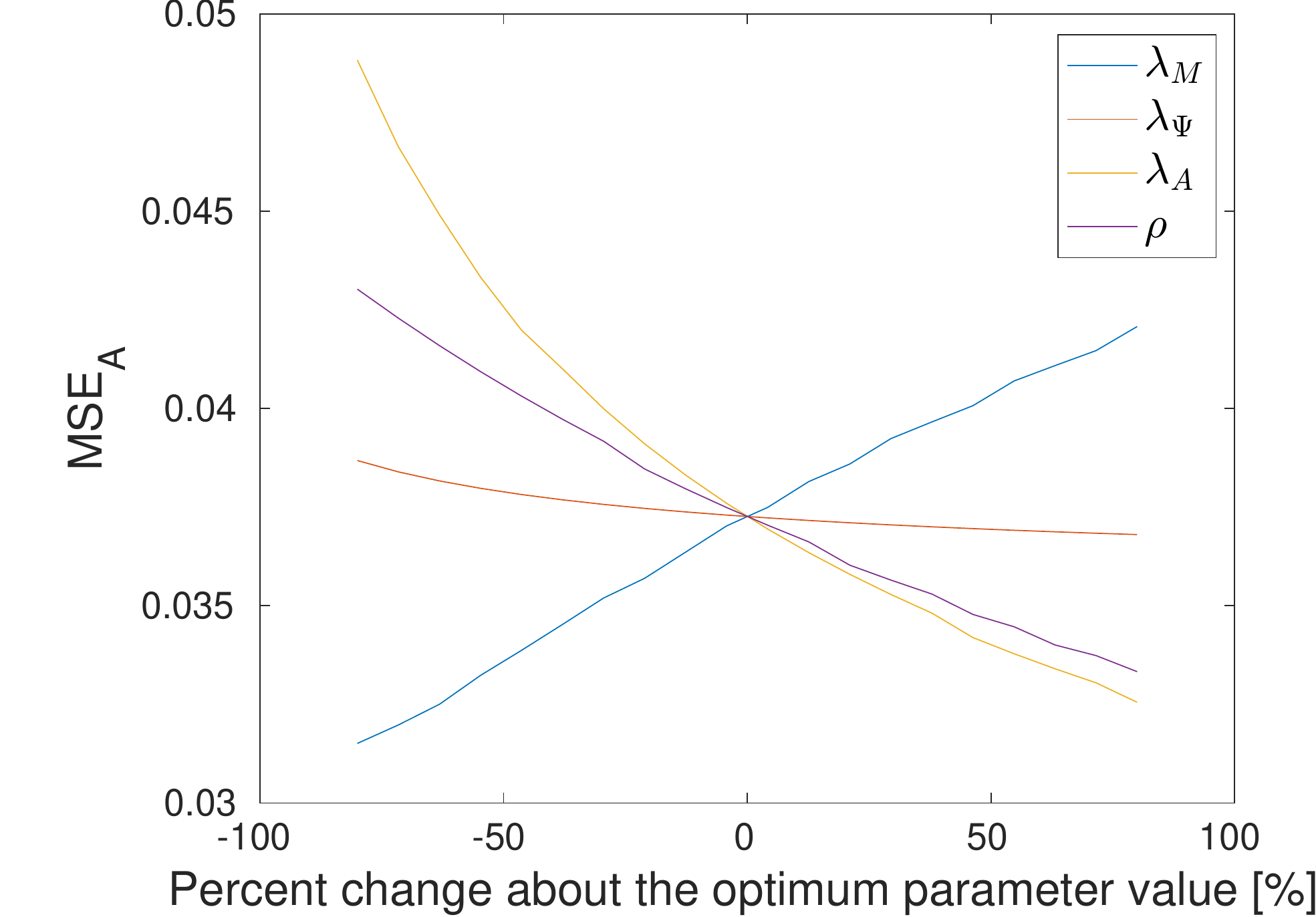}}
\end{minipage}
\begin{minipage}[b]{.43\linewidth}
  \centering
  \centerline{\raisebox{0.85cm}{\includegraphics[width=1\linewidth]{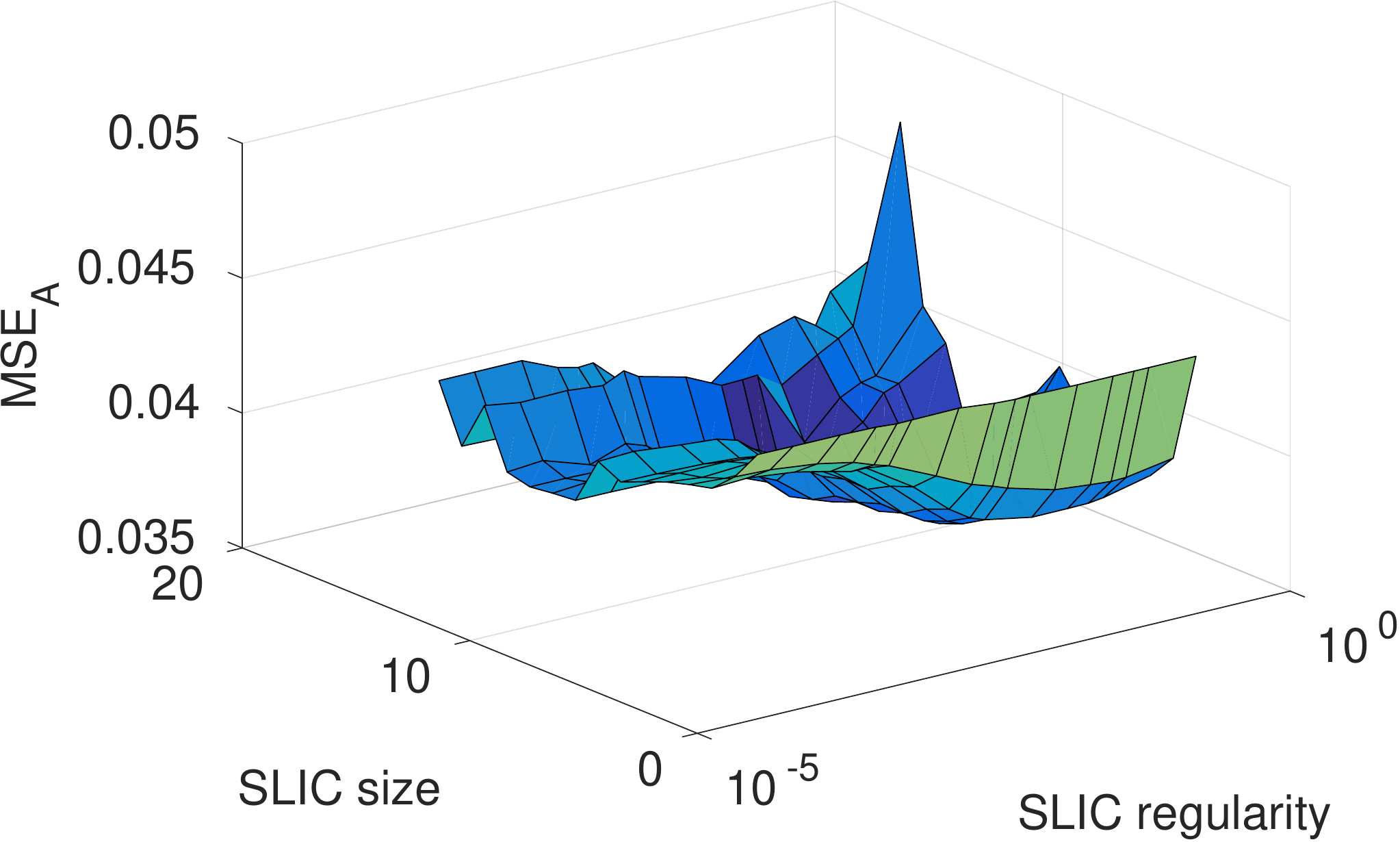}}}
\end{minipage}
\hspace{-0.75cm}
\caption{MSE variation due to relative changes in each parameter value about its optimal value (left) and MSE as a function of SLIC parameters~$\sqrt{N/S}$ and~$\gamma$ (right) for data cube DC2 with an SNR of 40dB.}
\label{fig:supp_sensitivity_6}
\end{figure}

\begin{figure} [!htbp]
\centering
\hspace{-1.0cm}
\begin{minipage}[b]{.4\linewidth}
  \centering
  \centerline{\includegraphics[width=1\linewidth]{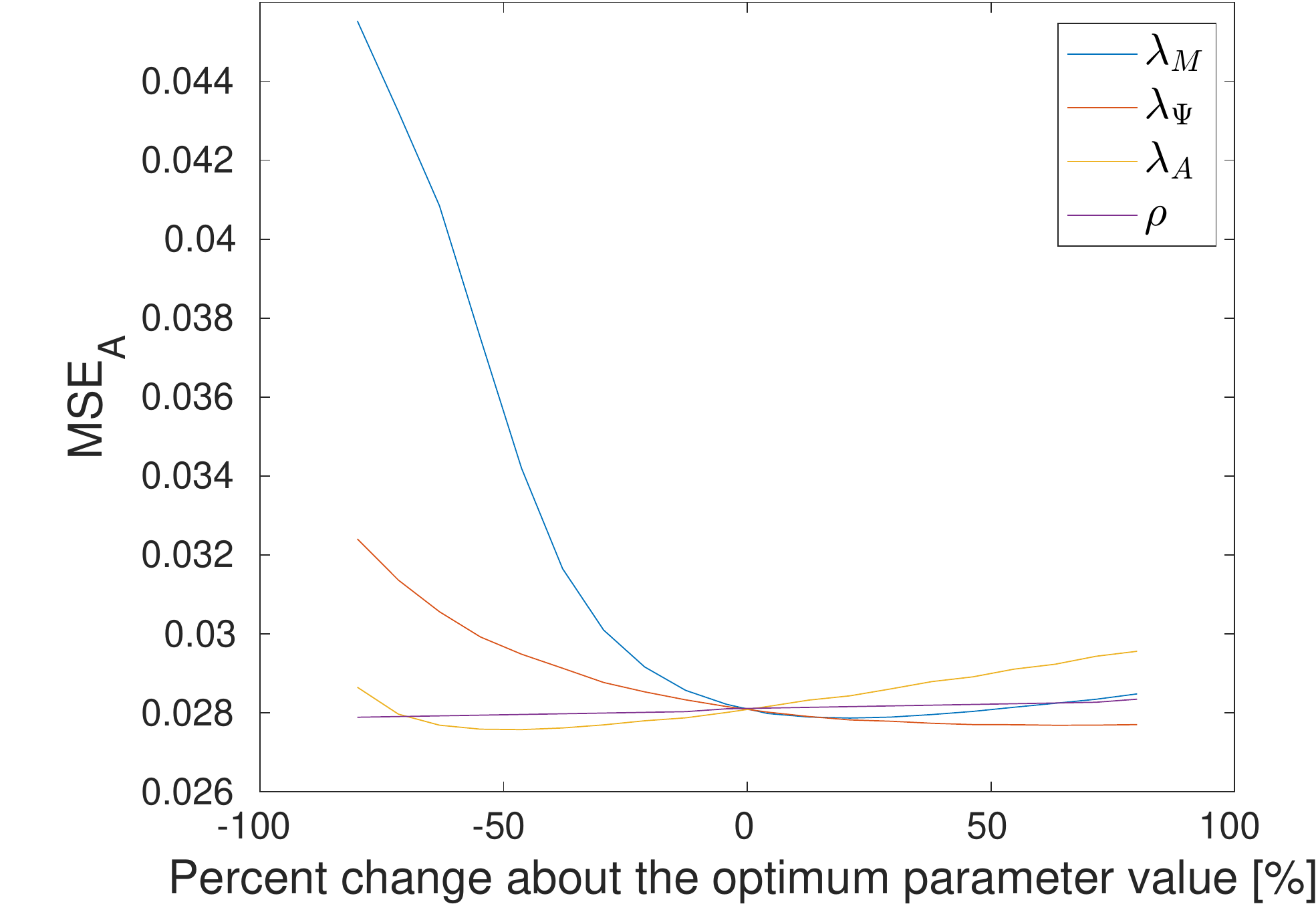}}
\end{minipage}
\begin{minipage}[b]{.43\linewidth}
  \centering
  \centerline{\raisebox{0.85cm}{\includegraphics[width=1\linewidth]{figures/sensitivity/SLIC_DC2_SNR30}}}
\end{minipage}
\hspace{-0.75cm}
\caption{MSE variation due to relative changes in each parameter value about its optimal value (left) and MSE as a function of SLIC parameters~$\sqrt{N/S}$ and~$\gamma$ (right) for data cube DC3 with an SNR of 20dB.}
\label{fig:supp_sensitivity_7}
\end{figure}

\begin{figure} [!htbp]
\centering
\hspace{-1.0cm}
\begin{minipage}[b]{.4\linewidth}
  \centering
  \centerline{\includegraphics[width=1\linewidth]{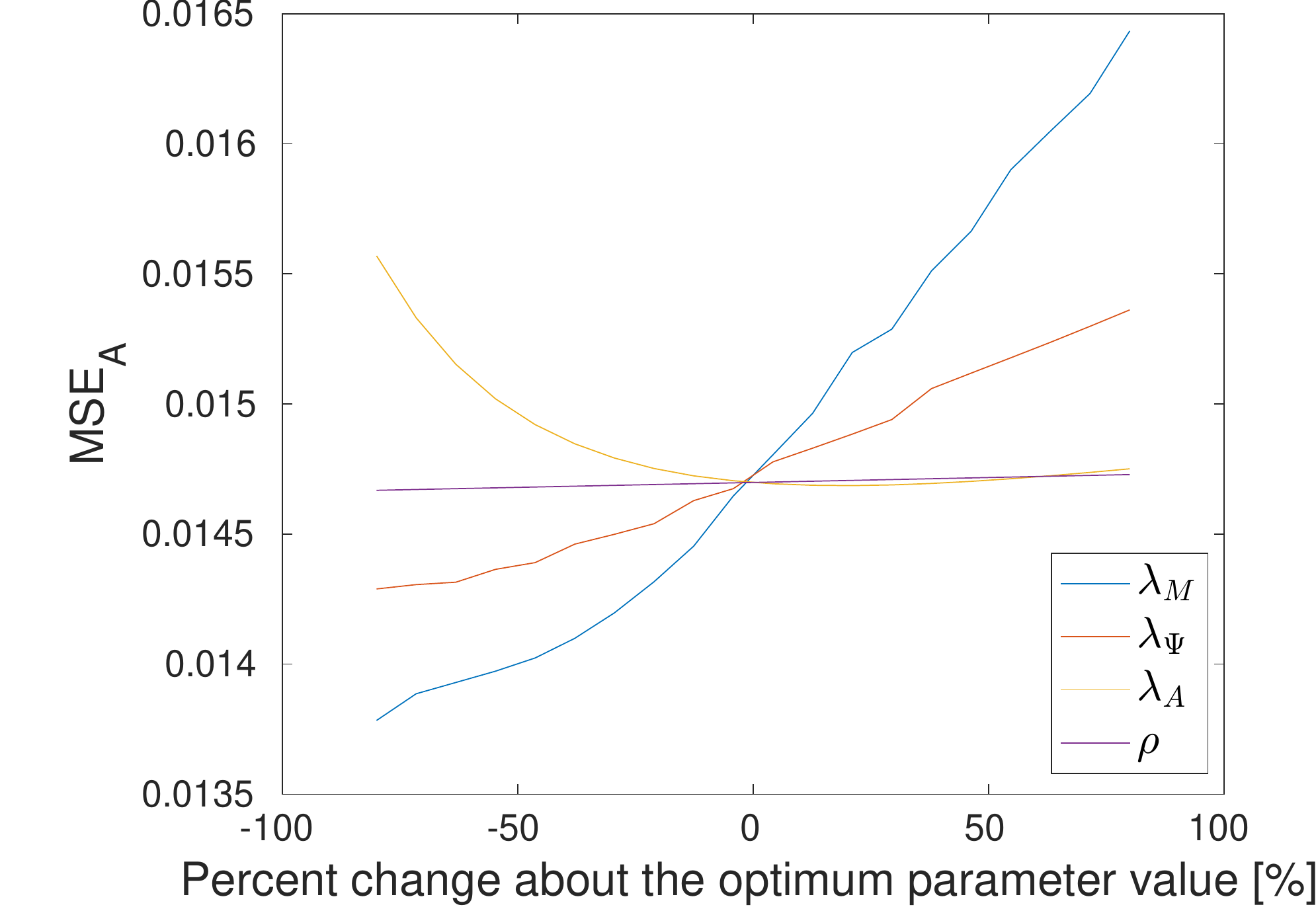}}
\end{minipage}
\begin{minipage}[b]{.43\linewidth}
  \centering
  \centerline{\raisebox{0.85cm}{\includegraphics[width=1\linewidth]{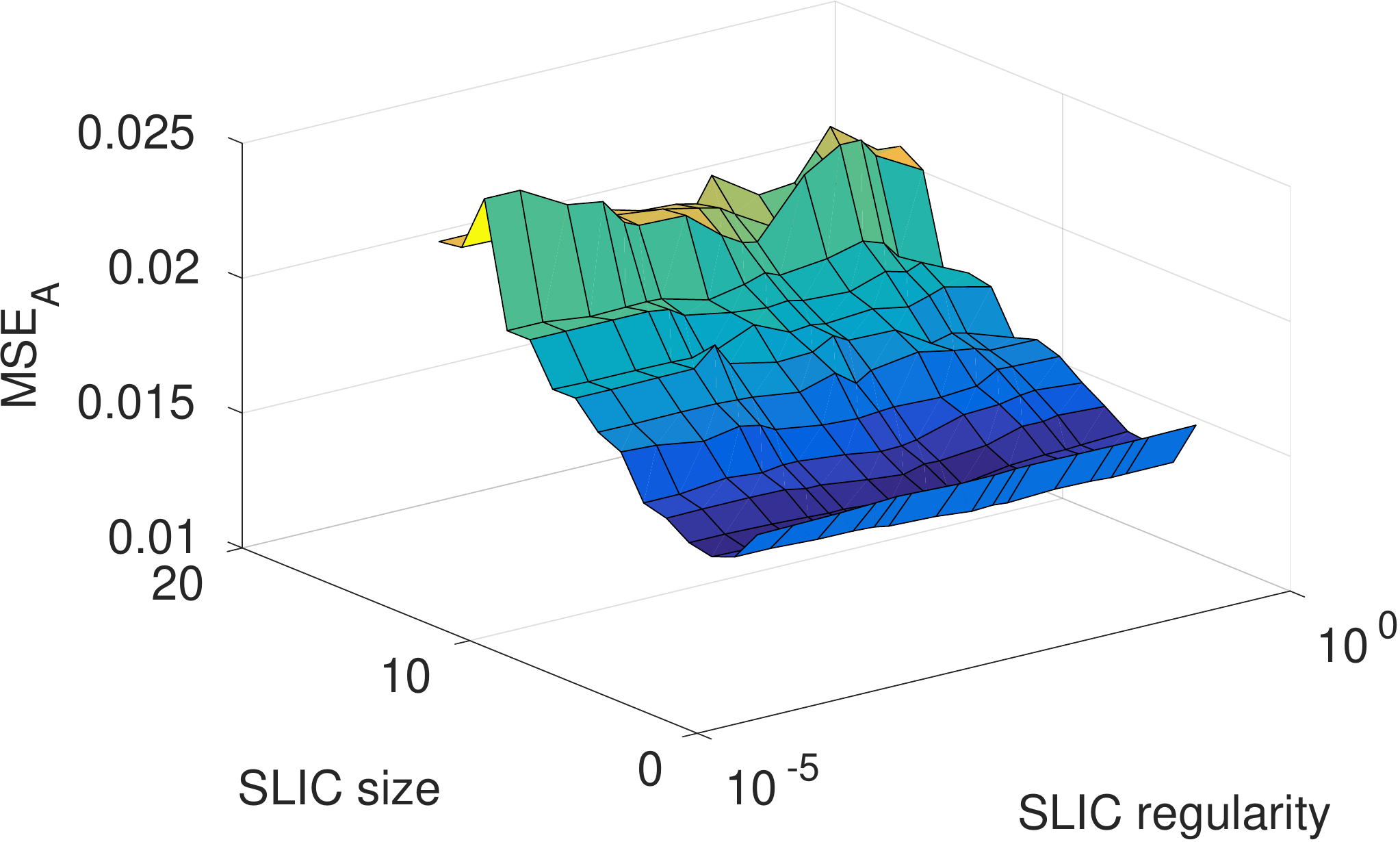}}}
\end{minipage}
\hspace{-0.75cm}
\caption{MSE variation due to relative changes in each parameter value about its optimal value (left) and MSE as a function of SLIC parameters~$\sqrt{N/S}$ and~$\gamma$ (right) for data cube DC3 with an SNR of 30dB.}
\label{fig:supp_sensitivity_8}
\end{figure}

\begin{figure} [!htbp]
\centering
\hspace{-1.0cm}
\begin{minipage}[b]{.4\linewidth}
  \centering
  \centerline{\includegraphics[width=1\linewidth]{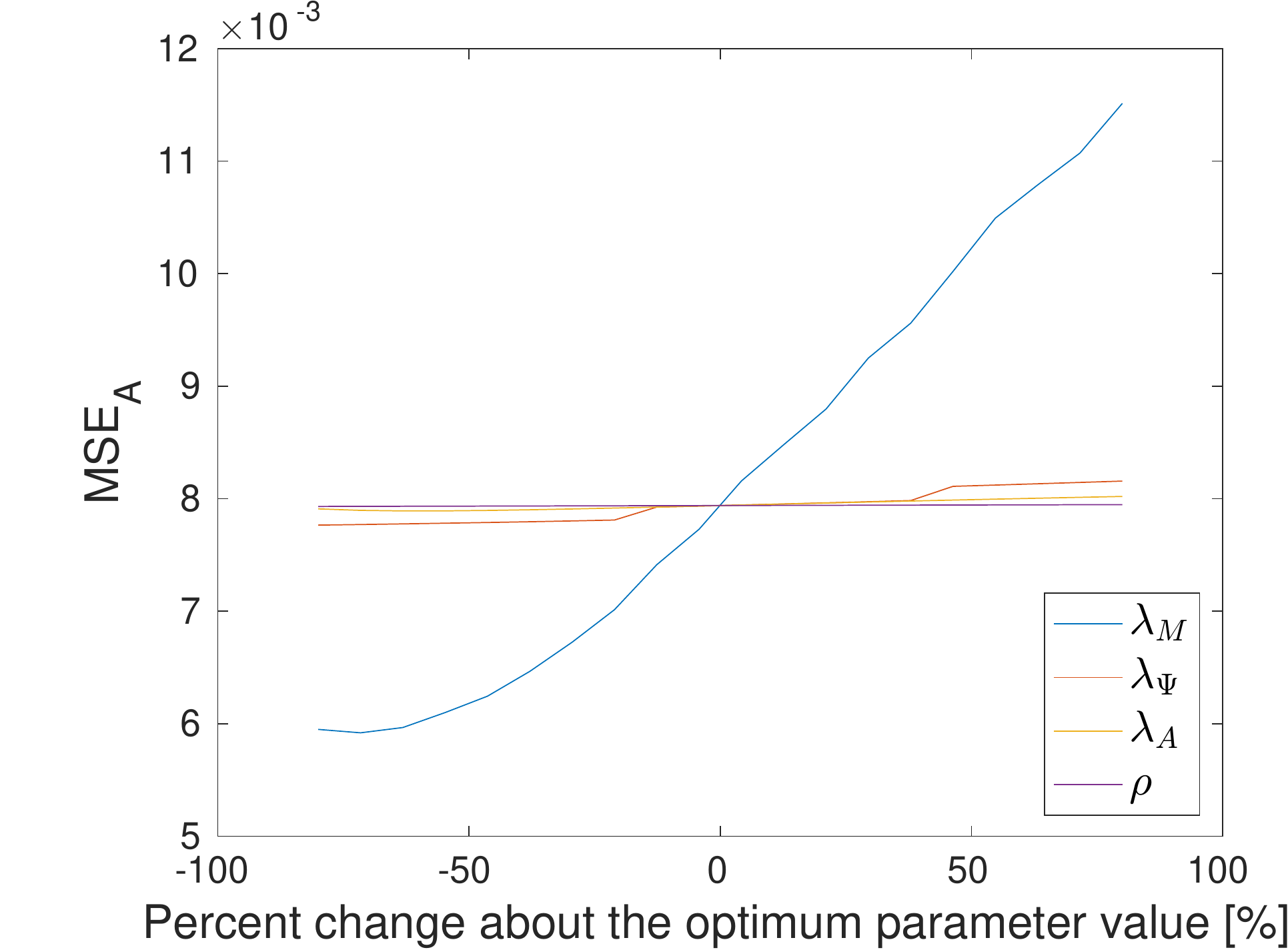}}
\end{minipage}
\begin{minipage}[b]{.43\linewidth}
  \centering
  \centerline{\raisebox{0.85cm}{\includegraphics[width=1\linewidth]{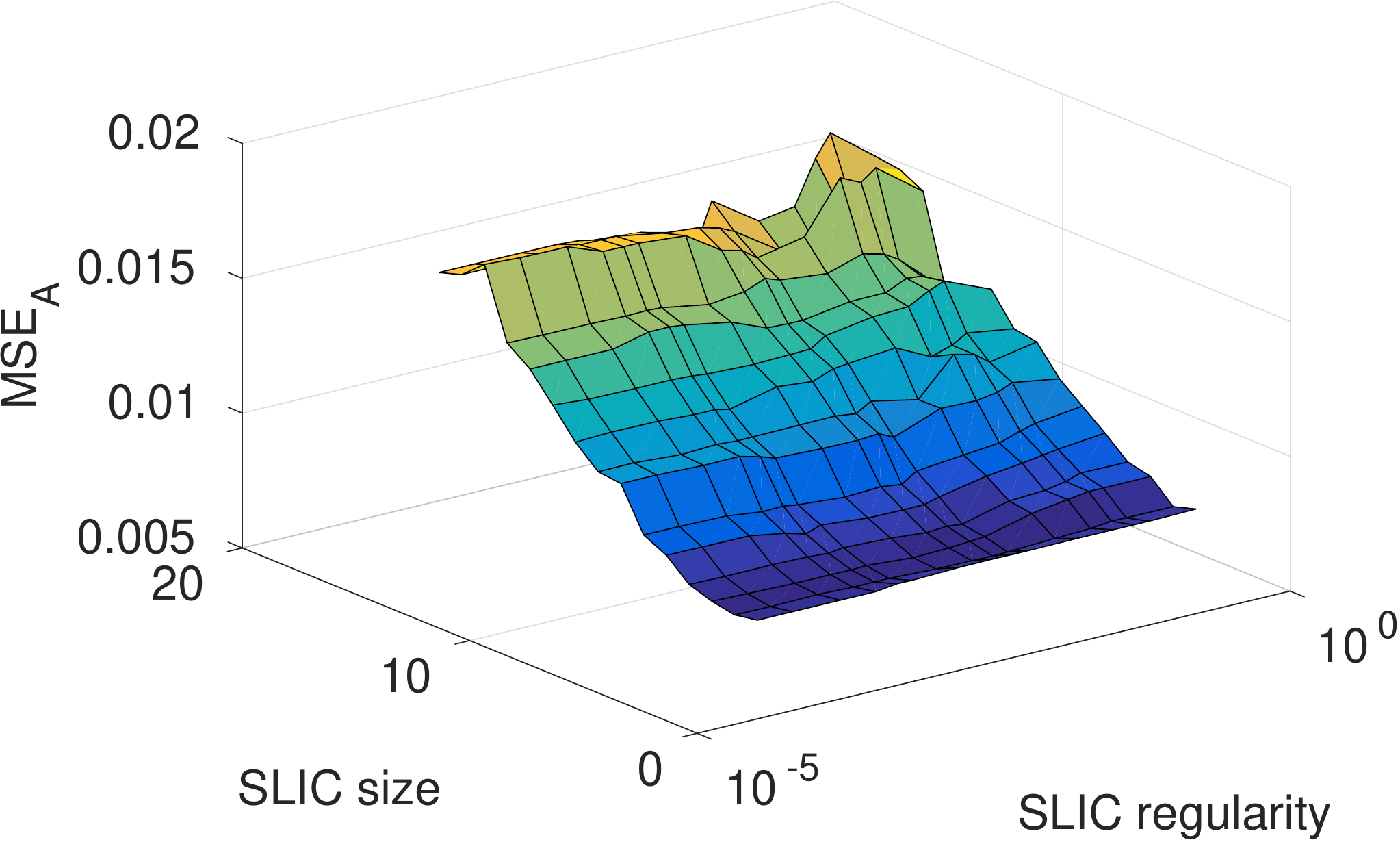}}}
\end{minipage}
\hspace{-0.75cm}
\caption{MSE variation due to relative changes in each parameter value about its optimal value (left) and MSE as a function of SLIC parameters~$\sqrt{N/S}$ and~$\gamma$ (right) for data cube DC3 with an SNR of 40dB.}
\label{fig:supp_sensitivity_9}
\end{figure}

The simulations discussed in Section V.B in the manuscript are replicated here for all datasets DC1, DC2 and DC3, and all SNR values 20, 30 and 40dB. Figures~\ref{fig:supp_sensitivity_1},~\ref{fig:supp_sensitivity_2} and~\ref{fig:supp_sensitivity_3} present the sensitivity for the DC1 data cube, Figures~\ref{fig:supp_sensitivity_4},~\ref{fig:supp_sensitivity_5} and~\ref{fig:supp_sensitivity_6} for the DC2 data cube, and Figures~\ref{fig:supp_sensitivity_7},~\ref{fig:supp_sensitivity_8} and~\ref{fig:supp_sensitivity_9} for the DC3 data cube, for SNRs of 20, 30 and 40dB respectively. The results corroborate the discussion presented in Section V.B of the manuscript.

\bigskip\bigskip

\centerline{REFERENCES}

\begin{itemize}
    \item[{[S1]}] R. Achanta, A. Shaji, K. Smith, A. Lucchi, P. Fua, and S. Süsstrunk, “SLIC superpixels compared to state-of-the-art superpixel methods,” IEEE transactions on pattern analysis and machine intelligence, vol. 34, no. 11, pp. 2274–2282, 2012.
    \item[{[S2]}] B. Hapke, Theory of Reflectance and Emittance Spectroscopy. Cambridge University Press, 1993.
    \item[{[S3]}] L. Drumetz, M.-A. Veganzones, S. Henrot, R. Phlypo, J. Chanussot, and C. Jutten, “Blind hyperspectral unmixing using an extended linear mixing model to address spectral variability,” IEEE Transactions on Image Processing, vol. 25, no. 8, pp. 3890–3905, 2016.
    \item[{[S4]}] R. Webster, P. Curran, and J. Munden, “Spatial correlation in reflected radiation from the ground and its implications for sampling and mapping by ground-based radiometry,” Remote sensing of environment, vol. 29, no. 1, pp. 67–78, 1989.
    \item[{[S5]}] E. Tola, K. Al-Gaadi, R. Madugundu, A. Zeyada, A. Kayad, and C. Biradar, “Characterization of spatial variability of soil physicochemical properties and its impact on rhodes grass productivity,” Saudi journal of biological sciences, vol. 24, no. 2, pp. 421–429, 2017.
    \item[{[S6]}] A. Najafian, M. Dayani, H. R. Motaghian, and H. Nadian, “Geostatistical assessment of the spatial distribution of some chemical properties in calcareous soils,” Journal of Integrative Agriculture, vol. 11, no. 10, pp. 1729–1737, 2012.
    \item[{[S7]}] Y.-C. Wei, Y.-L. Bai, J.-Y. Jin, F. Zhang, L.-P. Zhang, and X.-Q. Liu, “Spatial variability of soil chemical properties in the reclaiming marine foreland to yellow sea of china,” Agricultural Sciences in China, vol. 8, no. 9, pp. 1103–1111, 2009.
    \item[{[S8]}] J. Hou-Long, L. Guo-Shun, W. Xin-Zhong, S. Wen-Feng, Z. Rui-Na, Z. Chun-Hua, H. Hong-Chao, and L. Yan-Tao, “Spatial variability of soil properties in a long-term tobacco plantation in central china,” Soil Science, vol. 175, no. 3, pp. 137–144, 2010.
    \item[{[S9]}] J. K. Crowley, “Visible and near-infrared spectra of carbonate rocks: Reflectance variations related to petrographic texture and impurities,” Journal of Geophysical Research: Solid Earth, vol. 91, no. B5, pp. 5001–5012, 1986.
    \item[{[S10]}] R. N. Clark, “Spectroscopy of rocks and minerals, and principles of spectroscopy,” in Remote Sensing for the Earth Sciences: Manual of Remote Sensing, A. N. Rencz, Ed. New York, NY, USA: Wiley, 1999, vol. 3, pp. 3–58.
\end{itemize}

\end{document}